\documentclass[10pt, journal]{IEEEtran} % DO NOT CHANGE THIS

\usepackage{graphicx}
\usepackage{natbib}
\usepackage{framed,multirow}
\usepackage{amssymb}
\usepackage{amsmath}
\usepackage{latexsym}
\usepackage{subfig}
\usepackage{algorithm}
\usepackage{algpseudocode}
\usepackage{soul}
\usepackage{enumitem}
\usepackage{url}
\usepackage{xcolor}
\usepackage{hyperref}
\usepackage{newfloat}
\usepackage{listings}
\usepackage{float}

\begin{document}

\setcounter{page}{1}

%%%%%%%%  TITLE  %%%%%%%%%%%%%%%%%%%%%%%%%%%%%%%%%%%%%%%%%%%%%%%%%
\title{DRG-Net: Interactive Joint Learning of Multi-lesion Segmentation and Classification for Diabetic Retinopathy Grading}

\author{Hasan Md Tusfiqur\textsuperscript{\rm 1,*}, Duy M. H. Nguyen\textsuperscript{\rm 1,2 *}, Mai T. N. Truong\textsuperscript{\rm 3}, Triet A. Nguyen\textsuperscript{\rm 4}, Binh T. Nguyen\textsuperscript{\rm 5}, \\
Michael Barz\textsuperscript{\rm 1}, Hans-Jürgen Profitlich\textsuperscript{\rm 1}, Ngoc T. T. Than\textsuperscript{\rm 6}, Ngan Le\textsuperscript{\rm 7}, Pengtao Xie\textsuperscript{\rm 8}, Daniel Sonntag\textsuperscript{\rm 1,9}%
	\thanks{\textsuperscript{\rm *}\textit{Equal contribution}}
	\thanks{\textsuperscript{\rm 1}\textit{German Research Center for Artificial Intelligence (DFKI)}}
 \thanks{\textsuperscript{\rm 2}\textit{Machine Learning and Simulation Lab, University of Stuttgart, Germany}}
 \thanks{\textsuperscript{\rm 3}\textit{Department of Multimedia Engineering, Dongguk University, South Korea}}
 \thanks{\textsuperscript{\rm 4}\textit{Department of Mathematics,  University of Architecture Ho Chi Minh City}}
 \thanks{\textsuperscript{\rm 5}\textit{AISIA Lab, University of Science - VNUHCM, Vietnam}}
 \thanks{\textsuperscript{\rm 6}\textit{Byers Eye Institute, Stanford University, USA}}
 \thanks{\textsuperscript{\rm 7}\textit{Department of Computer Science and Computer Engineering, University of Arkansas, USA}}
 \thanks{\textsuperscript{\rm 8}\textit{Department of Electrical and Computer Engineering, University of California San Diego, USA,}}
\thanks{\textsuperscript{\rm 9}\textit{Applied Artificial Intelligence, Oldenburg University, Germany}}}
 
\maketitle
%%%%%%%%%%%%%%%%%%%%%%%%%%%%%%%%%%%%%%%%%%%%%%%%%%%%%%%%%%%%%%%%%%

\begin{abstract}
Diabetic Retinopathy (DR) is a leading cause of vision loss in the world, and early DR detection is necessary to prevent vision loss and support an appropriate treatment. In this work, we leverage interactive machine learning and introduce a joint learning framework, termed DRG-Net, to effectively learn both disease grading and multi-lesion segmentation. Our DRG-Net consists of two modules: (i) DRG-AI-System to classify DR Grading, localize lesion areas and provide visual explanation and (ii) DRG-Expert-Interaction to receive feedback from user-expert and improve the DRG-AI-System. To deal with sparse data, we utilize transfer learning mechanisms to extract invariant feature representations by using Wasserstein distance and adversarial learning-based entropy minimization. Besides, we propose a novel attention strategy at both low- and high-level features to automatically select the most significant lesion information and provide explainable properties. In terms of human interaction, we further develop DRG-Net as a tool that enables expert users to correct the system’s predictions, which may then be used to update the system as a whole.
%\hl{Finally, the strategy can reduce perturbations in labels made by users using attention networks, thereby saving time and accelerating the data annotation process.  Duy: What does this mean?} 
Moreover, thanks to the attention mechanism and loss functions constraint between lesion features and classification features, our approach can be robust given a certain level of noise in the feedback of users. We have benchmarked DRG-Net on the two largest DR datasets, i.e., IDRID and FGADR, and compared it to various state-of-the-art deep learning networks. In addition to  outperforming other SOTA approaches, DRG-Net is effectively updated using user feedback, even in a weakly-supervised manner.
%The empirical experiments validate out method: we outperform common baselines of state-of-the-art systems by a significant margin. Also, our system’s performance improves over time when more user feedback is fed into the network, even in a weakly-supervised form.
\end{abstract}

% ============================ Introduction
\section{Introduction}
\label{sec-intro}
\begin{figure}[!t]
 \centering
\includegraphics[width=0.95\columnwidth]{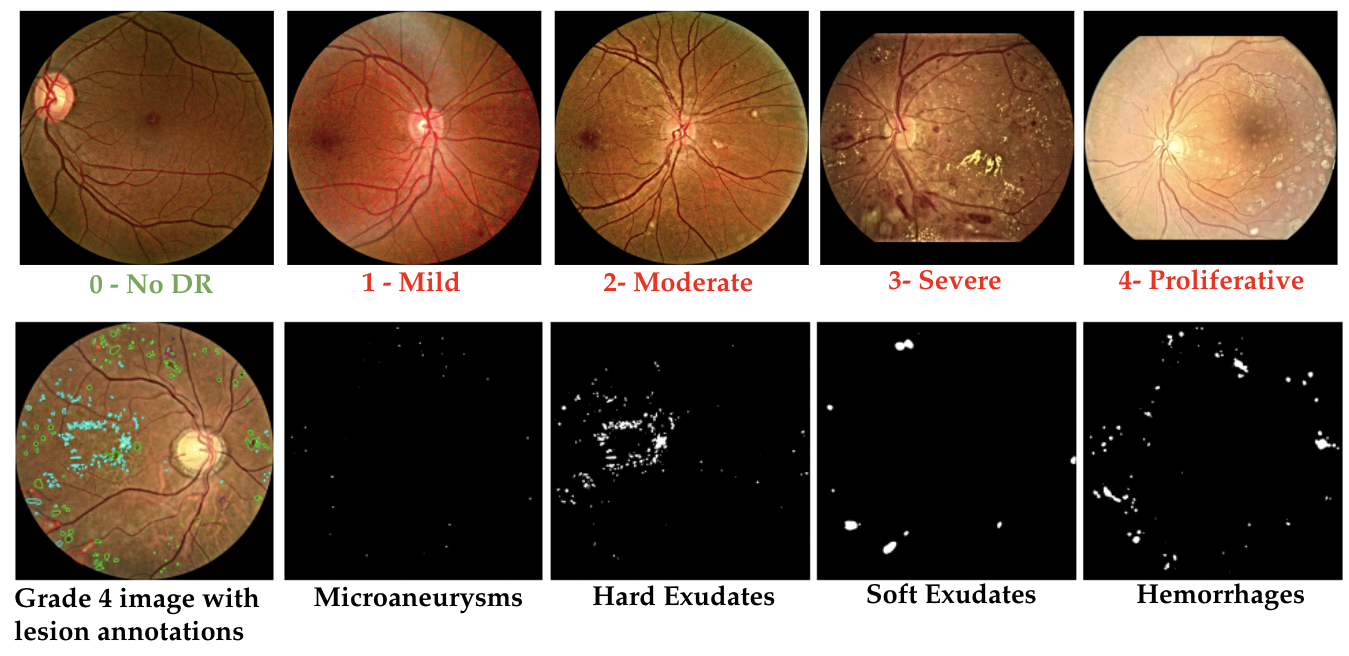}\\
    \caption{Illustration of diabetic retinopathy retinal images. Top row - examples of different stages of DR disease. Bottom row - a retinal image with lesion information annotated by domain expert (left most image). The remaining 4 images are lesion mask predicted by the Dense U-Net Segmentation model \cite{zhou2020benchmark, li2018h}.}
    \label{fig:retina_example}
\end{figure}

\textit{Diabetes} is a chronic health condition that is estimated to affect about one in every ten people worldwide \cite{saeedi2019global}. According to \citet{yang2021robust},
40\% to 45\% of people with diabetes may develop \textit{Diabetic Retinopathy (DR)} during their lifetime. DR is a kind of ocular disease that damages the retina's blood vessels, and it is one of the leading causes of irreversible blindness. Although the symptoms are diagnosed only in the later stage, people suffering from this disease start to lose vision from an early stage.
To assess the complexity of DR, \textit{International Clinical Diabetic Retinopathy Disease Severity Scale}  \cite{gulshan2016development} has used five grades (0-4), including \textit{no DR, mild, moderate, severe, and proliferative}  as shown in Figure \ref{fig:retina_example}. In practice, accurate DR grading is a time-consuming task and there is a huge shortage of qualified ophthalmologists. Therefore, an automated system can significantly aid the diagnosis process.

While the existing methods \cite{gulshan2016development,antal2012ensemble,wang2017zoom} can assist ophthalmologists in validating and interpreting the automated prediction, they introduce three main issues: (i) They are designed as black-box models and unable to utilize lesion features to improve the models. (ii) Lesion segmentation performance is limited because of irregularities in shape, noisy, ill-defined boundaries, intra-class lesion diversities and less labeled data, (iii) Though disease classification and lesion discovery are strongly correlated, they are mostly studied as independent tasks.

%\textcolor{red}{Ngoc: reasoning why segmentation multi-lesions is important for DR-Grading tasks (general descriptions), for e.g. show a case study for 3-severse and add references.}

Joint learning has recently shown promising progress in medical imaging e.g., breast cancer \cite{mehta2018ynet}, skin lesion \cite{al2020multiple}, COVID-19 detection \cite{nguyen2021attention,Wu_2021}, and retinal blood vessel \cite{li2020joint}. However, the performance of joint learning-based medical imaging is generally limited due to domain-shift and scarcity of labeled data. To address the aforementioned limitations, we propose a joint learning-based multi-lesion segmentation and classification for DR Grading, termed DRG-Net. Our DRG-Net includes a novel \textit{Domain Invariant Lesion Feature Generator} and \textit{Attention-Based Disease Grading Classifier}. While the former aims to identify and distinguish important lesion regions of different semantics, the later one aims to incorporates the predicted lesion information in the learning process of the disease classification network. Furthermore, our framework is designed as a collaborative learning system between user and machine learning (ML) model, i.e., (i) ML model provides both lesion prediction with its corresponding relevant clinical evidence; (ii) the system allows users to inspect the predicted results and be able to enhance accuracy given new samples which are roughly annotated by users, e.g., drawing a bounding box around lesion regions rather than a precise segmentation boundary.

%\hl{the system allows users to inspect predicted results and be able to enhance accuracy given new samples annotated by users which usually contain to a certain noise level. Duy: What does this mean?} \textcolor{red}{Duy: it means the user don't need to provide a precise annotation to update the system, epsecially for segmentation tasks. For e.g., just drawing a bounding box around lesion regions is enough rather than a precise boundary segmentations. This property therefore reduces labor cost.}

Our contributions are summarized as followed: 
\begin{itemize}[noitemsep]
    \item We construct a novel lesion segmentation model by formulating a domain invariant learning scheme combining a new transfer learning scheme and domain adaption concepts.
    \item We addresses both the challenges of scarcity of labeled data and domain shift. 
    \item We train our segmentation model with adversarial learning \cite{goodfellow2014generative} to enhance its robustness and accuracy.
    \item Our framework is explainable and provides diverse information to the end-users with interpretation to support its decision.
    \item Our framework allows user to provide feedback on the output and if required, re-annotate the predicted lesion features, which can be used to fine-tune the model further.
    \item Our learning method can be robust to a certain threshold, and pixel-wise correction is not required.
    \item We conduct comprehensive experiments and compare with the recent state-of-the-art methods to illustrate the effectiveness of our proposed DRG-Net.
    
\end{itemize}

The rest of the paper is organized as follows: In section \ref{sec-related-work}, we review some of the related works and some preliminaries for our method. Next, we detail our proposed DRG-Net in section \ref{sec-method}. Our DRG-Net is evaluated and compared with other SOTA existing work in section \ref{sec-experiments}. Finally, we discuss the proposed method and future 

%\hl{Need to rewrite and add more meat}
% ============================ Related Work
\section{Related Work} 
\label{sec-related-work}
\subsection{DR Grading Analysis}
DR Grading analysis aims to classify fundus images into different DR severity classes and extract the relevant lesion regions. Most of early works focus on classification task only, i.e. classifying DR into multiple classes of grading \cite{gulshan2016development,sankar2016earliest,alban2016automated}. However, such existing approaches treat the problem of classification task as a \textit{black box} and did not consider the fine-grained DR-related lesion information in the learning process \cite{nguyen2021self,zhou2020benchmark}. Some recent approaches attempt to integrate lesion information to improve the grading classification. For instance, \citet{yang2017lesion} propose two-stages Convolutional Neural Networks (CNNs), which contains two CNNs: the former plays the role of local network and aims to learn a weighted attention lesion map; the later plays the role of global network and exploits image features in holistic level for DR grading by utilizing the former's outcomes. \citet{lin2018framework} propose a similar framework to automatically detect missing lesion features and integrate with global image features using a classification network for grading prediction.

However, the aforementioned approaches are designed in a cascade framework i.e., extract lesion information first and then classify DR Grading. With such network design, both tasks are optimized independently while they have a strongly coherency. To address such limitation, \citet{zhou2019collaborative} propose a collaborative learning pipeline to jointly learn both lesion segmentation and DR Grading classification by semi-supervised learning with an attention mechanism. Lately, \citet{yang2021robust} also introduce an attention-based learning scheme, which collaboratively learns to predict lesion maps patches and DR grading. Leveraging multi-head attention mechanism, \citet{sun2021lesion} combine self-attention and cross-attention mechanisms to direct the overall attention of a network to consider lesion region diversity and their importance separately in the final DR grading prediction. While obtaining promising performance, these methods suffer from \emph{domain shift problem} i.e. difference in data distribution between source (used to train the model) and target (encountered during inference). Furthermore, none of the exiting approaches can explain how lesion maps influences the final DR grading prediction while lesion regions do not share the same contribution to DR grade. In this paper, we propose an explainable interactive joint learning framework, which does not only results lesion localization and its corresponding DR Grading score but also provides an informative explanation of the outcomes. 

\subsection{Visual explanations of CNNs.} 
Explaining the CNNs prediction is crucial for the system acceptance in critical applications such as DR Grading. In general, generating saliency maps that highlight the image regions contributing to the final CNNs prediction is one of the most common techniques. There have been a large number of research proposed to visualize a CNNs predictions by highlighting important regions that are believed to be intuitive to end-users. %\textcolor{red}{Duy: we should list out some general directions about explainable AI. For e.g., backpropagation-based and Pertubation-based methods (See Table 1 in \cite{ras2022explainable}. Then provide some abstract descriptions on each category. We next mention that we follow the back propagation-based approahces in this paper and develop constraints based on this heatmap. In the next paragraph, we can talk about CAM and its variation as current.}

For instance in image classification, visual explanation is considered to be good if it is able to be localize the category in the image (i.e. class-discriminative). Class Activation Mapping (CAM) \cite{zhou2016learning} is considered as one of the most common technique for identifying informative areas by a CNNs-based classification. In short, CAM utilizes the last convolutional layer and combine weighted activation maps to produce explainable heatmaps. As a result, it is highly class-discriminative, however, the output heatmaps is blurry (low-resolution). As an improvement of CAM, \cite{selvaraju2017grad} propose Grad-CAM, which does not require any specific network architecture. However, gradients can be noisy and may get vanished in activation function like signmoid, ReLU. To address this issue, \cite{wang2020score} proposed Score-CAM, which gets rid of the dependence on gradients by obtaining the weight of each activation map. In Score-CAM, the final result is obtained by a linear combination of weights and activation maps. Furthermore, to address the problem of low-resolution heatmaps, \cite{jalwana2021cameras} propose CAMERAS to produce high-resolution heatmaps with multi-scale accumulation and fusion of the activation maps and backpropagated gradients to compute precise saliency maps. CAM-based explainable mechanisms have been recently utilized by many research in medical imaging. For example, \citet{nguyen2021attention} proposed a CAM-based explainable COVID-19 detection method using CT images,  \citet{Wu_2021} adopt the Grad-CAM technique to perform the joint task of classification and segmentation for COVID-19 detection. 
% \section{Transfer learning and Domain Adaptation}
In this work, we propose a constraint between heatmap regions computed by GradCam in Grading Network with multi-lesion regions extracted by Lesion generators. This step aim to generate lesion prior-guided DR grading predictions. Additionally, the proposed distilling strategy using attention mechanism enables automatically selecting major lesion parts used in specific disease grading tasks and making the model to be robust. This sets us apart with other works \cite{nguyen2021attention, Wu_2021}. In the experiments, we demonstrate that this tactic improves prediction performance and provide helpful explanations to the experts (Figures 2 and 15).

\subsection{Transfer learning and Domain Adaptation} 
Domain adaptation can be regarded as a special case kind of transfer learning, thus, we first review the definitions of each to provide the differences. In the standard transfer learning setting, there are two main concepts: domain and task. While domain means the feature space of a particular dataset and the marginal probability distribution of features, task means the label space of a dataset and an specific objective function. In general, transfer learning aims to transfer the knowledge learned from a task $\mathcal{T}_a$ on domain $\mathcal{A}$ (i.e. source) to another task $\mathcal{T}_B$ on domain $\mathcal{B}$ (i.e target). Domain adaptation, as a special case of transfer learning, it assumes that both source and target share the same domain feature spaces and tasks and whereas only marginal distributions between the source and target domains are different. In medical image analysis (MIA), domain adaptation and transfer learning have been promisingly used for dealing with limited labeled data issue. 

\noindent
\textit{Transfer Learning in MIA:}  Rather than training a network with limited training data from a target task, the network is first trained for a task with potentially larger source datasets, creating a more robust model. This pre-trained network is then down-streamed on target task. However, a large-scale analysis by \citet{cats2019}; \citet{he2020sample} suggests that transfer learning might be not better than random initialization for most medical tasks. Medical images are significantly different from the natural image (e.g. ImageNet dataset); therefore, features learned from the external data maybe not effective helpful for the target medical domain task. Furthermore, medical data often face to the problem of imbalanced data at both instance level and pixel level \cite{le2021offset}. To address those limitations, \citet{nguyen2021tatl} propose a novel transfer learning, termed \textit{Task Agnostic Transfer Learning (TATL)}, motivated by dermatologists' behavior in the skincare context. TATL is learnt by a two-stage learning step i.e., first, an attribute-agnostic network is trained, which detects all the lesion regions irrespective of their labels; then,  the knowledge from this network is transferred to a set of attribute-specific classifiers to label each particular region. However, the performance of DNN trained on a particular source domain and transferred to a different target domain (e.g., different vendor, acquisition parameters), can drop unexpectedly due to domain shift. \cite{yan2019domain}.

\noindent 
\textit{Domain Adaptation in MIA:}  
In this settings, the networks are trained with domain adaptation constraints to address the domain-shift problem. \citet{tzeng2014deep} proposed a Deep Domain Confusion (DDC) method to reduce the divergence between two distributions by minimizing the maximum mean discrepancy (MMD) loss \citep{gretton2012kernel}. MMD is a nonparametric metric and can be used to compute distances between distributions as distances between mean embeddings of features. In their method, a network is trained with data from multiple distributions using a loss function that consists of both task-specific loss and MMD loss. Some studies apply adversarial optimization to remove the domain discrepancy by incorporating generative adversarial networks (GANs). GANs consists of two networks, i.e., a generator to generate a new data from a distribution and a discriminator to evaluate the new data \cite{goodfellow2014generative}. The two networks trained using game theory theorem, i.e. minimax.
\citet{tzeng2017adversarial} combined standard adversarial loss with the task-specific classification loss to minimize domain distances. At first, a discriminative model (e.g. CNNs) is trained using labeled data from the source domain. Then, a target encoder is learned using GANs adaptation, where discriminator aims to distinguish between encoded source and target examples.  \citet{shen2018wasserstein} describe Wasserstein distance guided representation learning (WDGRL) method to reduce the domain discrepancy by minimizing Wasserstein distance for each feature block of the encoder CNN. % \hl{In this work, we ... Duy: Please introduce your work here!}

In this work, we are the first take into accounts transfer learning and domain adaptation problems in the context of DR-Grading and Segmentation. These issues in fact are typically encountered when deploying an AI-assisted system in medicine applications, thus a robust framework under circumstances is critical. To this end, we properly designed support deep networks (PD-Net, AD-Net and W-Net) that jointly learn with lesion generator S-Nets. These support network operate based on the cutting-edge domain adaptation and adversarial leaning techniques which have been shown impressing results on various applications such as object recognition, self-driving car, etc. In expeiments, we demonstrate that all of these components contribute to improve performance and robustness.

%\textcolor{orange}{Duy: should we go further on technique descriptions or only talk on the abstract level as the above paragraph?}

\subsection{Interactive Machine Learning}
Interactive Machine Learning (IML), which focuses on interaction between users and learning systems \citet{amershi2014power}, with people interactively supplying information in a more focused, frequent, and incremental way compared to traditional ML. Typically, IML is an active ML technique in which models are designed and implemented with human-in-the-loop manner. While ML algorithms usually requires a large-scale data to yield accurate result; however, such requirement is limited in some domains like clinical trials, medical analysis. A comprehensive review of IML medical analysis is extensively studied in \cite{maadi2021review}. To build such a computer-based interactive system, the initial challenge is how to to represent the semantics in a machine-readable form using medical ontologies \citet{sonntag2010pillars} so that the information is easily exchangeable between human experts and machines. RadSpeech's \textit{Mobile Dialogue System for Radiologist} \citet{sonntag2012radspeech} provides a multi-modal interaction system for radiology image annotations. It is a user-friendly semantic search interface where users can annotate medical image regions with a specific medical, structured diagnosis using speech and pointing gestures. \citet{prange2017multimodal} presents a medical decision support system inside virtual reality (VR). In this system, a doctor can visualize patients' records and clinical images, as well as therapy predictions which are computed in real-time using a pre-trained deep learning model. The aforementioned studies provide the techniques to capture user feedback (annotations on inputs) for an intelligent learning system. However, they did not consider the inclusion process of these feedbacks in an IML loop.
\citet{sonntag2020skincare} describes the functionality and interface of an interactive decision support system for differential diagnosis of malignant skin lesions, where the user can re-annotate an input that is used to fine-tune the prediction model based on the explanation. Recently, \citet{dai2021deep} presents a real-time deep learning based interactive system for diabetic retinopathy where the architecture consists of three sub-networks to perform different tasks one by one. However, the coherent relation between lesion discovery and disease classification was mostly ignored, making it hard to conclude that lesion-feature information predicted by the segmentation part has any influence on the classification result.

%\hl{In this work, we ... Duy: Please introduce your work here!}
%\textcolor{orange}{Duy: We should trim the above paragraph. It is so long now. }

In this paper, we introduce the first time a human interactive framework for the DR-Grading and Segmentation tasks. The expert user benefit from our AI-assist system with the following advantageous. First, we provide an explainable predictions by correlation between heatmap of the classification network  with medical lesion regions. This helps the doctor inspect predictions generated by system by bridging with their prior knowledge. Second, due to transfer learning and domain adaptation, we permit a fast deployment our system on a new domain (new hospital, new data, etc.) while not demanding large annotations to pre-train the system (only classification annotation is required, don't need segmentation in beginning), leading to saving labour costs.  Finally, the model is designed to learn from user's feedback in weak and noise annotation forms (bounding boxes rather than precise segmentation masks, noise in new data can be damped through constraints among classification, segmentation, and attention loss functions).

\section{DRG-Net: Overview and Problem Setup} 
\subsection{Overview}
\begin{figure*}[ht]
 \centering
\includegraphics[width=1.8\columnwidth]{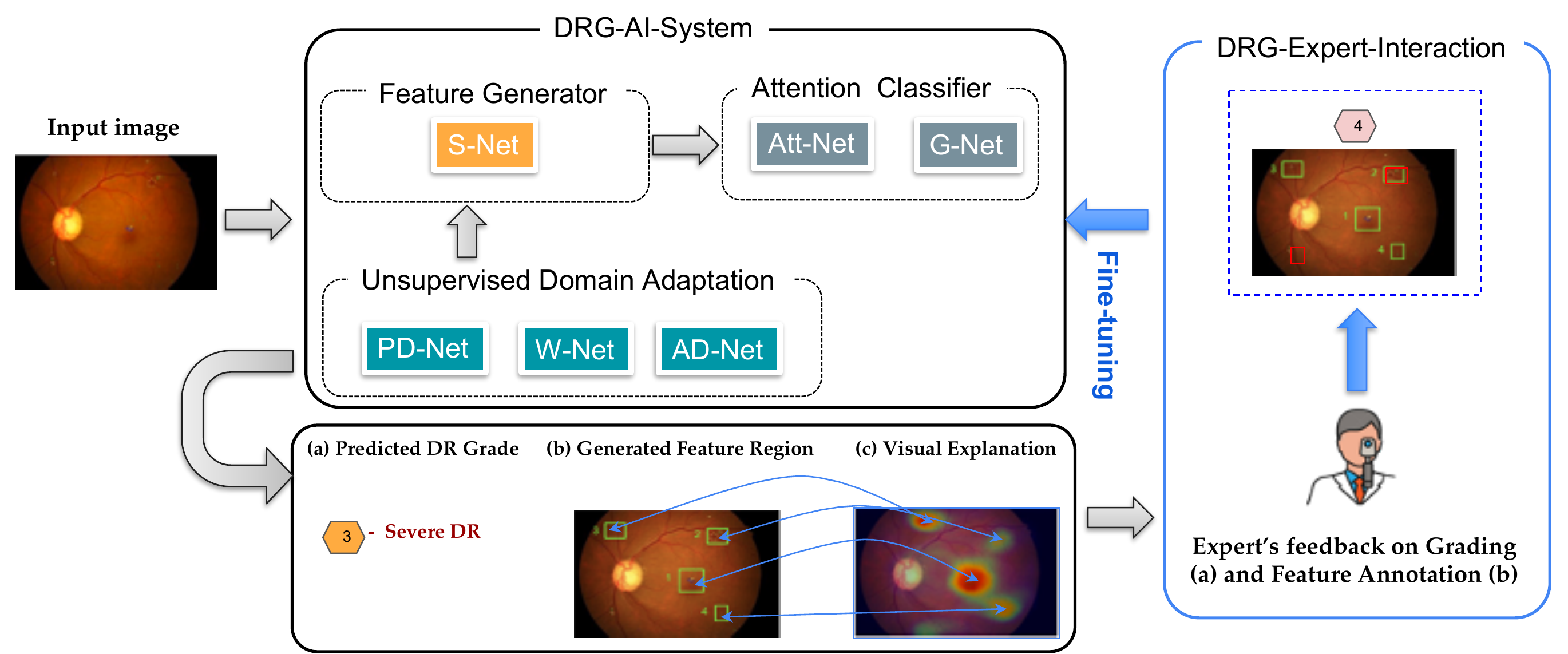}\\
    \caption{A high-level overview of our proposed DRG-Net system that contains two main module of DRG-AI-System and DRG-Expert-Interaction. DRG-AI-System: Given a retinal image, our deep learning models will simultaneously generate three outcomes prediction (DR grade, lesion region, visual explanation). DRG-Expert-Interaction: By using an Intelligent User Interface (IUI), ophthalmologists can observe the system's predictions and provide feedback to fine-tune the model.}
    \label{fig:high_level}
\end{figure*}

Our DRG-Net consists of two modules: (i) \emph{DRG-AI-System} to classify DR Grading, localize lesion areas and provide visual explanation and (ii) \emph{DRG-Expert-Interaction} to receive feedback from user-expert and improve the DRG-AI-System. The overall framework of our proposed DRG-Net is depicted in Fig. \ref{fig:high_level}. Given a retina image input, our first module DRG-AI-System will simultaneously returns three outputs: (a) DR grade severity on a scale of 0 to 4 by \textit{Attention-based Classifier}, comprising Att-Net and G-Net, (b) lesion areas by \textit{Feature generator}, comprising S-Nets and support systems (PD-Net, AD-Net, W-Net), (c) and a correlation among visual explanation, lesion regions and grading prediction. To address real-world challenges of DRG analysis including scarcity labeled data, small lesion localization, and domain shift, we propose to design DRG-AI-System in an unsupervised domain adaptation (UDA) manner, where the system was trained on source data with annotation and test on target data without label. In the second module of DRG-Expert-Interaction, the expert can provide feedback on the grading and lesion information, which is then utilized to fine-tune the DRG-AI-System network.

\subsection{Problem Setup}
In DRG-AI-System, we consider two correlated tasks at both image-level and pixel-level. The former task aims to classify DR Grading (i.e., associate an input image with a DR Grading class), whereas the later one aims to localize lesion areas (i.e., partition an input image into various  lesion areas).

Let $\mathbf{x}_s \in \mathcal{X}_s$, $\mathbf{\hat{t}}_s \in \mathcal{T}_s$, and $\mathbf{\hat{y}}_s \in \mathcal{Y}_s$ be an input image and its corresponding image-level and pixel-level label in the source domain drawn from the source distribution $p_s$. Similarly, we denote
$\mathbf{x}_t \in \mathcal{X}_t$, $\mathbf{\hat{t}}_t \in \mathcal{T}_t$ and $\mathbf{\hat{y}}_t \in \mathcal{Y}_t$ be the input image and its corresponding image-level and pixel-level label in the target domain drawn from the target distribution $p_t$.  As for image-level label, there are $C$ classes of No DR (0), Mild (1), Moderate (2), Severe (3), Proliferative (4). As for pixel-level label, there are $L$ classes of Microaneurysms (MA), Hard Exudates (HE), Soft Exudates (SE), Hermorrhages (EX).

In our domain adaptation setting, the ground-truth pixel-level label $\mathbf{\hat{y}}_t$ of image $\mathbf{x}_t$ is not available. 

The DRG-AI-System is trained on source data and tested on target data. Given a target retinal image $\textbf{x}_t \in \mathcal{X}_t$, the output from DRG-AI-System is a tuple of grading level, localizing lesion areas and explainable visual representation.

The following Table \ref{tb:symbols} summarizes and briefly describes important symbols that are used throughout this manuscript:

\begin{table*}[]
\caption{Summation of symbols and their brief description.}
\begin{tabular}{l|l||l|l}
\hline
Variables & Description & Network & Description \\ \hline
$L$         &   Number of classes of lesions         & S-Net = $\{S^e, S^d\}$  &   Segmentation network, enoder $S^e$ and decoder $S^d$\\ 
$\mathcal{X}_s$ & source domain &  $S_l = \{S_l^e, S_l^e\}$ & \shortstack{Binary segmentation network for a lesion $l$}\\ 
$\mathcal{X}_t$ & target domain &  PD-Net & Patch Discriminator \\
$\mathbf{x}_s \in \mathcal{X}_s$    &   an image in source domain   & W-Net & Wasserstein Discriminator             \\ 
$\mathbf{x}_t \in \mathcal{X}_t$ & an image in target domain           & AD-Net & Adversarial Domain Discriminator         \\ 
$\mathbf{x}$ & an image from any domain & Att-Net & Attention Network \\
$\mathbf{\hat{t}}_s$, $\mathbf{\hat{t}}_t$    &  GT  image-level label of source/target image &   G-Net      &     Grading Network        \\
$\mathbf{\hat{y}}_s$, $\mathbf{\hat{y}}_s$     &  GT  pixel-level label of source/target image & & \\
$\mathbf{\hat{y}}$     &  predicted  pixel-level label  & & \\
$\mathbf{h}_s$, $\mathbf{h}_t$     &  encoder feature map of a source/target image  & & \\
$\mathbf{g}^i$ &  Conv. block & & \\
$\mathbf{o}^i$ &  output of each Conv. block $\mathbf{g}^i$ & & \\
$h$, $w$ & height and width of an image & & \\  
$C$ & number of grading labels & & \\ \hline
\end{tabular}
\label{tb:symbols}
\end{table*}

Each module will be detailed in the following sections.

\section{DRG-Net: DRG-AI-System} 
\label{sec-method}

% \subsection{Deep Network Architectures for Lesion Attributes Segmentation and DR Grading Prediction}
\label{sec:feature_generation}

Our proposed UDA DRG-AI-System consists of three main components: The first component, Feature Generator, plays the role of multi-lesion segmentation and is termed \textit{S-Net}. The second component plays the role of UDA including Patch Discriminator \textit{PD-Net}, Wasserstein Discriminator \textit{W-Net} and Adversarial Domain Discriminator \textit{AD-Net}. The second component aims to support the first component in dealing with scarcity labeled data, small objects and domain shift.  The third component, Attention-based Classifier, is designed as a combination of an Attention Network \textit{Att-Net} and a Grading Network \textit{G-Net}. 
The overall DRG-AI-System is given in Fig.\ref{fig:DRG-AI-System}. 

\subsection{Feature Generator: multi-lesion segmentation S-Net}
\label{subsec:Snet}
S-Net aims to partition an input retinal image into $L$ binary segmentation masks, each represents a lesion mask. In this paper, our proposed S-Net is based on an encoder-decoder Unet-like architecture \cite{ronneberger2015unet} consisting of encoder $S^e$ and decoder $S^d$. %We follow.... \hl{Duy: Details of S-Net will be included, TransUnet or Mask2Former?}. 

Furthermore, we focus on real-world applications where S-Net is trained on source domain $\mathcal{X}_s$ and tested on target domain $\mathcal{X}_t$. Under this setup, annotation is only available during training including image-level annotation $\mathbf{\hat{t}}_s \in \mathcal{T}_s$, and pixel-level annotation $\mathbf{\hat{y}}_s \in \mathcal{Y}_s$. Due to scarcity of training data, small lesion area and domain shift, standard Unet-based networks are limited when working on retinal images. In this work, we propose to utilize Task Agnostic Transfer Learning (TATL) framework \cite{nguyen2021tatl} to propose a novel UDA approach to alleviate the aforementioned challenges. Our proposed UDA consists of three sub-networks, i.e. Patch Discriminator PD-Net, Wasserstein Discriminator W-Net and Adversarial Domain Discriminator AD-Net, which will be detailed in Section \ref{subsec:UDA} whereas we only focus describing S-Net in this section.

In the TATL settings, S-Net is first pre-trained at the pretext task on a multi-lesion region first. Then, it is then fine-tuned on the downstream task for segmenting an individual lesion region. Each particular lesion $l$ is segmented by a network $S_l$, all $S_l$, $l \in [1, L]$ shares a similar network architecture as S-Net. For a given image $\mathbf{x}_s \in \mathcal{X}_s$, for each network $S_l$, we obtain a binary segmentation mask $S_l(\mathbf{x}_s),\ (l \in L)$, where $L = \{\mathrm{MA},\,\mathrm{HE},\,\mathrm{SE},\,\mathrm{EX}\}$. 
% An illustration of utilizing TATL is given in Fig. \ref{fig:tatl}.
%  \begin{figure*}[ht]
%  \centering
% \includegraphics[width=1.7\columnwidth]{images/method/tatl_dr.pdf}\\
%     \caption{Task agnostic transfer learning (TATL) procedure used in the pre-training step in lesion attribute generation. \textit{Pretext Task}: an U-shaped model is trained to recognise all the regions containing any lesion attributes through an \textit{attribute agnostic mask}. \textit{Downstream Task}: Trained parameters from the \textit{Pretext Task} are transferred to the downstream tasks of locating and identifying each of the lesion attributes independently. The image is adopted from \cite{nguyen2021tatl} and modified as per fundus image lesion feature generation task.}
%     \label{fig:tatl}
% \end{figure*}
The segmentation network is trained by weighted binary cross-entropy loss as follows:

\begin{align}
    \mathcal{L}_{seg} = {\text{min}} \frac{1}{\lvert \mathcal{X}_s \rvert} \sum_{(\mathbf{x}_s, \hat{\mathbf{y}}_s) \in \mathcal{X}_s} \sum_{l \in L} \mathcal{L}_{wbce} (S^{l}(\mathbf{x}_s), \hat{\mathbf{y}}_{{s}_l})
    \label{eq:seg}
\end{align}
where $\hat{\mathbf{y}}_{{s}_l}$ is pixel-level annotation of lesion $l$. $\mathcal{L}_{wbce}$ is the weighted binary cross-entropy loss
\begin{align}
    \mathcal{L}_{wbce}(\hat{y}, {\mathbf{y}}) = -(\beta \cdot \hat{\mathbf{y}} \log ({\mathbf{y}}) + (1 - \hat{\mathbf{y}}) \log(1 - \mathbf{y}),
    \label{eq:wbce}
\end{align}
where $\hat{\mathbf{y}}$ is the ground-truth binary lesion mask and ${\mathbf{y}}$ is the model prediction, $\beta$ is the class balancing weight. The loss $\mathcal{L}_{wbce}$ in general can be replaced with other segmentation losses \cite{jadon2020survey}.

So far, we have trained the segmentation model $S_l$ in the source domain $\mathcal{X}_s$ using eq.\ref{eq:seg} where annotation is available. However, it is unfit to target domain where we do not have annotation. We propose to constrain the embedding space of the source and target domains are close to each other by minimizing Wasserstein distance \cite{wasser_gan,liu2019wasserstein} via \textit{W-Net}. Furthermore, to improve semantic segmentation accuracy on tiny objects, we follow prior studies \citep{hung2018adversarial} and incorporate a discriminate network for each lesion segmentation model $S_l$. We propose Patch Discriminator \textit{PD-Net}. Furthermore, to exploit the structural consistency between the source and target domain, we proposed \textit{AD-Net} to directly minimize an uncertainty prediction in the target domain. Details of PD-Net, W-Net and AD-Net are in Section \ref{subsec:UDA}

\subsection{Unsupervised Domain Adaptation}
\label{subsec:UDA}
\subsubsection{Patch Discriminator PD-Net}
\label{subsec:GAN-source}
\begin{figure}[ht]
 \centering
\includegraphics[width=1.0\columnwidth]{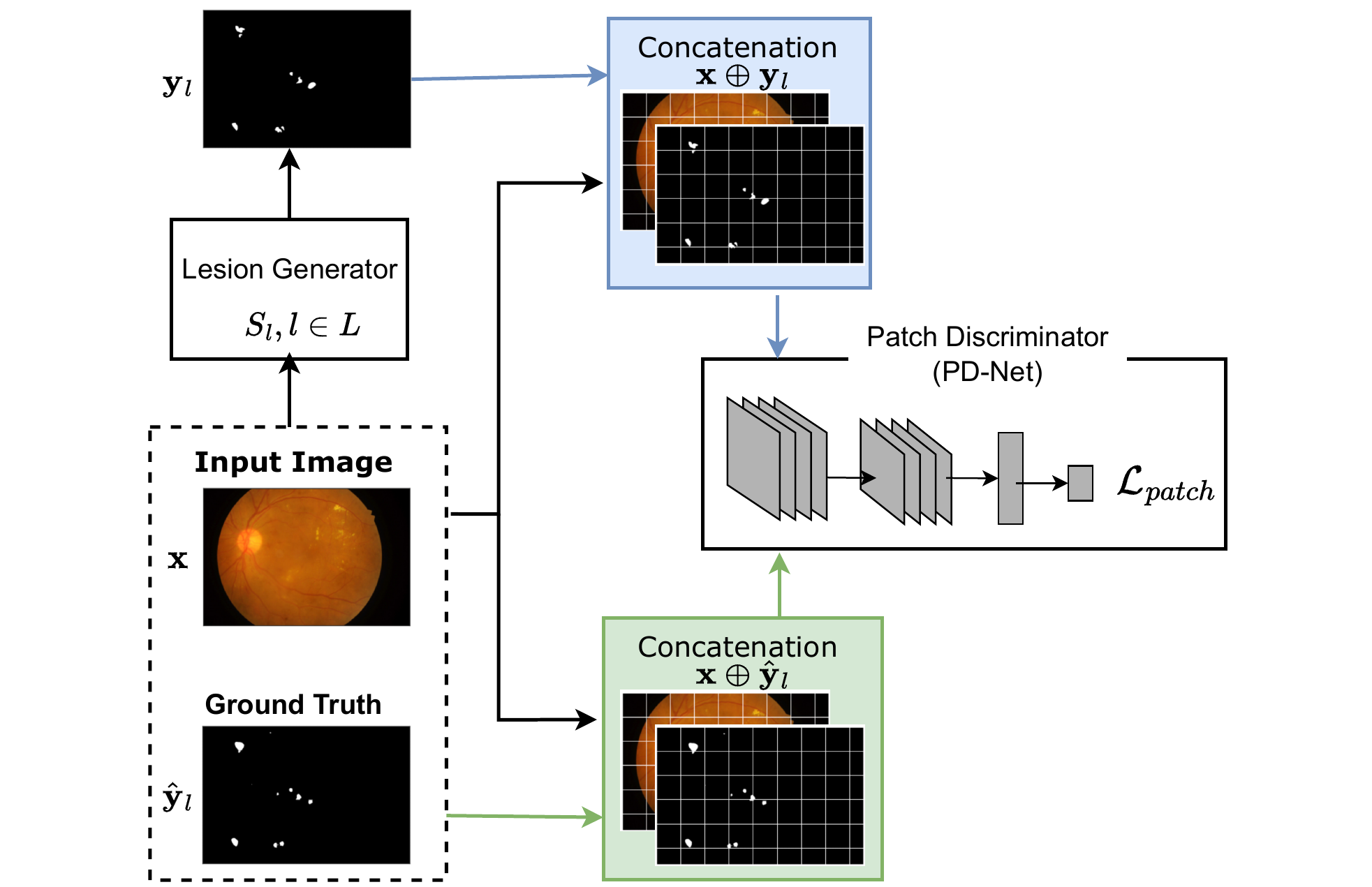}
    \caption{Our workflow for the supervised lesion segmentation using adversarial learning. Output from the segmentation network $S_l$ is concatenated with the input image and split into patches. Then, the patch discriminator \textit{PD-Net} computes the adversarial loss by predicting whether each patch is real or fake.}
    \label{fig:patch}
\end{figure}

To improve semantic segmentation accuracy on tiny objects, we follow prior studies \citep{hung2018adversarial} and incorporate a discriminate network for each lesion segmentation model $S_l$. We propose Patch Discriminator \textit{PD-Net}, a conditional GAN \cite{mirza2014conditional}, which aims to distinguish \textit{real} samples from the generated ones. The architecture of \textit{PD-Net} is similar to \cite{Xiao_2019}, which is formed based on the ideas of PatchGAN \cite{isola2018imagetoimage} as illustrated in Fig. \ref{fig:patch} .

In particular, an input image is split into $16 \times 16$ smaller patches, and each of these patches is applied with the cross-entropy loss to decide whether that patch is fake or real. The input for the PD-Net is the concatenation of the original image patch with its corresponding lesion map predicted from $S_l$ and actual ground truths. Thus, the discriminator PD-Net learns the joint distribution of both images and the lesion map, conditioned on the input data. In other words, the PD-Net will force the output of $S_l$ to look 'real' as the ground-truth data as much as possible given the input image $\textbf{x}_s \in \mathcal{X}_s$. The objective loss function for this formulation can be defined as:
% For class label 1 and 0 for ground-truth and prediction lesion map respectively, the  patch-based adversarial loss $\mathcal{L}_{\mathrm{pathch}}$ can be defined in our case by the following:
\begin{align}
  \mathcal{L}_{\mathrm{patch}} =   \underset{\theta_{S_{l}}}{\text{min}}\, \underset{\theta_{\mathrm{pd}}}{\text{max}} \frac{1}{\lvert \mathcal{X}_s \rvert} \sum_{(\mathbf{x}, \hat{\mathbf{y}}_{l}) \in \mathcal{X}_s} \left[ \mathrm{log}(\mathrm{PD}(\mathbf{x} \oplus \hat{\mathbf{y}}_l)) \right.\\
  \left.+  \log(1 - (\mathrm{PD}(\mathbf{x} \oplus {\mathbf{y}_l})))\right]
    \label{eq:cgan}
\end{align}

where $\theta_{S_{l}}$ and $\theta_{\mathrm{pd}}{\text{max}}$ are parameters of binary segmentation network $S_l$ and PD-Net. $\oplus$ is the concatenation operator. ${z_l}$ is output of  binary segmentation network $S_l$ i.e., $z_l = S_{l}(\mathbf{x})$ and $\hat{\mathbf{y}}_l$ is groundtruth corresponding to lesion $l$. Combining this adversarial objective with Eq.\,(\ref{eq:seg}), we derive an optimization problem for the segmentation model $S_l$ using the source domain $\mathcal{X}_s$ as:
\begin{align}
  \mathcal{L}_{\mathrm{source}} =   \mathcal{L}_{seg} + \lambda_{p}\,\mathcal{L}_{\mathrm{patch}.}
    \label{eq:supervised_joint}
\end{align}
where $\lambda_{p}$ is a parameter controls the contribution of $\mathcal{L}_{patch}$.

\subsubsection{Wasserstein Discriminator W-Net}
\label{sub2sec:w-net}
\begin{figure}[ht]
 \centering
\includegraphics[width=1.0\columnwidth]{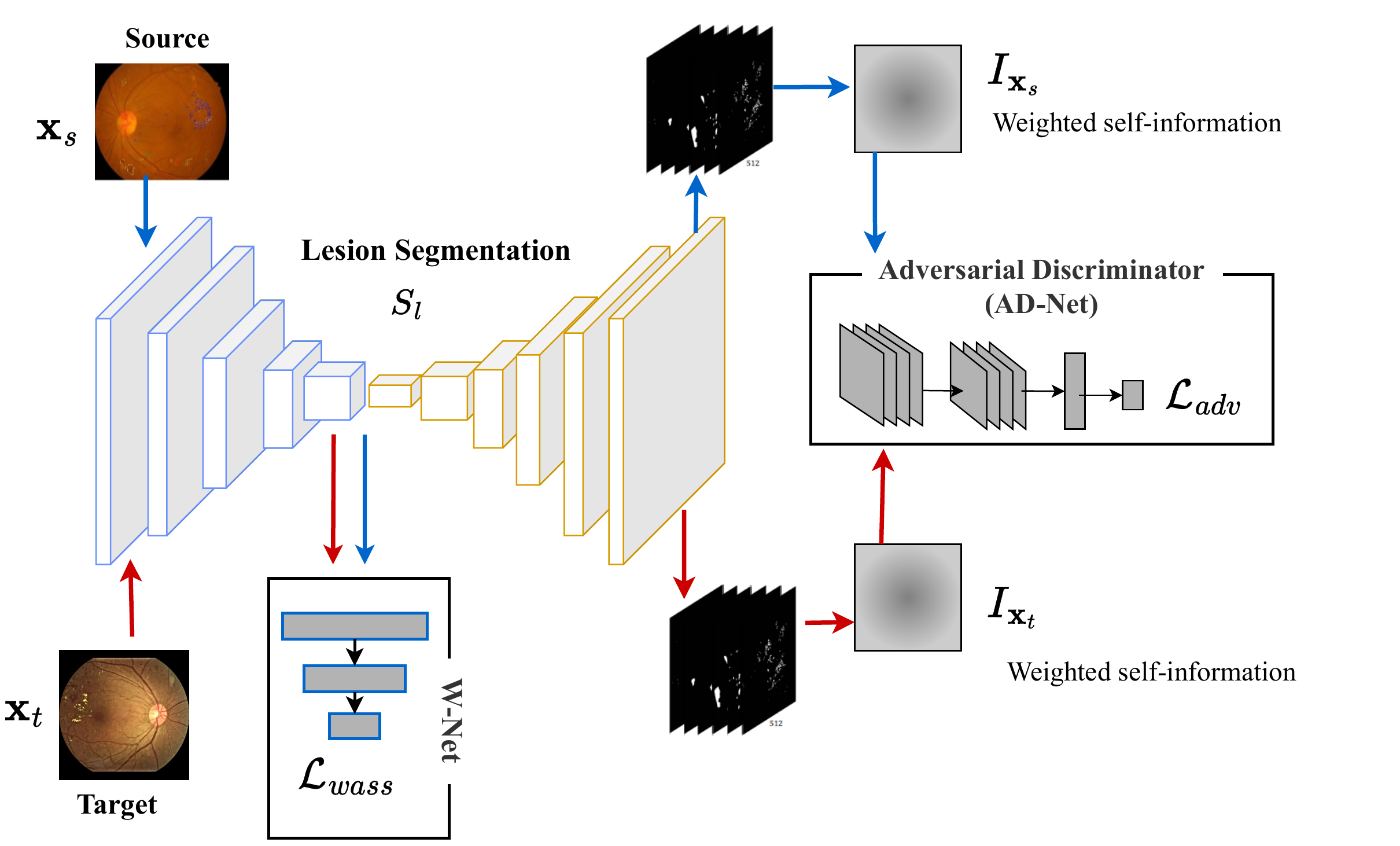}\\
    \caption{Domain Adaptation Components in our method with \textit{W-Net} and \textit{AD-Net}. The \textit{W-Net} is designed to minimize the Wasserstein Distance Minimization between two feature embedding spaces in the source and target domain. The \textit{AD-Net} is employed as the discriminate network to distinguish entropy distributions from two domains.}
    \label{fig:domain-adaptation}
\end{figure}

\begin{figure*}[ht]
 \centering
\includegraphics[width=2.0\columnwidth]{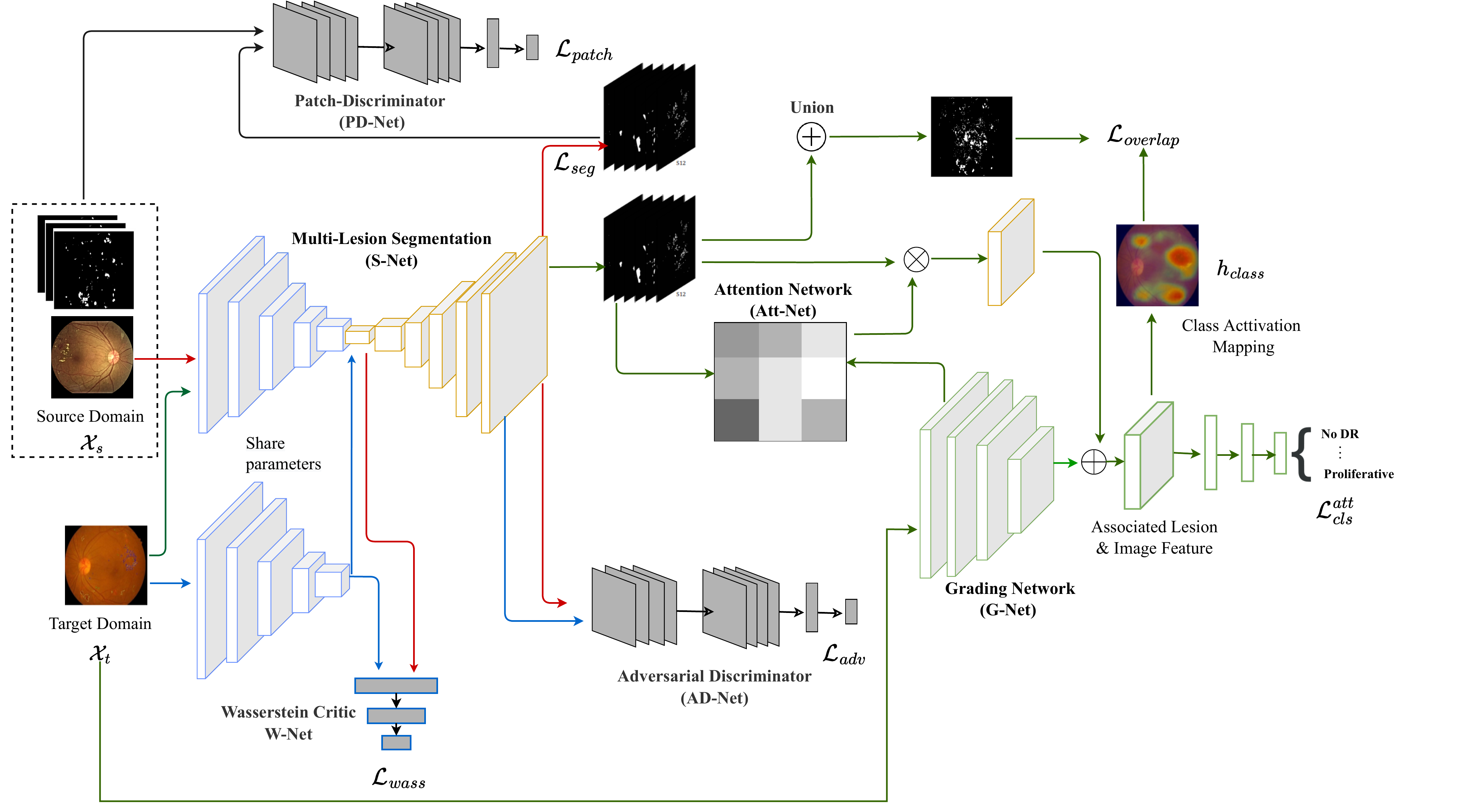}\\
\caption{Detailed architectures and pipeline of the proposed \emph{DRG-AI-System}. It consists of five sub-networks: \textit{S-Net}, \textit{PD-Net}, \textit{AD-Net}, \textit{W-Net}, \textit{Att-Net}, and \textit{G-Net}. These networks are learned in two phases. The first stage aims to train the multi-lesion segmentation networks \textit{S-Net}. To overcome the lack of training data in a source domain and domain adaptation issues in a target domain, \textit{PD-Net}, \textit{W-Net}, and \textit{AD-Net} are jointly integrated.  The second stage optimizes the grading network \textit{G-Net} using lesion information obtained from \textit{S-Net}, where the attention network \textit{Att-Net} automatically decides the most influential parts of a given image.}
    \label{fig:DRG-AI-System}
\end{figure*}

For each lesion segmentation network $S_l$, we aim to minimize discrepancies between different domain's feature embedding estimated at the encoder layers $S_{l}^{e}$ of $S_l$. To this end, we introduce the domain critic \textit{W-Net} parameterized by $\theta_w$, whose goal is to estimate the Wasserstein distance \cite{wasser_gan} between the source and target distribution in the feature representation space (Fig. \ref{fig:domain-adaptation}). Given an encoder feature representation $\mathbf{h} = S_{l}^{e}(\mathbf)$ for an image $\mathbf{x}$ from any domain, we define:
\begin{align}
    \mathbf{h}_s &= S_{l}^{e}(\mathbf{x}_s),\ \mathbf{x}_s \in \mathcal{X}_s,\   \\
    \mathbf{h}_t &= S_{l}^{e}(\mathbf{x}_t),\ \mathbf{x}_t \in \mathcal{X}_t.
\end{align}
and the domain critic function $W(.) \colon \mathbb{R}^d \rightarrow \mathbb{R}$ which maps a feature representation to a real number. As proposed by \cite{shen2018wasserstein}, if the parameterized family of domain critic function $W(.)$ are all 1-Lipschitz, for the source and target distributions $\mathcal{X}_s$ and $\mathcal{X}_t$, an empirical Wasserstein distance can be approximated by \textit{maximising domain critic loss} $\mathcal{L}_{wd}$ with respect to $\theta_w$:
\begin{align}
    \mathcal{L}_{wd}(\mathcal{X}_s, \mathcal{X}_t) = \frac{1}{\lvert \mathcal{X}_s \rvert} \sum_{\mathbf{x}_s \in \mathcal{X}_s} W(\mathbf{h}_s) - \frac{1}{\lvert \mathcal{X}_t \rvert} \sum_{\mathbf{x}_t \in \mathcal{X}_t} W(\mathbf{h}_t).
    \label{eq:wasser_loss}
\end{align}
% \underset{\theta_{w}}{\text{max}}\, 
% \underset{\theta_{w}}{\text{min}}\,
To make the training progress to be stable, we further enforce parameters $\theta_w$ to \textit{minimize the Lipschitz constraint} using gradient penalty $\mathcal{L}_{grad}$ proposed by \cite{Improved-wgan} of the domain critic as:
\begin{align}
    \mathcal{L}_{grad}(\hat{\mathbf{h}}) =  \left( 
    \left\Vert \bigtriangledown_{\hat{\mathbf{h}}} W(\hat{\mathbf{h}})  \right\Vert_{2} - 1\right)^2, 
    \label{eq:wasser_grad}
\end{align}
where $\hat{\mathbf{h}} = \{\mathbf{h}_s, \mathbf{h}_t, \mathbf{h}\}$ is the feature representations at which to penalize the gradients and it is defined by the source and target representations $\mathbf{h}_s, \mathbf{h}_t$ as well as at the random points $\mathbf{h}$ along the straight line between source and target representation pairs \cite{shen2018wasserstein,Improved-wgan}. 

Combining Eq.\,(\ref{eq:wasser_loss}) and Eq.\,(\ref{eq:wasser_grad}), the optimization problem for the \textit{W-Net} and the lesion attribute segmenter  $S_{l}\ (l \in L)$, which minimizes discrepancy between two domains $\mathcal{X}_s$ and $\mathcal{X}_t$ in terms of Wasserstein distance, is defined as:
% maximise the critic score thus minimizing $\theta_e$ by enforcing encoder to minimize domain discrepancy:
\begin{align}
\mathcal{L}_{wass} = 
    \underset{\theta_{S_{l}}}{\text{min}}\, \underset{\theta_w}{\text{max}} \{\mathcal{L}_{wd} - \eta \mathcal{L}_{grad} \},
    \label{eq:wasser_total}
\end{align}
where $\eta$ is the regularization coefficient. 

\subsubsection{Adversarial Domain Discriminator AD-Net}
\label{sub2sec:AD-net}
In order to exploit the structural consistency between the source and target domain, we employ an entropy loss $\mathcal{L}_{ent}$ to directly minimize an uncertainty prediction \cite{advent,ma2021adversarial} in the target domain of $S_{l}$. Given an input image in the target domain $\mathbf{x}_{t} \in \mathcal{X}_{t}$, we compose the Shannon Entropy map \cite{entropy} $\textbf{\textit{E}}_{\mathbf{x}_{t}} \in [0, 1]^{h \times w}$:

%\hl{Duy: Please check this equation. It is strange when sum across all channels. C is number of classes in classification whereas L is class in segmentatio, thus C or L}

\begin{align}
\textbf{\textit{E}}_{\mathbf{x}_{t}}^{(h,w)} = \frac{-1}{\log (C)} \sum_{i=1}^C P_{\mathbf{x}_{t}}^{(h,w,i)} \log P_{\mathbf{x}_{t}}^{(h,w,i)}, 
    \label{eq:entropy_map}
\end{align}
at each pixel $(h,w)$. Here, $C$ being the number of output class, i.e., $C = 2$ in our setting with binary segmentation for each lesion attribute $l \in L$, and $P_{\mathbf{x}_{t}}^{(h,w,i)}$ is the pixel-wise predicted class score estimated from  $S_{l}(\mathbf{x}_{t})$. 

%\textcolor{red}{Duy-answer: The Eq.(11) is correct. This indeed computes the entropy at each pixel. For e.g, at pixel (h,w) in image $x_t$, we have probability scores (0.7, 0.3) where 0.7 is for object and 0.3 is for the foreground. The entropy in this case is $\frac{-1}{\log(2)}[0.7*\log(0.7) + 0.3*\log(0.3)]$ (Shannon Entropy)}.

We now define an entropy loss $\mathcal{L}_{ent}$ which is the sum of all pixel-wise normalized entropy:
\begin{align}
    \mathcal{L}_{ent}(\mathbf{x}_t) = \sum_{h,w} \textbf{\textit{E}}_{\mathbf{x}_{t}}^{(h,w)}
    \label{eq:entropy_loss}
\end{align}

By combining Eq.(\ref{eq:seg}) and Eq. (\ref{eq:entropy_loss}), we can jointly optimize the supervised segmentation loss with the samples in the source domain $\mathcal{X}_s$ and an unsupervised entropy loss on the target domain $\mathcal{X}_t$ by:
\begin{align}
     \underset{\theta_{S_{l}}}{\text{min}} \frac{1}{\lvert \mathcal{X}_s \rvert} \sum_{(\mathbf{x}, \mathbf{z}_{l}) \in \mathcal{X}_s} \mathcal{L}_{wbce} (S_{l}(\mathbf{x}), \mathbf{z}_{l}) + \frac{\lambda_{ent}}{\lvert \mathcal{X}^t \rvert} \sum_{\mathbf{x}_t \in \mathcal{X}_t} \mathcal{L}_{ent}(\mathbf{x}_t),
    \label{eq:joint_seg_loss_1}
\end{align}
% problem but combining equation \ref{eq:seg} and \ref{eq:entropy_loss} where $\mathcal{L}{seg}$ is the supervised segmentation loss on the samples from source domain and $\mathcal{L}_{ent}$ is the unsupervised entropy loss on the target samples:
% \begin{align}
%      \underset{\theta_{S_{l}}}{\text{min}} \frac{1}{\lvert X_s \rvert} \sum_{(x, z_{l}) \in X_s} \mathcal{L}_{wbce} (S_{l}(x), z_{l}) + \frac{\lambda_{ent}}{\lvert X^t \rvert} \sum_{x_t \in X_t} \mathcal{L}_{ent}(x_t),
%     \label{eq:joint_seg_loss_1}
% \end{align}
with $\lambda_{ent}$ is the weight factor for the entropy part $\mathcal{L}_{ent}$. 

Training the lesion segmentation model $S_{l}$ with the joint loss function in Eq.\,(\ref{eq:joint_seg_loss_1}) has shown improvement in experiments. However, merely minimization this does not entirely capture the structural dependencies between the local semantics. In other directions, authors in \cite{learning_to_adapt} have argued that adaptation to the structural output space is favorable for unsupervised domain adaptation in semantic segmentation tasks. Similarly, \citet{shen2018wasserstein} have also shown that adaptation in the latent space can enhance the model generalization ability. These premises are based on the fact that the source and target domain usually share strong similarities in the semantic layout. To exploit such observations, we incorporate an adversarial training framework to implicitly guide the target domain's entropy distributions to be similar to the source ones \cite{advent,ma2021adversarial}. Our motivation is based on the fact that a trained model naturally produces a high confidence score for one target class and low for the rest on source-like images. Therefore, entropy for the source-like images will be low, and that of target images will be higher.
% similar to \cite{advent,ma2021adversarial}, 
% Instead of direct entropy minimization, similar to \cite{advent}, we incorporate an adversarial training framework to indirectly minimize the entropy to have target domain's entropy distribution similar to the source. 

% It is motivated by the fact that a trained model naturally produces a high confidence score for one target class and low for the rest on source-like images. Therefore, entropy for the source-like images will be low and that of target images will be higher.
Using the equation (\ref{eq:entropy_map}), we define a \textit{weighted self-information map} $I_x$ for any input image $\mathbf{x}$ composed  of pixel-level vectors at each pixel $(h,w)$ as:
\begin{align}
    I_{x}^{(h,w)} = - P_{x}^{(h,w)} \circ \log P_{x}{(h,w)}
    \label{eq:self-information}
\end{align}
%\hl{Duy: it is not clear is P is a vector length C. Is so, how to compute the Eq. 14}
where $\circ$ stands for hadamard product, $P_{x}^{(h,w)} = \left[P_{x}^{(h,w,i)}\right]_{i=1}^{C}$ is the probability score for $C$ classes at $(h,w)$ estimated by $S_{l}(x)$.

Given this, we define $I_{\mathbf{x}_s},\,I_{\mathbf{x}_t}$ as weighted self-information maps for the source and target domain $\mathcal{X}_s, \mathcal{X}_t$ respectively computed by:
\begin{align}
    I_{\mathbf{x}_s}^{(h,w)} &= - P_{\mathbf{x}_s}^{(h,w)} \circ \log P_{\mathbf{x}_s}{(h,w)} \\
    I_{\mathbf{x}_t}^{(h,w)} &= - P_{\mathbf{x}_t}^{(h,w)} \circ \log P_{\mathbf{x}_t}{(h,w)}
    \label{eq:self_information_formula}
\end{align}
% Given pixel-wise predicted class score $P_{x}^{(h,w,c)}$ for $X$, the self-information is defined as $- \log P_x^{(h,w,c)}$ \cite{Tribus1961ThermostaticsAT}. $I_x$ is composed of pixel-level vectors as hadamard product $\circ$ of class score and self-information:

% \begin{align}
%     I_{x}^{(h,w)} = - P_{x}^{(h,w)} \circ log P_{x}{(h,w)}
%     \label{eq:self-information}
% \end{align}

We now formulate the adversarial network \textit{AD-Net} with parameters $\theta_{ad}$ as convolutational networks. This network takes $I_{x}$ as an input and produces domain classification as an output with class label 1/0 for the source/target domain respectively (right side in Fig. \ref{fig:domain-adaptation}). Note that, \textit{AD-Net} architecture and its parameters are different from the discriminator network \textit{PD-Net} that we have used to optimised $S_{l}$ in Eq.\,(\ref{eq:supervised_joint}). Similar to the learning procedure of the original GAN method \cite{goodfellow2014generative}, the discriminator \textit{AD-Net} is trained to discriminate outputs coming from source and target images and simultaneously, the lesion  network $S_{l}$ is trained to fool the discriminator \textit{AD-Net}. The optimization objective of the discriminator is:
\begin{align}
    \mathcal{L}_{adv} = 
    \underset{\theta_{S_{l}}}{\text{min}}\ 
    \underset{\theta_{ad}}{\text{max}}\  \frac{1}{\lvert \mathcal{X}_s \rvert} \sum_{\mathbf{x}_s \in \mathcal{X}_s} \log(AD(I_{\mathbf{x}_{s}})) \\+ \frac{1}{\lvert \mathcal{X}_t \rvert} \sum_{\mathbf{x}_t \in \mathcal{X}_t} \log( 1 - AD(I_{\mathbf{x}_{t}})))
    \label{eq:ad_net}
\end{align}
% where $\mathcal{L}_{bce}$ is the cross-entropy classification loss. The adversarial objective for training the lesion segmentation network will be:

% \begin{align}
%     \underset{\theta_{S}}{\text{min}} \frac{1}{\lvert X_t \rvert} \sum_{x_t \in X_t} \mathcal{L}_{adv}(I_{x_{t}}, 1)
%     \label{eq:ad_net_adv}
% \end{align}
In summary, by jointly optimizing equations \,(\ref{eq:supervised_joint}),\,(\ref{eq:wasser_total}), and (\ref{eq:ad_net}), we end up with a total loss function:
\begin{align}
    \mathcal{L}_{total} = \mathcal{L}_{seg} + \lambda_{p}\,\mathcal{L}_{\mathrm{patch}} + \lambda_{w}\mathcal{L}_{wass} + \lambda_{adv}\mathcal{L}_{adv}.
    \label{eq:final_segmentation_formula}
\end{align}
where, $\lambda_p,\,\lambda_{w}$ and $\lambda_{adv}$ are the weighting factors for the patch-based adversarial learning, Wassertein term, and entropy-based adversarial term respectively.

\subsection{Attention-based Classifier}
To this end, using lesion segmentation models $S_{l}$ from the previous step, we generate $L$ distinctive lesion maps ${\mathbf{z}}_{{l=1}_l}$ for each input image $\mathcal{x}$. The attention module \textit{Attn-Net} then uses this preliminary information to select the most relevant parts contributing toward the better performance of the disease classification network \textit{G-Net}. 

\begin{figure*}[ht]
 \centering
\includegraphics[width=1.5\columnwidth]{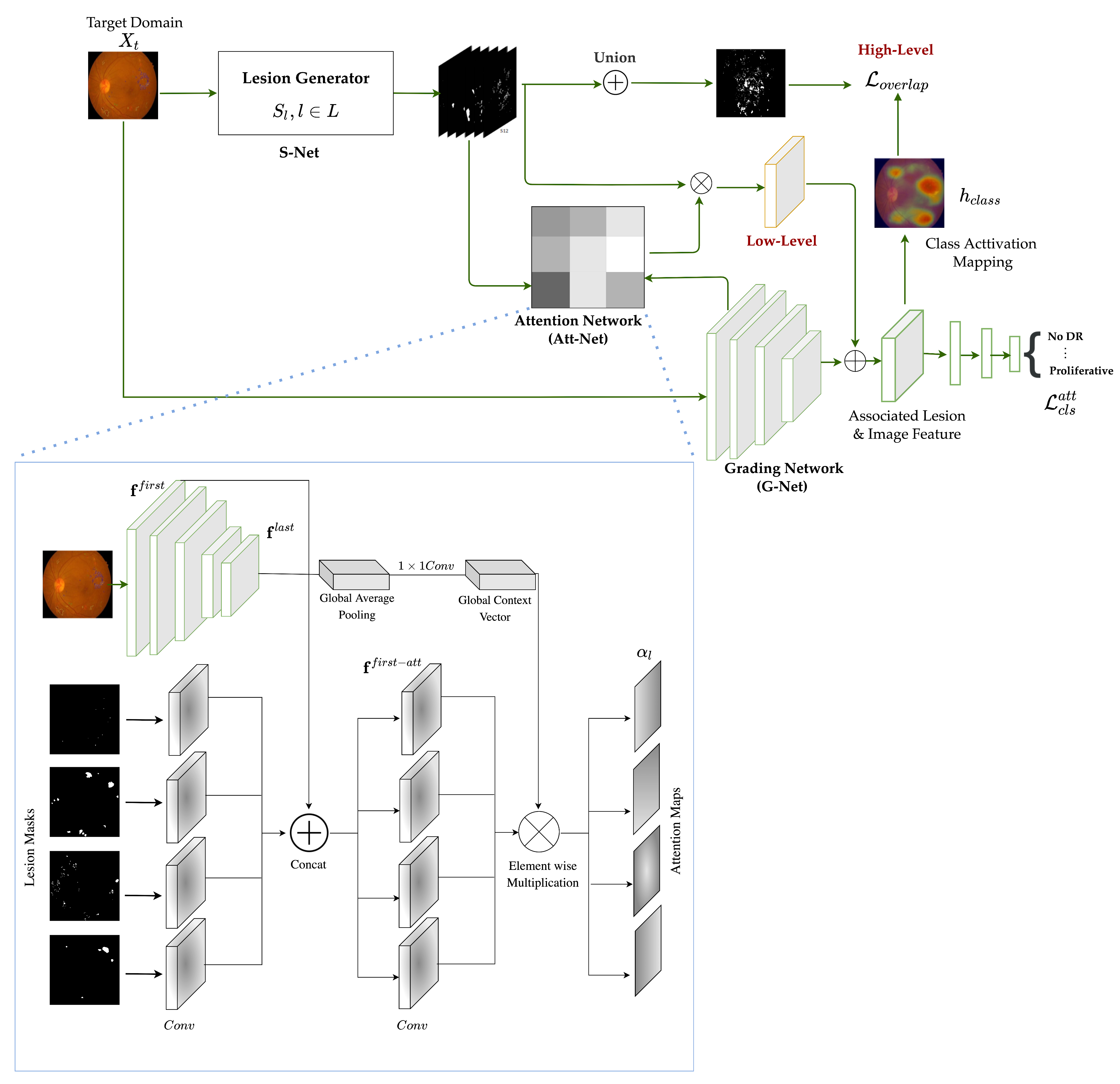}\\
    \caption{An overview of our attention mechanisms jointly learning with the DR grading network at both low-level and high-level concepts. In the blue box, internal architecture of low-level attention mechanism is illustrated in detail.}
    \label{fig:overview-attention}
\end{figure*}

This work investigates two different jointly learning frameworks between \textit{Att-Net} and \textit{G-Net} for network architectures based on CNN (section \ref{subsec:lesion-cnn}) and Transformer (section \ref{subsec:lesion-trans}). Furthermore, besides integrating lesion information at a feature level through attention gates, denoted as \textit{low-level concepts} (section \ref{subsec:lesion-cnn}), we also formulate a novel overlapping constraint among heatmap regions of \textit{G-Net} and segmented multi-lesion regions obtained from \textit{S-Net}, denoted as \textit{high-level concepts} (section \ref{subsec:lesion-high-level}). Fig. \ref{fig:overview-attention} gives an illustration for these attention concepts. In our setting, we develop these constraints for both CNN-based and Transformer-based architectures and discover that they contribute in improving accuracy for several baselines by a large margin (tables \ref{table:grading_vit}, \ref{table:grading_cnn}). Last but not least, these strategies enable an explainable visualization of DR grading predictions when ophthalmologists can observe the correlation of high-responding areas in trained networks with medical priors represented as lesion regions.

\subsubsection{Attention Lesion Regions at Low-level Concepts: Attention Network Att-Net and Grading Network G-Net}
\label{subsec:lesion-cnn}
% Most attention mechanism for vision related tasks are designed using high-level features. \citet{visual_attention} has proposed a guided attention inference network (GAIN) which utilizes the class activation map (CAM) of the convolution block of the neural network for inference.
Inspired by \cite{zhou2019collaborative}, we consider attention mechanisms at feature maps of the classification network \textit{G-Net} given  multi-lesion maps predicted by $S_l$. For this, feature maps at the first and last layers of \textit{G-Net} are jointly combined with lesion maps $\mathbf{z}_{l=1}^L$ to define attention maps that return high responses to different lesion regions characterizing the disease. It is worthy to note that
we do not integrate the lesion masks initially predicted by $S_l$ as a direct input for the classification model \textit{G-Net} because the initially predicted masks are usually very tiny and sparse. Their contributions rather than being controlled by the attention network \textit{Att-Net}. In our setting, we choose the ResNet-50 \cite{resnet50}, including five CNN blocks for the architecture of \textit{G-Net}, and therefore have to modify formulations in \cite{zhou2019collaborative} as follows.

We indicate five CNN blocks of \textit{G-Net} as $G = \{\mathbf{g}^1, \mathbf{g}^2, \mathbf{g}^3, \mathbf{g}^4, \mathbf{g}^5\}$.  The feature maps of an image $\mathbf{x}$ at the first block are computed by:
\begin{align}
    \mathbf{f^{\mathrm{first}}} = \mathbf{g}^{1}(\mathbf{x}) 
    \label{eq:first-fea}
\end{align}
Similarly, the feature maps at the last layer estimated by:
\begin{align}
    \mathbf{f^{\mathrm{last}}}  = \mathbf{g}^{5}(\mathbf{o}^{5})
    \label{eq:last-fea}
\end{align}
where 

%\hl{Duy: h was used Eq. 6,7: Are they the same?}
% $ h^{i} = \mathbf{g}^{i-1}(h^{i-1}),\ i \in [2,5],\ h^{1} = x$. 
\begin{gather}
    \mathbf{o}^{i} = \mathbf{g}^{i-1}(\mathbf{o}^{i-1}),\ i \in [2,5] \\
    \mathbf{o}^{1} = \mathbf{x}
\end{gather}
where $\mathbf{o}^{i}$ indicates the output of each convolutions block.
%\textcolor{red}{Duy: $\mathbf{h}^{i}$ here indicates the output of each convolutions block wherein $\mathbf{h}_s$, $\mathbf{h}_t$ in equations 6, 7 indicate for outputs of the encoder layers. Therefore, we can change $\mathbf{h}$ notations in Eq.(21, 22, 23) with another one, for e.g., $\mathbf{o}^{i} = \mathbf{h}^{i}$}.

In the first step, we train \textit{G-Net} with all blocks in a fully supervised manner using the image-level annotations in the target domain $(\mathbf{x}_t, \hat{\mathbf{t}}_t) \in \mathcal{X}_{t}$. The optimization problem is:
\begin{align}
    \mathcal{L}_{cls} = \underset{\theta_{g}}{\text{min}} \frac{1}{\lvert \mathcal{X}_{t} \rvert} \sum_{(\mathbf{x}_t, \hat{\mathbf{t}}_t) \in \mathcal{X}_{t}}  \mathcal{L}_{mce} (G(\mathbf{x}_t), \hat{\mathbf{t}}_t), 
    \label{eq:classification}
\end{align}
where $\mathcal{L}_{mce}$ is the multi-class cross entropy classification loss defined as:
\begin{align}
    \mathcal{L}_{mce}(\mathbf{t}_t,\hat{\mathbf{t}}_t) = - \sum_{k=1}^{C} \hat{\mathbf{t}}_t^{(k)} \log \mathbf{t}_t^{(k)} 
    \label{eq:multi-cross-entropy}
\end{align}
with $\hat{\mathbf{t}}^{(k)}$  is $0$ or $1$, indicating whether class label $k$ is the correct classification.

%\hl{Duy: Some confusions: $\hat{\mathbf{t}}^{(k)}$ is a pixel or segmentation map. K is as same as L?}

%\textcolor{red}{Duy-Answer\ :$\mathbf{\hat{t}}_{t}$ and $\mathbf{t}_{t}$ represent the ground-truth and predictions in the form of one-hot vector. For e.g., $\mathbf{\hat{t}}_{t} = [0, 0, 0, 1]$,  $\mathbf{t}_{t} = [0.1, 0.1, 0.1, 0.7]$. The correct sum is $\sum_{k=1}^C$ where $C$ is the class numbers.}

Once the model is pre-trained, the feature maps $\mathbf{f}^{\mathrm{first}}$ and $\mathbf{f}^{\mathrm{last}}$ are computed using equations (\ref{eq:first-fea}),\,(\ref{eq:last-fea}). Then we define \textit{an attentive feature for the $l$-th lesion} $\mathbf{y}_l$ obtained from $S_{l}$ by:
\begin{align}
    \mathbf{f}^{\mathrm{first-att}}_{l} = \mathrm{ReLU}(\mathbf{W}_l^{\mathrm{first}} \text{concat}(\mathbf{y}_l, \mathbf{f}^{\mathrm{first}}) + b_l^{\mathrm{first}}), 
\end{align}
where concat(.) is the channel-wise concatenation; $\mathbf{W}_l^{\mathrm{first}}$ and $b_l^{\mathrm{first}}$ are additional learnable parameters and bias terms for the $l$-th lesion.

In a next step, the last feature maps $\mathbf{f}^{\mathrm{last}}$ acting as a global feature embedding is correlated with the first-level attentive features to generate attention weights for the $l$-th lesion:
\begin{align}
    \alpha_l = \text{Sigmoid}(\mathbf{W}_l^{\mathrm{last}} [\mathbf{f}_l^{\mathrm{first-att}} \odot \mathbf{f}^{\mathrm{last}}] + b_l^{\mathrm{last}}),
    \label{eq:att-lesion-l}
\end{align}
where $\odot$ is the element-wise multiplication; $\mathbf{W}_l^{\mathrm{last}}$ and $b_l^{\mathrm{last}}$ are other parameters and bias terms to learn attention features at the global level. To make $\mathbf{f}^{\mathrm{last}}$ and $\mathbf{f}_l^{\mathrm{first-att}}$ be compatible in channel dimensions, we also use a $1\times1$ convolution over the $\mathbf{f}^{\mathrm{last}}$.

By applying Eq.\,(\ref{eq:att-lesion-l}) for each lesion $l \in L$, we aggregate all attention lesion maps and use them to separately conduct element-wise multiplication with the first-level features $\mathbf{f}^{\mathrm{first}}$ of \textit{G-Net}. The output feature vectors then are concatenated and utilized as final attention features to fine-tune the DR Grading network \textit{G-Net}. In terms of optimization, parameters of \textit{Att-Net} include $\theta_{Att} = \{\left(\mathbf{W}_{l}^{\mathrm{first}},\,\mathbf{W}_{l}^{\mathrm{last}},\,b_{l}^{\mathrm{first}},\,b_{l}^{\mathrm{last}}\right)_{l=1}^{L}\}$. 

Finally we compose an optimization, that jointly learns the grading network \textit{G-Net} and the attention-based lesion \textit{Att-Net} at the \textit{low-level concept} as:
\begin{align}
    \mathcal{L}_{cls}^{att}  = \underset{\theta_{g},\theta_{att}}{\text{min}} \frac{1}{\lvert \mathcal{X}_{t} \rvert} \sum_{(\mathbf{x}_t, \hat{\mathbf{t}}_t)\in \mathcal{X}_{t}} \mathcal{L}_{mec} \left(G(\mathbf{x}_t)\cdot Att \left(\,S_{l}(\mathbf{x}_t)_{l=1}^{L}\right),\,\hat{\mathbf{t}}_t\right).
    \label{eq:classification_attention}
\end{align}
where $\mathcal{L}_{mec}$ is the  multi-class cross entropy defined in Eq.\,(\ref{eq:multi-cross-entropy}).  We characterize the loss $\mathcal{L}_{cls}^{att}$ as the low-level concept because it purely combines \textit{G-Net} and \textit{Att-Net} at the feature level through attention weights. While this approach boosts performance, it is still a black box to end-users. In the next section, we introduce a new constraint between two types of networks which are able to enhance accuracy and provide explainable properties for the learned system.

%\hl{Duy: In Eq. 28: How is Attention network applied in to L binary segmentation maps}

\subsubsection{Attention Lesion Regions at High-level Concepts}
\label{subsec:lesion-high-level}
The attention procedure that we have used in the previous section is on the structural feature level of the grading network. In this section, we further impose a new constraint that directly compares the class activation map of the grading network with the lesion maps. By considering this property, we explicitly guide the network on which parts of the image they should focus on and make the predictions more transparent to end-users by observing correlations between trained networks' activation map and medical priors (Fig. \ref{fig:explantion_outputs}). This sets us apart from prior studies in DR grading problems \cite{zhou2019collaborative,sun2021lesion}. Below we present the formulation to integrate this constraint seamlessly in our learning framework for the case of \textit{G-Net} based on CNN architecture.   

% Though, it showed significant improvement in the performance, the learning procedure is still a black-box for the human understating. 

% As we discussed previously, the class activation maps are the models visual explanation about its prediction for a particular class output. We introduce an explanation loss that directly compares the class activation map of the grading network with the lesion maps. 

Based on Grad-CAM, we can get the class activation map of the last layer in the grading model \textit{G-Net}. For a input image $\mathbf{x}$, let $f_{l,k}$ be the activation of unit $k$ in $l$-th layer. For each of the ground-truth class $c$, we compute the corresponding gradient score $s^c$, with respect to activation maps of $f_{l,k}$. These gradient scores then pass through $1 \times 1$ convolution of global average pooling layer to obtain the neuron importance weights $w_{l, k}^c$:
\begin{align}
    w_{l,k}^c = \text{GAP}\left(\frac{\partial s^c}{\partial f_{l,k}}\right),
 \end{align}
where GAP(.) is the global average pooling. Because $w_{l,k}^{c}$ indicates the important of activation map $f_{l,k}$ contributing the prediction of class $c$, we thus apply the weight matrix $w_{c}$ as a kernel and apply 2D convolution over the feature map $f_{l}$ to aggregate all activation maps, followed by a ReLU function to get the activation map $AM^{c}$ for the $c$-th class: 
\begin{align}
    AM^{c} = \text{ReLU}(\mathrm{conv}(f_l, w^c))
\end{align}
where $l$ represents the last convolution layer. 

In the next step, we normalize the $AM^{c}$ so that its class channel values are normalized to $[0, 1]$, denoted as $\widetilde{AM}^{c}$, using a thresholding operation $T(.)$ \cite{visual_attention} as follows:
\begin{gather}
        \widetilde{AM}^{c} = T(AM^{c}) \\
        T(AM^c) = \frac{1}{1 + \exp\left(-\omega(AM^{c} - \boldsymbol{\sigma})\right)}
\end{gather}
where $\boldsymbol{\sigma}$ is the threshold matrix whose elements are equal to $\sigma$. $\omega$ is the scale parameter forcing $T(AM^{c})_{i,j}$ approximately equals to $1$ if $AM^{c}_{ij}$ is greater than $\sigma$, or to $0$ otherwise.

Finally, we propose an overlapping loss function $\mathcal{L}_{overlap}$ for images whose lesion areas are not empty as:
\begin{align}
    \mathcal{L}_{overlap} = \underset{\theta_{g},\theta_{att}}{\text{min}} \frac{1}{hw}\    \Vert\widetilde{AM}^{c} - L_{U} \Vert_{2}
    \label{eq:overlap}
\end{align}
where $h,\,w$ are the height and width of image $x$, $L_{U}$ is the union region of all of lesion types computed by:
\begin{align}
    L_{U} = \bigcup_{l \in L} S_{l}(x)
\end{align}
Note that we only compute the $\mathcal{L}_{overlap}$ for an image $x$ if its grading label is different $0$, i.e., input image has some stages of DR disease. By optimizing $\mathcal{L}_{overlap}$, we jointly learn parameters for both grading network \textit{G-Net} and the attentive network \textit{Att-Net}. 

Combining equations (\ref{eq:overlap}), (\ref{eq:classification_attention}), we end up with a new total object loss function that incorporates attention mechanisms for DR grading task at both low-level and high-level constraints:
\begin{align}
    \mathcal{L}_{grading} = \mathcal{L}_{cls}^{att} + \mathcal{L}_{overlap}
    \label{eq:grading_final}
\end{align}

At a glance, our proposed loss $\mathcal{L}_{overlap}$ is comparable to existing semi-supervised learning \cite{visual_attention} or Covid-19 detection \cite{Wu_2021}; however, we extend it for the multi-lesion scenario in the context of the DR grading task. Furthermore, our system is superior to these works and current approaches to the DR problem \cite{zhou2019collaborative,sun2021lesion} in that we develop lesion information-based attention mechanisms for classification tasks at both feature-level and high-level concepts, thereby improving performance and providing explainable properties to the entire system (table \ref{table:grading_cnn}).

% where $AM_{p,c}^{norm}$ is the activation mapping of the grading class channels with values normalized to $[0, 1]$. and $L$ is the ground-truth for the lesion maps. Combining the losses our final loss for the grading classification with lesion information is:
\subsubsection{Integrating Lesion Features for Transformer-based Methods}
\label{subsec:lesion-trans}

\begin{figure*}[ht]
 \centering
\includegraphics[width=1.7\columnwidth]{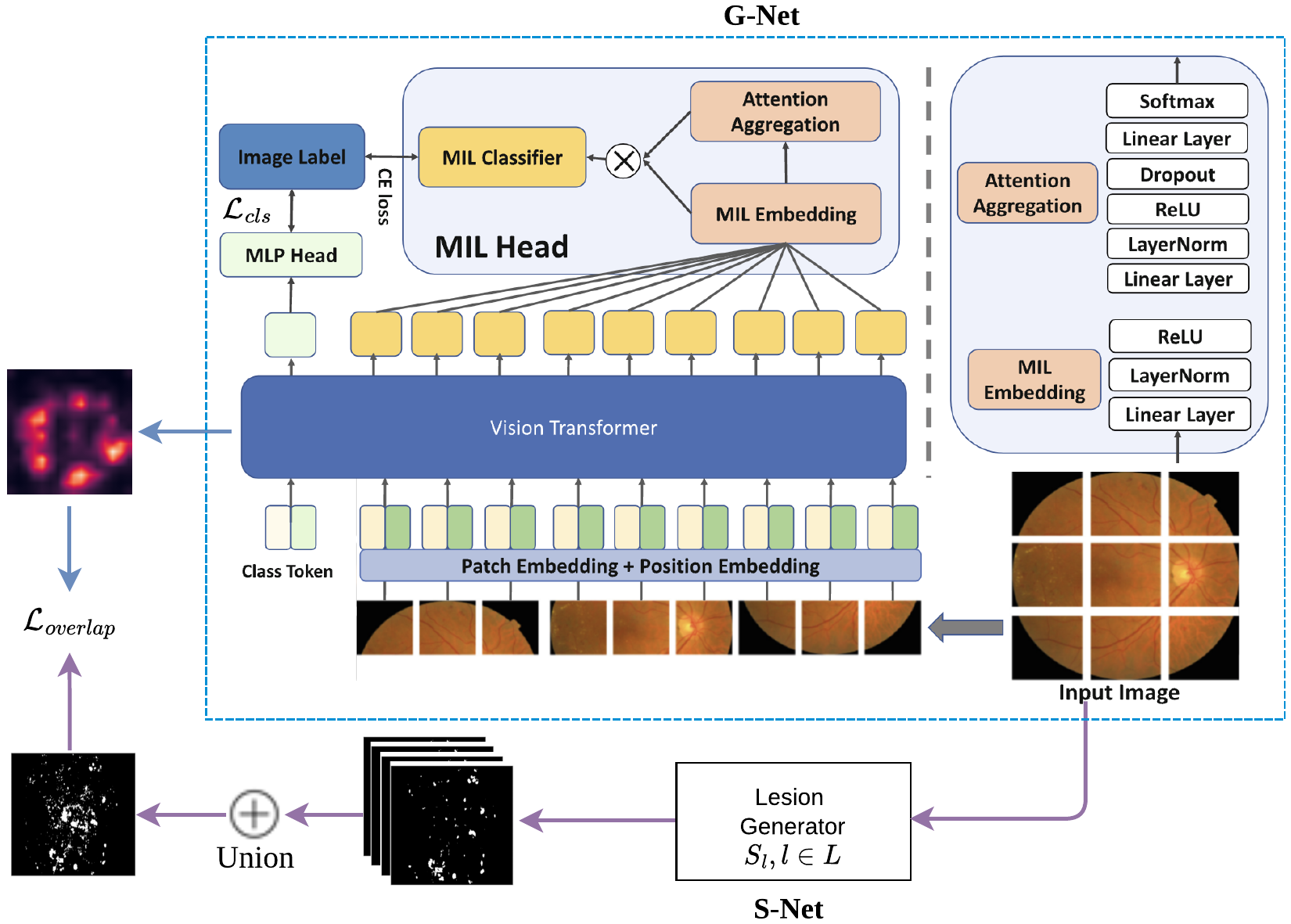}\\
    \caption{Our proposed DR-grading transformer architecture. We integrate predicted lesion feature maps using lesion generator $S_l$ for an image with the multi-head lesion generator \textit{G-Net} based on MIT-VT \cite{yu2021mil}. We compute $\mathcal{L}_{overlap}$ and the grading model is trained using both $\mathcal{L}_{cls}$ and $\mathcal{L}_{overlap}$.}
    \label{fig:att-transformer}
\end{figure*}

% Transformer models have been widely used for natural language processing (NLP) and achieved superior performance \cite{bert}. These models use inherent self-attention techniques automatically learn to focus on the task-specific important regions during the training process. Recently, transformer-based models such as ViT \cite{vit} have been exploited in vision-related tasks, and they have achieved competitive performance against the CNN. The main difference between CNN and transformer models is that CNN uses pixel arrays whereas vision transformers split the images into visual tokens. In Transformer, an input image $x \in \mathbb{R}^{h \times w \times 3}$ is divided into individual patches with size $p \times p$. Therefore, we can obtain $N = hw / p^2$ patches from one single image. The patches are further flattened into a 1D format and then get embedded together using a linear layer of $D$ dimensions to make a compatible data format for the transformer networks. 

%\hl{I am kind of lost here. Where is it in the network?} 

In this section, we investigate the performance of Transformer architecture for the DR grading task. For this, we choose the MIL-VT method proposed by \citet{yu2021mil} (Fig. \ref{fig:att-transformer}). Compared with ViT,  MIL-VT further adds multiple-instance learning heads to leverage the features extracted from individual patches. However, MIL-VT solely employs image-level data in the training stage and ignores the responsibilities of lesion areas.
This motivates us to incorporate the MIL-VT attention's mechanism with our overlapping heatmap concepts presented in section \ref{subsec:lesion-high-level}. To establish such a constraint, we apply the Attention Rollout technique \cite{att-roll,vit} to compute the heatmap regions for MIL-VT. 

In particular, Attention rollout is a concept used to track the information propagated from the input layer to the embeddings in the higher layers. Given a Transformer with $N$ layers, in each layer $n \in N$, this technique takes the average of all attention weights across all heads to form an attention matrix $A_{n}$ where $(A_{ij})_{l}$ defines how much attention is going to flow from token $j$ in the layer $n-1$ to token $i$ in the layer $n$. Then the attention rollout matrix at layer $n$, denoted as $\tilde{A}_{n}$, is computed in a recursive way:
\begin{align}
    \tilde{A}_{n} = (A_{n} + I)\,\tilde{A}_{n-1}\
    \label{eq:attention-roll}
\end{align}
where $I$ is the identity matrix.

By computing equation \ref{eq:attention-roll} at the last layer $N$, we can account for the combination of attention across tokens through all layers. In our method, we choose the attention map $AM^{c}$ as:
\begin{align}
AM^{c} = \tilde{A}_{N}.
\end{align}
Then the loss $\mathcal{L}_{overlap}$ in equation \ref{eq:overlap} is defined in analogous way (Fig. \ref{fig:att-transformer}). In experiment, we discover that this extended version significantly enhances MIL-VT's performance in a variety of situations (table \ref{table:grading_vit}).

\section{DRG-Net: ARG-Expert-Interaction}
\begin{figure*}[ht]
 \centering
\includegraphics[width=1.7\columnwidth]{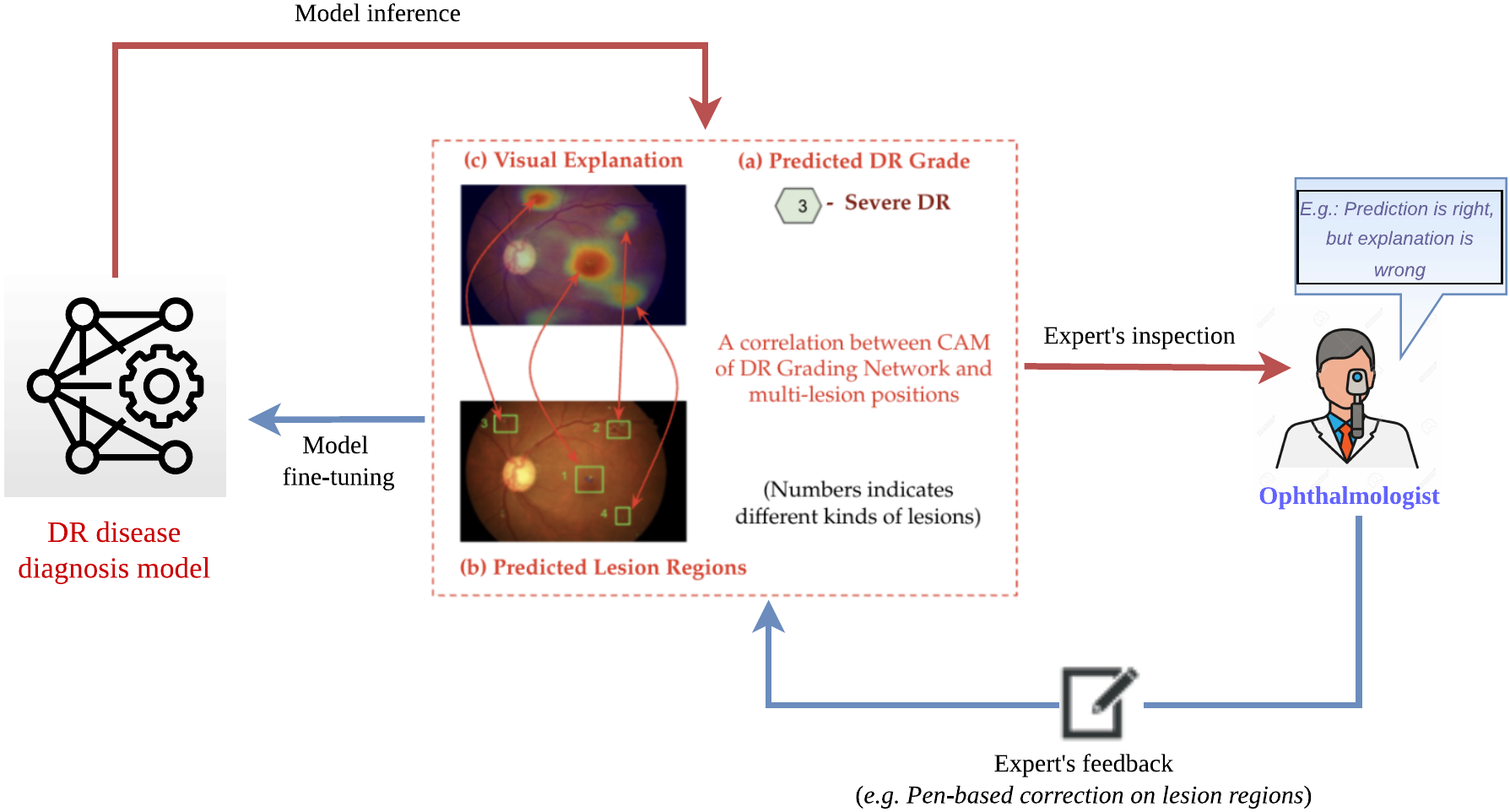}\\
    \caption{Illustration for the interaction between experts and the trained model. The model can infer explanatory predictions via presenting disease grade, related lesions, and class activation mapping. Upon inspection, the expert validates the results and, if necessary, provides input (e.g., drawing bounding box on missing lesion regions). This input is used to improve model performance in subsequent model retraining.}
    \label{fig:user_interaction}
\end{figure*}

As a joint disease diagnosis system, given a retina fundus image, the system can automatically detect the associated lesion attributes and predict its disease grade. In addition, the detected lesion masks from the segmentation module can support the disease grading prediction by the grading module. Unlike most of the medical decision support systems \citep{sonntag2020skincare, mehta2018ynet, dai2021deep} where the lesion maps and disease grading predictions modules are independent of each other, our architecture is trained to learn these tasks collaboratively. In the learning phase of our grading model, we exploit 
predicted lesion information. Besides, the system also facilitates the utilization of expert feedback throughout the training process.
Generally, our diagnosis system provides an interactive way with the expert user, as shown in Fig. \ref{fig:user_interaction}. For this, the method can deliver a visual explanation to the user about its predictions and incorporate user feedback to fine-tune the model. More details are as follows,
% Also, the method is robust enough to incorporate user feedback in weakly-supervised annotations and can use them to fine-tune the model. %We will discuss in the following sections detailing the interaction process.
\subsection{DR Grading Predictions with Explainable Properties}

% Explainability is an imperative feature required for intelligent decision support systems, especially in the healthcare domain. When a healthcare model predicts a disease, the medical practitioner needs to know which factors the model is taking into account. 
Our method focuses on the local interpretability \cite{lime,selvaraju2017grad} of our prediction model, which is computed based on the class activation map $AM^{c}$.
% . This can help us to understand how the model makes decisions for a single instance. We compute the gradient-based class activation map $AM^{c}$ for the disease grading prediction model. 
This gives us the discriminative regions used by the model to predict the disease grade for a certain class. By drawing the $AM^{c}$ over input image $x$ (Fig. \ref{fig:user_interaction} - \textit{Visual Explanation}), we can observe the highlighted region on the input $\mathbf{x}$, which the grading network \textit{G-Net} thought to be the most essential region for its decision (Fig. \ref{fig:user_interaction} - \textit{Predicted DR Grade}).
Moreover, detected lesion maps can be used as visual explanations. In particular, considering the overlapping region between lesion maps $\hat{z}_{{l=1}_l}$ (Fig. \ref{fig:user_interaction} - \textit{Predicted Lesion Regions}) with activation maps $AM^{c}$ of $x$ gives a visual interpretation of how much these lesions influenced the grading model prediction (Fig. \ref{fig:user_interaction}).

\subsection{Improving System's Performance through User Feedback}
\label{subsec:feedback}
% In general, user feedback is the information that a user sends to a learning agent in order to update the agent's knowledge. In interactive machine learning, user feedback can guide the intelligent system to achieve the desired behavior \cite{amershi2014power}. Our proposed method is inherently designed to be able to integrate user feedback in its learning procedure.
Getting user feedback can costly and time-consuming in the medical domain. For instance,  lesion regions in diabetic disease can be so tiny that each region covers only a small group of pixels in the image. Our framework alleviates this issue by not requiring pixel-level training data in the target domain to train segmentation models. It instead is handled by the domain adaptive networks as the Wasserstein network \textit{W-Net} (Section \ref{sub2sec:w-net}) and the Adversarial Discriminator-based on Entropy \textit{AD-Net} (Section \ref{sub2sec:AD-net}). Given this, we can generate different lesion regions while only using pixel-level labels from another source domain.

When the system is deployed in practice, the expert can provide two feedback forms. Firstly, given new annotations on lesion masks, we can fine-tune our lesion generator models \textit{S-Net}. Specifically, for each lesion generator model $S_l$, we update the segmentation model using \textit{new labeled data in the target domain} by training:
\begin{align}
   \mathcal{L}_{seg} = \underset{\theta_{S_{l}}}{\text{min}} \frac{1}{\lvert \mathcal{X}_t \rvert} \sum_{(\mathbf{x}, z_{l}) \in \mathcal{X}_t} \mathcal{L}_{wbce} (S_{l}(\mathbf{x}), \hat{\mathbf{y}}_{l})
\label{eq:seg-update}
\end{align}
where $\mathcal{L}_{wbce}$ is the weighted binary cross-entropy loss defined in equation (\ref{eq:wbce}). Note that in this case, we can ignore the domain adaptation parts, which are already optimized during training with data in the source domain $X_s$.

Secondly, when users provide both new image grading and lesion segmentation labels, we can update either:
\begin{itemize}
    \item  Segmentation models \textit{S-Net} using $\mathcal{L}_{seg}$ in equation (\ref{eq:seg-update}).
    \item  Attention-based disease grading model \textit{Att-Net} and \textit{G-Net} by learning the objective function $\mathcal{L}_{grading}$ defined in equation (\ref{eq:grading_final}).
\end{itemize}

It is worth noting that our technique automatically learns to identify critical lesion locations influencing the DR Grading task by utilizing attention mechanisms; hence, it can be resilient to a certain degree of noise in segmentation annotations provided by experts. In other words, rather than a precise segmentation mask, the expert can mark the region of interest in the form of bounding boxes or circles around the lesion locations using any pen-based input device. As the attention model inherently learns to filter out the noise and focus on the area of interest, our experiments confirm that only these soft annotations are sufficient to boost the performance of the model (table \ref{table:user_feedback}).

\section{Experimental Results} \label{sec-experiments}
\subsection{Datasets}
Retina images are generally captured in two forms, which are Optical Coherence Tomography (OCT) capturing a cross-sectional images of retinasa and Color Fundus retinal photography capturing 3D retina images using fundus cameras. We employed datasets with the later one for all our experiments in this work. For domain-invariant lesion attributes segmentation, we used two publicly available datasets. To the best of our knowledge, these two are the only datasets that provide pixel-level annotations for diabetic retinopathy. The datasets are:

\begin{itemize}
    \item \textit{IDRID segmentation }\cite{h25w98-18}: This dataset contains 81 high resolution ($4288 \times 2800$ pixels) images with four different types of lesion annotations and is split into 54 training images and 27 testing images. The lesions are: microaneurysms (MA), haemorrhages (HE), hard exudates (EX) and soft exudates (SE). Each of these lesions are annotated in the forms of binary masks.
    \item \textit{FGADR segmentation }\cite{zhou2020benchmark}: Similarly as IDRID, this dataset contains 1843 images with 4 kinds of lesion masks (MA, HE, EX, and SE). The training set consists of 1500 images and the remaining 343 images are for test set. This dataset has imbalance in the lesion classes. The class HE is available in 1471 images which is 78\% of the total images where the class MA is available for only 610 images which consists of 33\% of the total images.
\end{itemize}

For the classification task, we used three publicly available datasets.

\begin{itemize}
    \item \textit{IDRID classification } \cite{h25w98-18}: This dataset consists of 413 training and 103 test images with 5 severity grading labels. The five labels include: \textit{\textit{normal, mild, moderate, severe and proliferative}}, which are annotated as \textit{0, 1, 2, 3, 4}, respectively. These images only have image-level labels and do not have any pixel-level lesion masks annotations.
    
    \item \textit{EyePACS } \cite{kaggleEyePacsDataset}: This dataset consists of 35,126 training images and 53,576 testing images with similar grading protocol of classes with 5 categories. Images in this dataset is captured using different types of cameras settings under various light conditions.
    
    \item \textit{FGADR classification }\cite{zhou2020benchmark}: this dataset consists of 1500 training images and 343 testing image for diabetic retinopathy grading. This is the only one that contains both image-level pixel disease grading and pixel-level lesion annotation maps. 
    
\end{itemize}

\begin{figure*}[ht]
 \centering
\includegraphics[width=1.5\columnwidth]{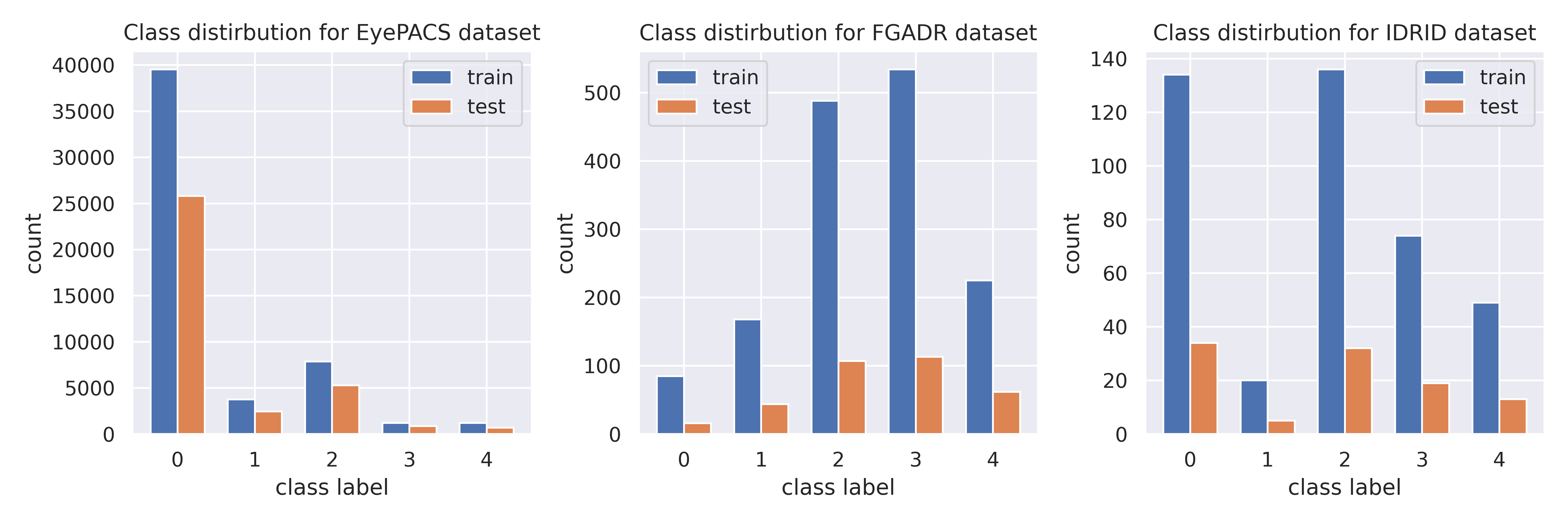}\\
    \caption{Class distributions in the three classification datasets used in our work.}
    \label{fig:grade_class}
\end{figure*}

Figure \ref{fig:grade_class} shows the class distribution for the three classification datasets we used in our work. We can observe that there are great numbers of class imbalance in all three datasets.

\subsection{Evaluation Metric}

To evaluate the segmentation performance of multi-lesion segmentation, we used two common metrics i.e., Area Under the Curve of Receiver Operating Characteristics (AUC-ROC) and Area Under the Curve of Precision-Recall (AUC-PR). The positive class means the existence of lesion and negative is the otherwise. For evaluating multi-class disease grading classification task, we used Accuracy and Quadratic Weighted Kappa (Kappa). Detailed definitions of these measures are as follows:

\begin{itemize}
    \item \textit{AUC-ROC} measures the class separability at various threshold settings. ROC is the probability curve and AUC represents the degree of measures of separability. It compares true positive rate (sensitivity/recall) versus the false positive rate (1 - specificity). The higher the AUC-ROC, the bigger the distinction between the true positive and false negative.
    
    \item \textit{AUC-PR:} It combines the precision and recall, for various threshold values, it compares the positively predicted value (precision) vs the true positive rate (recall). Both precision and recall focus on the positive class (the lesion) and unconcerned about the true negative (not a lesion, which is the majority class). Thus, for class imbalance, PR is more suitable than ROC. The higher the AUC-PR, the better the model performance.
    \item \textit{Accuracy:} The normal classification accuracy which simply measures how many observations are correctly classified. 
    \item \textit{Quadratic Weighted Kappa (Kappa):} It is similar to the Cohen's kappa metric \cite{mchugh2012interrater} by the weights are set to 'Quadratic'. Cohen's kappa measures the agreement between two raters who each classify $N$ number of items into $C$ mutually exclusive categories. The formulation is:
    \begin{align}
        \mathrm{kappa} = \frac{p_{o} - p_{e}}{1 - p_{e}},
    \end{align}
    where $p_o$ denotes the relative observed agreement between the raters and $p_e$ is the hypothetical probability of chance agreement. 
\end{itemize}

\subsection{Implementation Details}

We implemented all experiments using PyTorch framework \cite{NEURIPS2019_9015} and executed on a computer 2 NVIDIA TITAN RTX 24GB GPUs. For all experiments, the sizes of the input images are $512 \times 512$. 

\subsubsection{Domain Invariant Lesion Attribute Learning}

For the experiments in lesion attribute segmentation discussed in section \ref{subsec:Snet}, we used pixel values in the range [0, 1] for retina images and ground-truth binary lesion segmentation masks. In this step, the input image are labeled images from source domain and unlabeled images for the target domain. We applied prepossessing steps as proposed in \cite{Xiao_2019} and data augmentation for input images.
% \begin{itemize}
%     \item \textit{Contrast limited adaptive histogram equalization (CLAHE):} Instead of processing the entire image, CLAHE processes  smaller regions. We applied the CLAHE technique with to $8 \times 8$ patches.
%     \item \textit{Denoising}: Assuming that the images contain Gaussian white noise, we applied Non-local means denoising algorithm \cite{buades2011non} with a filter strength of 10.
% \end{itemize}

% Moreover, we applied the built-in data augmentation procedure of PyTorch framework. Each image and its corresponding lesion masks jointly are randomly crop to $512 \times 512$ pixel and randomly rotated with a maximum angle of $30^{\circ}$. Finally, we applied normalization to each channel of the input lesion image with $\mathrm{mean} = [0.485, 0.456, 0.406]$ and $\mathrm{std} = [0.229, 0.224, 0.225]$.

Our \textit{S-Net} model described in section \ref{subsec:Snet} is a U-shaped \cite{ronneberger2015unet} encoder-decoder segmentation network. The encoder architecture is identical to the convolutional block of the ResNet50 \cite{resnet50} model and the decoder architecture is up-scaled accordingly. The architecture of the \textit{PD-Net} described in section \ref{subsec:GAN-source} is similar to the InfoGAN \cite{chen2016infogan}. This discriminator model provides stable performance for our case. It consists of two convolution layers with 64 and 128 kernels, respectively, followed by two fully-connected layer of 1024 dimensions and then the sigmoid layer. 

For training the \textit{PD-net}  \ref{subsec:GAN-source}, we used SGD with momentum \cite{liu2018accelerating} as optimizer with initial learning rate of $10^{-4}$ and momentum value of 0.9. We adopted the cyclic learning scheduler \cite{smith2017cyclical} with max value of $10^{-2}$. The class balancing hyper-parameter weight $\beta$ for the segmentation loss $\mathcal{L}_{wbce}$ described in equation \ref{eq:wbce} is set to 10 following \cite{Xiao_2019} to address class imbalance for lesion and non-lesion class. The value of $\lambda_{p}$ for the source domain adversarial joint optimization in equation (\ref{eq:supervised_joint}) is set to $10^{-2}$. 
For domain adaptation fine-tuning training, pretrained \textit{S-Net} from \ref{subsec:GAN-source} is trained with Wasserstein critic model \textit{W-Net} (section \ref{sub2sec:w-net}) and entropy-based discriminatory network \textit{AD-Net}. \textit{W-Net} is a fully connected neural network which takes 1D encoder block outputs from \textit{S-Net} as inputs. W-Net has two fully connected layers with number of neurons of 128 and 64, respectively, and final output is linear layer which gives a single scalar value, the critic score. \textit{AD-Net} architecture is adopted from \cite{advent} which have five convolutional layers. The input for AD-Net is the weighted self-information maps computed on the output of S-Net using equation (\ref{eq:self_information_formula}) and output is the binary class probability score on input being source or target domain. 

For training both \textit{W-Net} and \textit{AD-Net}, we used Adam optimizer with a learning rate $10^{-4}$. The optimizer hyper-parameters for S-Net in this step are kept unchanged and they are similar to the one discuss in the previous step. Following \cite{advent}, the weight factor $\lambda_{ent}$ in Eq.\, (\ref{eq:joint_seg_loss_1}) and $\lambda_{adv}$ in Eq.\, (\ref{eq:final_segmentation_formula}) are $10^{-3}$. We used batch size of 8 for our experiments.

\subsubsection{Lesion Attentive Grading Model}
To train the lesion attentive grading models \textit{G-Net} and \textit{Att-Net}, we followed the similar pre-possessing and augmentation steps for the input images as with the grading model. %We also used data augmentation functions in PyTorch as
% random-crop, horizontal-flip, vertical-flip, color-distortion, rotation, and translation. 

The backbone of our lesion attentive grading model is a ResNet50 \cite{resnet50} architecture with the output fully connected layer modified to predict five classes.  The attention module \textit{Att-Net} consists of multiple convolutional layers. This network is optimized using Eq.\, (\ref{eq:grading_final}) using the cross-entropy loss. As we can see from figure \ref{fig:grade_class}, there is a significant class imbalance 
among different types of lesions across all datasets, we thus compute class-weight for each lesion attribute during the optimization step. We used the SGD optimizer \cite{ruder2016overview} with initial learning rate $10^{-3}$ and applied the cyclic learning strategy as in lesion segmentation steps. For all the experiments, we varied the batch size between 16 and 32. For all of our experiments we have used 10\% of training data as validation set and final results are computed on test set. 

\subsection{Performance of Multi-Lesion Segmentation Task}
\subsubsection{Influence of Pre-training Steps on Lesion Generator Models}
\label{subsec:result_source}
As discussed in section \ref{subsec:Snet}, our lesion segmentation model \textit{S-Net} is firstly pre-trained using the lesion segmentation labels from the source domain data in a fully supervised manner. To evaluate the effectiveness of task agnostic transfer learning (TATL) (\cite{nguyen2021tatl}) and the adversarial learning \textit{(PD-Net)} during the pre-training step, we test with three settings:
\begin{itemize}
    \item \texttt{S-Net}: Lesion feature generator segmentation model (\textit{S-Net}) is trained only using the segmentation loss $\mathcal{L}_{seg}$ in equation \ref{eq:seg}. We do not adapt either TATL strategy or adversarial trainings.
    \item \texttt{S-Net + PD-Net:} We incorporate adversarial learning strategy on source domain as discussed in section \ref{subsec:GAN-source}. We jointly the \textit{PD-Net} with \textit{S-Net} segmentation model by optimization the formula in equation \ref{eq:supervised_joint}.
    \item \texttt{S-Net + PD-Net + TATL:} We apply both \textit{task agnostic transfer learning (TATL)} and adversarial learning with \textit{PD-Net} to learn \textit{S-Net} using data from the source domain.
\end{itemize}

Table \ref{table:tatl_idrid} and \ref{table:tatl_fgadr} compare the results of different settings for \textit{IDRID} and \textit{FGADR} segmentation datasets respectively. The segmentation performance is evaluated by AUC-ROC and AUC-PR values on four different lesions, including microaneurysms (MA), haemorrhages (HE), hard exudates (EX) and soft exudates (SE). For \textit{IDRID-segmentation} dataset in table \ref{table:tatl_idrid}, we can observe slight improvement in performance for settings \texttt{S-Net + PD-Net} compared with original \texttt{S-Net} settings. With the combinations of TATL in settings \texttt{S-Net + PD-Net + TATL}, we observe significant improvements in results for all of the lesion classes. For instance, comparing with \texttt{S-Net} settings, we gained 2.5\% and 5.5\% on AUC-ROC and AUC-PR scores respectively, averaged over all four lesions. For the lesion class \textit{HE}, our method achieved highest improvement in AUC-PR scores of 9.4\%. 

We can observe the similar trend in results on the test set of \textit{FGADR-segmentation} dataset in table \ref{table:tatl_fgadr}. There is an improvement of 8.5\% in the AUC-PR scores averaged over all four lesion classes. To summarize from the results, we conclude that combining adversarial training using the \textit{PD-Net} and the \textit{TATL} strategy provided a better pre-training performance for \textit{S-Net} in the source domain supervised learning. 

% For \textbf{FGADR} segmentation dataset, we can observer the similar improvement in scores in all four kinds of lesions in table \ref{table:tatl_fgadr}. compare \textbf{S-Net} baseline method, \textbf{S-Net + PD-Net} has improved the AUC-ROC and AUC-PR scores by 1.38\% and 4.37\% on average and \textbf{S-Net + PD-Net + TATL} has an average improvement of 2.45\% in AUC-ROC and \textbf{8.5\%} in AUC-PR score. We observe the largest improvement in performance for \textbf{hard exudates (EX)} and \textbf{soft exudates (SE)} lesions. We have achieved 11.2\% and 14.3\% improvement for AUC-PR score respectively for these lesion classes.

\begin{table*}[!t]
\centering
\caption{Contributions of task agnostic transfer learning TATL and Adversarial training PD-Net (section \ref{subsec:GAN-source}) in learning lesion segmentation model S-Net in the source domain. Results are evaluated using the training and testing set of \textit{IDRID segmentation}.}
\scalebox{0.90}{
\begin{tabular}{lcccccccc}
% \toprule
\hline
\multirow{2}{*}{Methods}  & \multicolumn{2}{c}{MA} & \multicolumn{2}{c}{HE} & \multicolumn{2}{c}{EX} & \multicolumn{2}{c}{SE} \\ \cline{2-9}
                          & ROC  &  PR    & ROC  &  PR    &  ROC  &  PR  & ROC  &  PR    \\ \hline
S-Net                                             & 0.937           & 0.440           & 0.896           & 0.459          & 0.931           & 0.722          & 0.953          & 0.583          \\
S-Net + PD-Net                                    & 0.932           & 0.439          & 0.917           & 0.481          & 0.946           & 0.735          & 0.965          & 0.611          \\
S-Net+PD-Net+TATL & \textbf{0.953}  & \textbf{0.456} & \textbf{0.931}  & \textbf{0.553} & \textbf{0.961}  & \textbf{0.772} & \textbf{0.970} & \textbf{0.643} \\ %\bottomrule
\hline
\end{tabular}}
\label{table:tatl_idrid}
\end{table*}

\begin{table*}[!t]
\centering
\caption{Contributions of task agnostic transfer learning TATL (\cite{nguyen2021tatl}) and Adversarial training PD-Net (section \ref{subsec:GAN-source}) in learning lesion segmentation model S-Net in the source domain. Results are evaluated using the training and testing set of 
\textit{FGADR segmentation}.}
\label{table:tatl_fgadr}
\scalebox{0.90}{
\begin{tabular}{lcccccccc}
% \toprule
\hline
\multirow{2}{*}{Methods} & \multicolumn{2}{c}{MA} & \multicolumn{2}{c}{HE} & \multicolumn{2}{c}{EX} & \multicolumn{2}{c}{SE} \\ \cline{2-9}
                         & ROC  &  PR    & ROC  &  PR    &  ROC  &  PR  & ROC  &  PR    \\ \hline
{\color[HTML]{202124} S-Net}                      & {\color[HTML]{202124} 0.901}          & {\color[HTML]{202124} 0.373}          & {\color[HTML]{202124} 0.941}          & {\color[HTML]{202124} 0.611}          & {\color[HTML]{202124} 0.947}          & {\color[HTML]{202124} 0.602}          & {\color[HTML]{202124} 0.927}          & {\color[HTML]{202124} 0.410}          \\
{\color[HTML]{202124} S-Net + PD-Net}             & {\color[HTML]{202124} 0.926}          & {\color[HTML]{202124} 0.394}          & {\color[HTML]{202124} 0.955}          & {\color[HTML]{202124} 0.638}          & {\color[HTML]{202124} 0.959}          & {\color[HTML]{202124} 0.667}          & {\color[HTML]{202124} 0.941}          & {\color[HTML]{202124} 0.492}          \\
{\color[HTML]{202124} S-Net+PD-Net+TATL} & {\color[HTML]{202124} \textbf{0.937}} & {\color[HTML]{202124} \textbf{0.417}} & {\color[HTML]{202124} \textbf{0.963}} & {\color[HTML]{202124} \textbf{0.652}} & {\color[HTML]{202124} \textbf{0.970}} & {\color[HTML]{202124} \textbf{0.714}} & {\color[HTML]{202124} \textbf{0.954}} & {\color[HTML]{202124} \textbf{0.553}} \\ \hline
\end{tabular}}
\end{table*}

\vspace{0.2in}
\subsubsection{Influence of Domain Adaption on Lesion Generator Models}
\label{subsec:result_domain}

In this section, we present the results of our domain adaption approaches in different settings and compare with different methods. As discussed in section \ref{subsec:UDA}, the aim of domain adaptation is to train a neural network using available labeled data from source domain and secure a good accuracy on target domain. We evaluated our approach on IDIRD and FGADR datasets in the forms of \textit{source $\rightarrow$ target}:

\begin{itemize}
    \item \textit{IDRID $\rightarrow$ FGADR}: evaluation on \textit{FGADR-segmentation} test set when domain adaptive segmentation model is trained using the \textit{labeled} data from \textit{IDRID-segmentation} dataset and \textit{unlabeled} images from \textit{FGADR-segmentation} dataset.

    \item \textit{FGADR $\rightarrow$ IDRID}: evaluation on \textit{IDRID-segmentation} test set when domain adative segmentation model is trained using the \textit{labeled} data from \textit{FGADR-segmentation} dataset and \textit{unlabled} images from \textit{IDRID-segmentation} dataset.
\end{itemize}

% \usepackage{multirow}
\begin{table*}[!t]
\centering
\caption{\textit{IDRID $\rightarrow$ FGADR:} Semantic segmentation performance on \textit{FGADR} (\textit{target domain}). Models are trained with labeled data on \textit{IDRID} and increasingly labeled data in the target domain from $0\rightarrow 100\%$. \textcolor{red}{Red} indicates the best results for settings using 0\% labeled data from target domain in the training step. \textcolor{green}{Green} stands for settings using $40\%-60\%$ labeled data in target domain but outperform at least one of the methods trained with $100\%$ data. \textbf{Bold} are the best values in all methods.}
\label{table:idrid_to_fgadr}
\scalebox{0.85}{
\begin{tabular}{llcccccccc}
% \toprule
\hline
\multirow{2}{*}{Target Domain}  &  \multirow{2}{*}{Methods}  & \multicolumn{2}{c}{MA} & \multicolumn{2}{c}{HE} & \multicolumn{2}{c}{EX} & \multicolumn{2}{c}{SE} \\ \cline{3-10}
 &   &  ROC  &  PR    & ROC  &  PR    &  ROC  &  PR  & ROC  &  PR    \\ \hline
                                                  & S-Net              & 0.752                                 & 0.243                                 & 0.796                                                & 0.280                                 & 0.794                                 & 0.311                                 & 0.728                                 & 0.264                                 \\
                                                  & S-Net + \textit{Entropy}    & 0.807                                 & 0.313                                 & 0.843                                                & 0.357                                 & 0.855                                 & 0.407                                 & 0.819                                 & 0.341                                 \\
                                                  & S-Net + AD-Net     & 0.841                                 & 0.348                                 & 0.903                                                & 0.448                                 & 0.917                                 & 0.473                                 & 0.902                                 & 0.443                                 \\
\multirow{-4}{*}{\textit{0\%}}                    & S-Net+AD-Net+W-Net & {\color[HTML]{EA4335} \textbf{0.894}} & {\color[HTML]{EA4335} \textbf{0.357}} & {\color[HTML]{EA4335} \textbf{0.911}}                & {\color[HTML]{EA4335} \textbf{0.502}} & {\color[HTML]{EA4335} \textbf{0.939}} & {\color[HTML]{EA4335} \textbf{0.538}} & {\color[HTML]{EA4335} \textbf{0.915}} & {\color[HTML]{EA4335} \textbf{0.522}} \\ \hline
\textit{40\%}                                     & S-Net+AD-Net+W-Net & {\color[HTML]{009901} 0.938}          & {\color[HTML]{34A853} 0.411}          & 0.953                                                & 0.613                                 & 0.966                                 & 0.682                                 & 0.965                                 & 0.634                                 \\
\textit{60\%}                                     & S-Net+AD-Net+W-Net & 0.946                                 & 0.438                                 & {\color[HTML]{34A853} 0.969} & {\color[HTML]{34A853} 0.648}          & {\color[HTML]{000000} 0.979}          & {\color[HTML]{34A853} 0.728}          & 
{\color[HTML]{34A853} 0.971}          & 
{\color[HTML]{34A853} 0.655}          \\
\textit{80\%}                                     & S-Net+AD-Net+W-Net & 0.954                                 & 0.458                                 & 0.973                                                & 0.671                                 & {0.981}          & {\color[HTML]{000000} 0.732}          & 0.977                                 & 0.684                                 \\
\textit{100\%}                                    & S-Net+AD-Net+W-Net & \textbf{0.958}                        & \textbf{0.462}                        & \textbf{0.979}                                       & 0.676                                 & \textbf{0.990}                        & \textbf{0.739}                        & \textbf{0.984}                        & \textbf{0.693}                        \\ \hline
                                                  & FCN-8s \cite{zhou2020benchmark}            & 0.925                                 & 0.363                                 & 0.962                                                & 0.606                                 & 0.981                                 & 0.686                                 & 0.963                                 & 0.642                                 \\
                                                  & U-Net  \cite{zhou2020benchmark}            & 0.927                                 & 0.382                                 & 0.967                                                & 0.643                                 & 0.982                                 & 0.726                                 & 0.977                                 & 0.683                                 \\
                                                  & DL-V3+ \cite{zhou2020benchmark}           & 0.934                                 & 0.364                                 & 0.973                                                & 0.619                                 & 0.981                                 & 0.708                                 & 0.967                                 & 0.659                                 \\
\multirow{-4}{*}{\textit{100\%}}                  & Attention U-Net  \cite{zhou2020benchmark}  & 0.942                                 & 0.435                                 & 0.974                                                & \textbf{0.678}                        & 0.984                                 & 0.731                                 & 0.980                                 & 0.685                                 \\ %\bottomrule
\hline
\end{tabular}}
\end{table*}

% \usepackage{multirow}
We report the results for domain adaptation on tables \ref{table:idrid_to_fgadr} and \ref{table:fgadr_to_idrid}. Our domain adaption approach discussed in section \ref{subsec:UDA} consists of modules for Adversarial Entropy Minimization and Wasserstein based distance minimization between source and taget domain. To evaluate the effectiveness of different constraints for our domain adaptive segmentation model, we considered four experiments settings:

\begin{enumerate}
    \item \texttt{S-Net:} The baseline approach for our domain adaptation approaches. \textit{S-Net} is trained on source domain by adopting the pre-training method including PD-Net and TATL. This model is trained using only the equation \ref{eq:supervised_joint}. Neither domain adaptation constraints are considered.
    % As discussed in the previous section, this model is trained in a fully supervised manner using the optimization formula in equation \ref{eq:supervised_joint} to minimize segmentation loss $\mathcal{L}_{seg}$ and $\mathcal{L}_{patch}$. Here, we do not incorporate any domain adaptation constraints or any image/data information from the target domain in the training process of \textit{S-Net}.
    \item \texttt{S-Net + Entropy:} We apply direct entropy minimization constraint for target domain in the learning process of \textit{S-Net} as discussed in section \ref{subsec:Snet}. The entropy loss $\mathcal{L}_{ent}$ in equation\,(\ref{eq:entropy_loss}) is used for minimization on the unlabeled data from target domain and model is optimized with the segmentation loss $\mathcal{L}_{seg}$ using equation (\ref{eq:joint_seg_loss_1}).
    
    \item \texttt{S-Net + AD-Net:} Instead of using direct entropy minimization $\mathcal{L}_{ent}$, we use the proposed domain adaptation with minimizing entropy-based adversarial learning \textit{AD-Net}. \textit{S-Net} and \textit{AD-Net} are optimized together by combining adversarial loss $\mathcal{L}_{adv}$ in equation\,(\ref{eq:ad_net}) with $\mathcal{L}_{seg}$ in equation (\ref{eq:joint_seg_loss_1}). 
    % incorporate adversarial learning for entropy minimization on target domain by introducing domain discriminatory network \textit{AD-Net}.  \textit{S-Net} and \textit{AD-Net} are optimized together in an adversarial manner discussed in section \ref{sub2sec:ad-net}, using the labeled data from source domain and unlabeled data from target domain in order to minimize the domain gap in weighted self-information space Eq.\,(\ref{eq:self-information}). \textit{S-Net} and \textit{AD-Net} is optimized together by combining adversarial loss $\mathcal{L}_{adv}$ Eq.\,(\ref{eq:ad_net}) with $\mathcal{L}_{seg}$. 
    \item \texttt{S-Net + AD-Net + W-Net: } Additional domain critic \textit{W-Net} is added along with \textit{S-Net} and \textit{AD-Net} to minimize the domain gap in feature representation using Wasserstein loss $\mathcal{L}_{wass}$. This is our proposed setting where all modules are jointly trained using our optimization problem in equation (\ref{eq:final_segmentation_formula}). 
\end{enumerate}

\begin{table*}[!t]
\centering
\caption{\textit{FGADR $\rightarrow$ IDRID:} Semantic segmentation performance on \textit{IDRID} (\textit{target domain}). Models are trained with labeled data on \textit{FGADR} and increasingly labeled data in the target domain from $0\rightarrow 100\%$. \textcolor{red}{Red} indicates the best results for settings using 0\% labeled data from target domain in the training step. \textcolor{green}{Green} stands for settings using $40\%-60\%$ labeled data in target domain but outperform at least one of baselines trained with $100\%$ data. \textbf{Bold} are the best values in all methods.}
\label{table:fgadr_to_idrid}
\scalebox{0.85}{
\begin{tabular}{llcccccccc}
\hline
\multirow{2}{*}{Target Domain}  &  \multirow{2}{*}{Methods}  & \multicolumn{2}{c}{MA} & \multicolumn{2}{c}{HE} & \multicolumn{2}{c}{EX} & \multicolumn{2}{c}{SE} \\ \cline{3-10}
 &   &  ROC  &  PR    & ROC  &  PR    &  ROC  &  PR  & ROC  &  PR    \\ \hline
                                                   & S-Net                                   & 0.813                                                & 0.232                                                & 0.803                                 & 0.251                                 & 0.837                                 & 0.363                                 & 0.783                                 & 0.241                                 \\
                                                   & S-Net + Entropy                         & 0.884                                                & 0.331                                                & 0.874                                 & 0.375                                 & 0.902                                 & 0.531                                 & 0.865                                 & 0.404                                 \\
                                                   & S-Net + AD-Net                          & 0.925                                                & 0.408                                                & 0.902                                 & 0.447                                 & 0.911                                 & 0.697                                 & 0.879                                 & 0.447                                 \\
\multirow{-4}{*}{\textit{0\%}}                     & S-Net+AD-Net+W-Net                      & {\color[HTML]{EA4335} \textbf{0.947}}                & {\color[HTML]{EA4335} \textbf{0.436}}                & {\color[HTML]{EA4335} \textbf{0.911}} & {\color[HTML]{EA4335} \textbf{0.462}} & {\color[HTML]{EA4335} \textbf{0.927}} & {\color[HTML]{EA4335} \textbf{0.772}} & {\color[HTML]{EA4335} \textbf{0.890}} & {\color[HTML]{EA4335} \textbf{0.476}} \\ \hline
\textit{40\%}                                      & \multicolumn{1}{l}{S-Net+AD-Net+W-Net} & {\color[HTML]{34A853} 0.971} & {\color[HTML]{34A853} 0.483} & {\color[HTML]{34A853} 0.957}          & {\color[HTML]{34A853} 0.631}          & {\color[HTML]{34A853} 0.946}          & {\color[HTML]{34A853} 0.829}          & 0.937                                 & {\color[HTML]{34A853} 0.634}          \\
\textit{60\%}                                      & \multicolumn{1}{l}{S-Net+AD-Net+W-Net} & 0.983                                                & 0.502                                                & 0.968                                 & 0.658                                 & 0.969                                 & 0.841                                 & {\color[HTML]{34A853} 0.943}          & 0.669                                 \\
\textit{80\%}                                      & \multicolumn{1}{l}{S-Net+AD-Net+W-Net} & 0.985                                                & 0.510                                                & 0.979                                 & 0.682                                 & 0.977                                 & 0.847                                 & 0.959                                 & 0.710                                 \\
\textit{100\%}                                     & \multicolumn{1}{l}{S-Net+AD-Net+W-Net} & \textbf{0.988}                                       & \textbf{0.511}                                       & \textbf{0.982}                        & \textbf{0.700}                        & \textbf{0.978}                        & \textbf{0.851}                        & \textbf{0.961}                        & \textbf{0.714}                        \\ \hline
                                                   & Adv. HEDNet \cite{Xiao_2019}                             & -                                                    & 0.439                                                & -                                     & 0.483                                 & -                                     & 0.840                                 & -                                     & 0.481                                 \\
                                                   & AdvSeg \cite{hung2018adversarial}                                 & 0.961                                                & 0.470                                                & 0.924                                 & 0.592                                 & 0.945                                 & 0.79                                  & 0.939                                 & 0.675                                 \\
                                                   & ASDNet \cite{zhou2019collaborative}                                  & 0.969                                                & 0.478                                                & 0.932                                 & 0.628                                 & 0.950                                 & 0.809                                 & 0.948                                 & 0.692                                 \\
\multirow{-4}{*}{\textit{100\%}}                   & CoLL  \cite{zhou2019collaborative}                                 & 0.965                                                & 0.473                                                & 0.954                                 & 0.657                                 & 0.967                                 & 0.845                                 & 0.953                                 & 0.716                                 \\ %\bottomrule
\hline
\end{tabular}}
\end{table*}
\paragraph{Results on IDRID $\rightarrow$ FGADR:} In table \ref{table:idrid_to_fgadr}, we report the results for different settings on \textit{FGADR segmentation} test set. These models are trained using the labeled data from \textit{IDRID} training dataset and unlabeled data from \textit{FGADR} dataset. We can observe that the settings without any domain adaptation \textit{S-Net} performed poorly during the inference on \textit{FGADR} domain. The AUC-PR scores for all of the lesion classes are below 0.3. Direct entropy minimization method (\texttt{S-Net + Entropy}) has significant improvement in both the metrics with respect to \texttt{S-Net}. Introducing Adversarial network for entropy minimization \texttt{(S-Net + AD-Net)} has improved the AUC-ROC and AUC-PR scores by 12\% and 15.3\%, respectively, averaged over all four lesion classes. The best results among our domain adaptive settings with $0\%$ labeles data in target domain are highlighted in red. In short, we achieved the best results for all of the lesion classes when both adversarial entropy minimization and Wasserstein domain critic model are considered together.

\begin{figure}[ht]
\centering
\includegraphics[width=1.0\columnwidth]{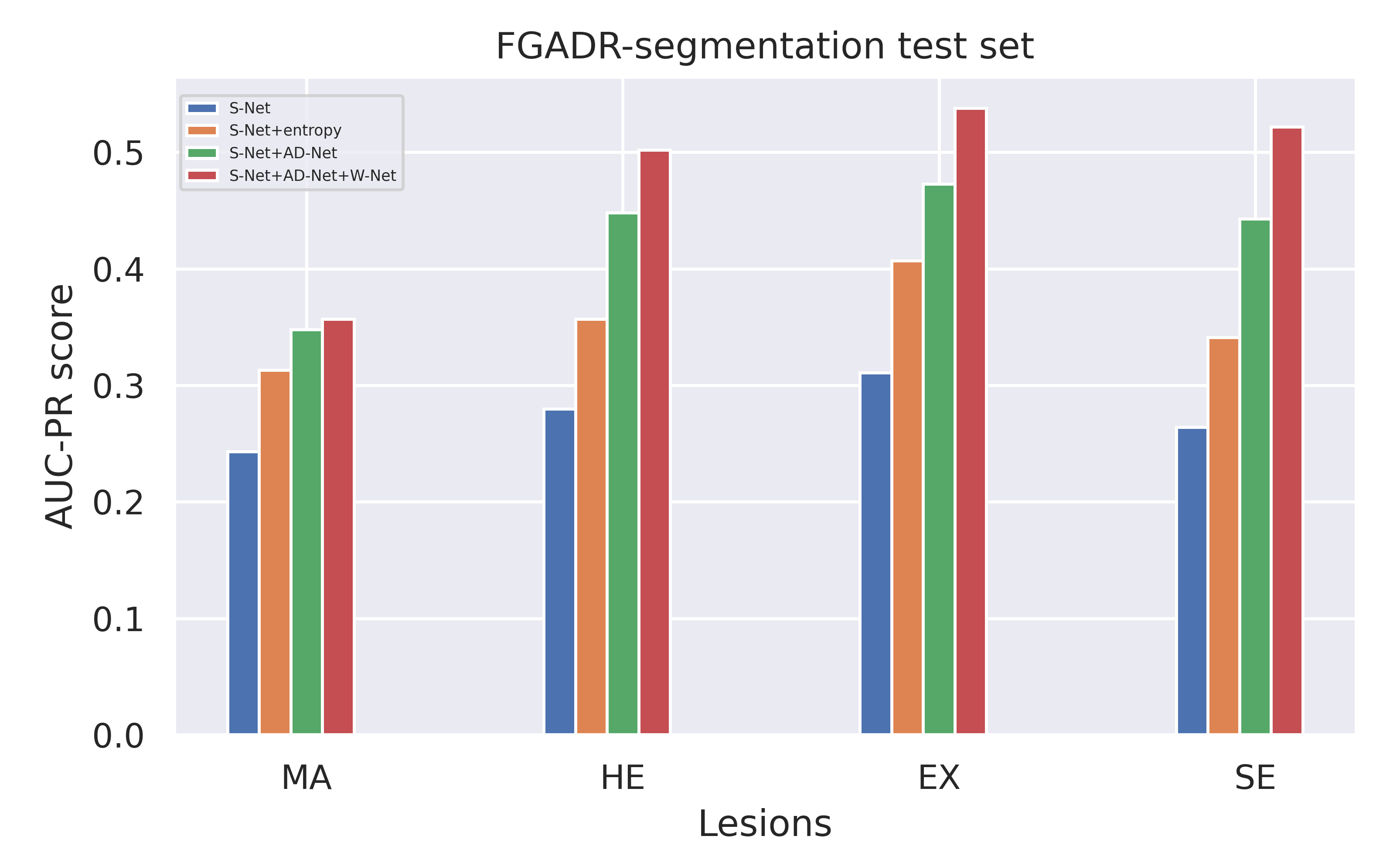}
\caption{Comparison of our different domain adaption approaches for lesion segmentation on four types of lesions. Results are evaluated on the setting \textit{IDRID $\rightarrow$ FGADR} with $0\%$ in target domain.}
\label{fig:fgadr_performance}
\end{figure}

\begin{figure*}[ht]
\centering
\includegraphics[width=1.7\columnwidth]{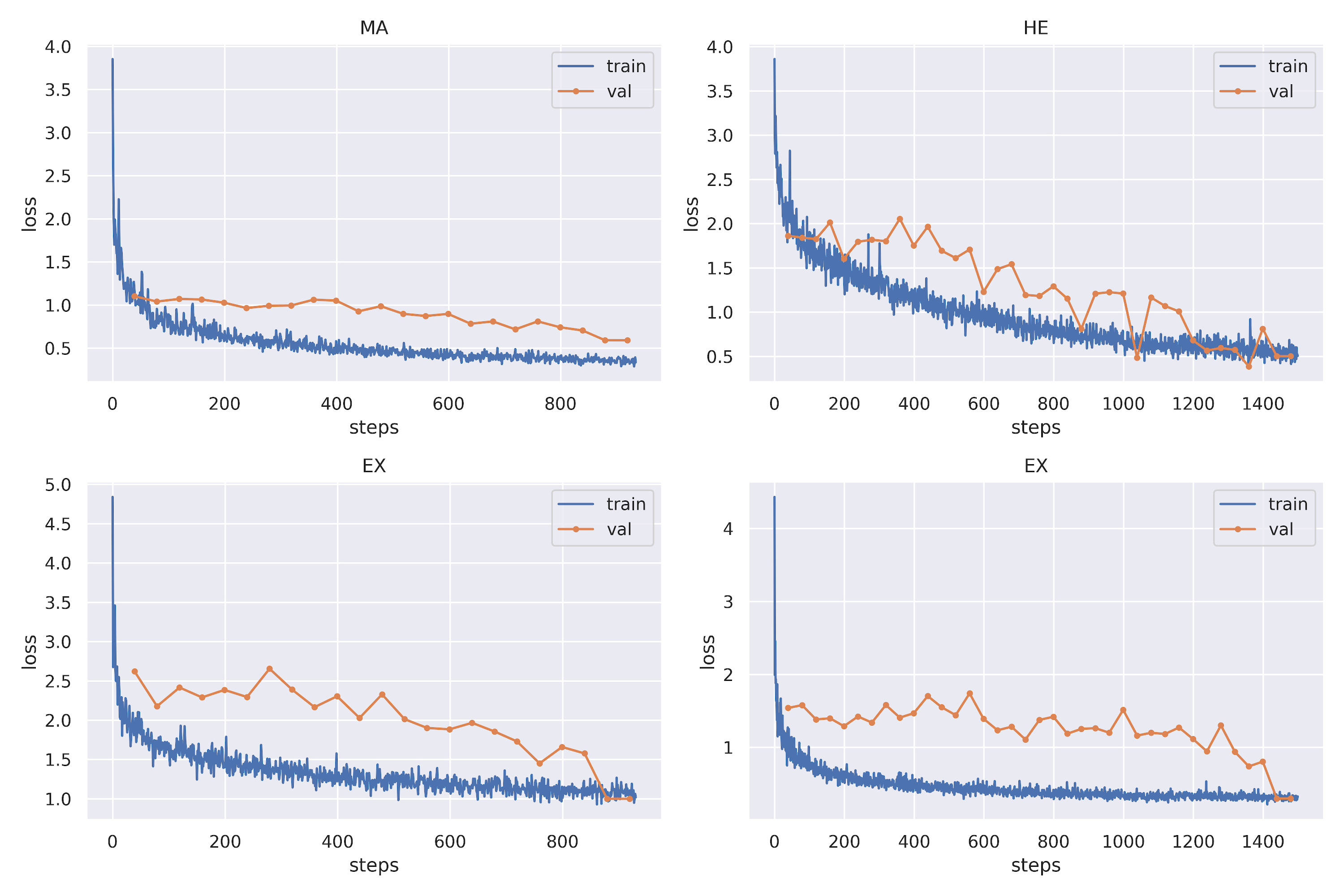}\\
    \caption{Training and validation loss curves for our domain adaptive lesion segmentation models. Results are on the setting \textit{IDRID $\rightarrow$ FGADR} with $0\%$ labeled data in the target domain. It can be seen that all training and validation curves tend to converge to stable points given more training steps.}
    \label{fig:fgadr_loss}
\end{figure*}

Plots in figure \ref{fig:fgadr_performance} compare the improvement in AUC-PR scores on the lesion classes of \textit{FGADR segmentation} test set for different domain adaptation settings. In summary, we observe that all of our domain adaptive settings contribute for significantly improving performance over non-adaptive settings \texttt{S-Net}. Besides, it can be seen in figure \ref{fig:fgadr_loss} that the training and validation loss curves of all lesion classes for the setting \texttt{S-Net + AD-Net + W-Net} using 0\% labeled data from the FGADR monotonously decline, confirming the stability in the learning process.

\begin{figure*}[ht]
    \centering
    \subfloat[Microaneurysms (MA)]{\includegraphics[width =3.5in]{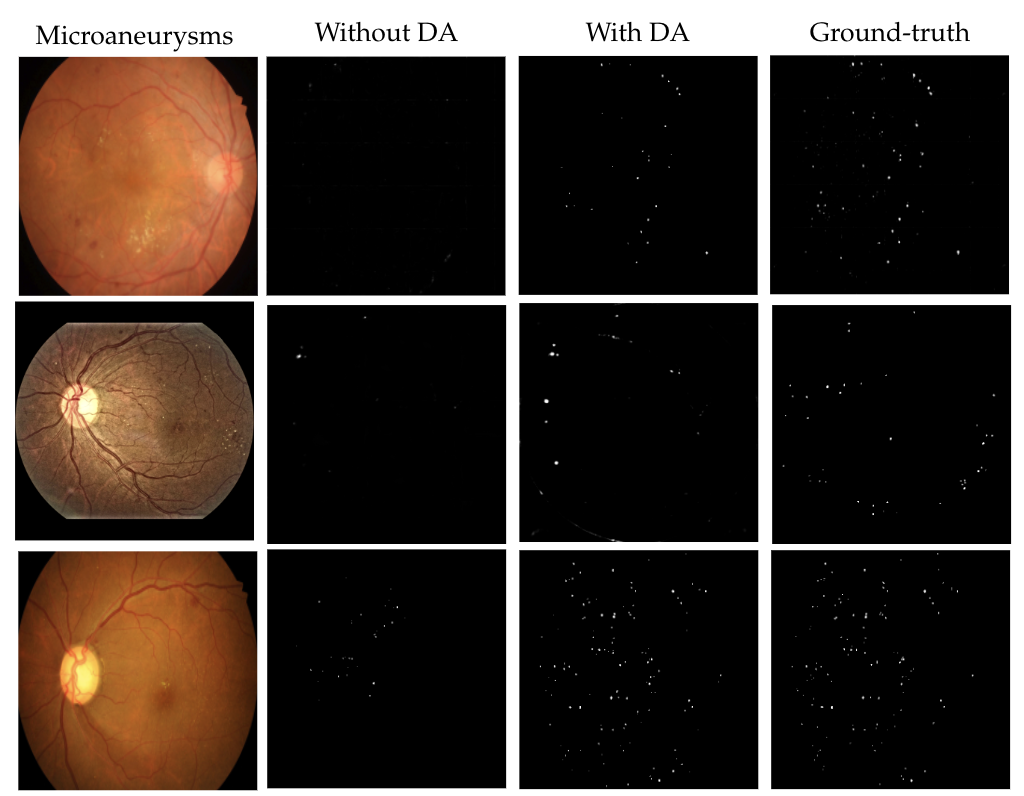}}%\hfill%
    \hspace{0.05cm}
    \subfloat[Haemorrhages (HE)]{\includegraphics[width =3.5in]{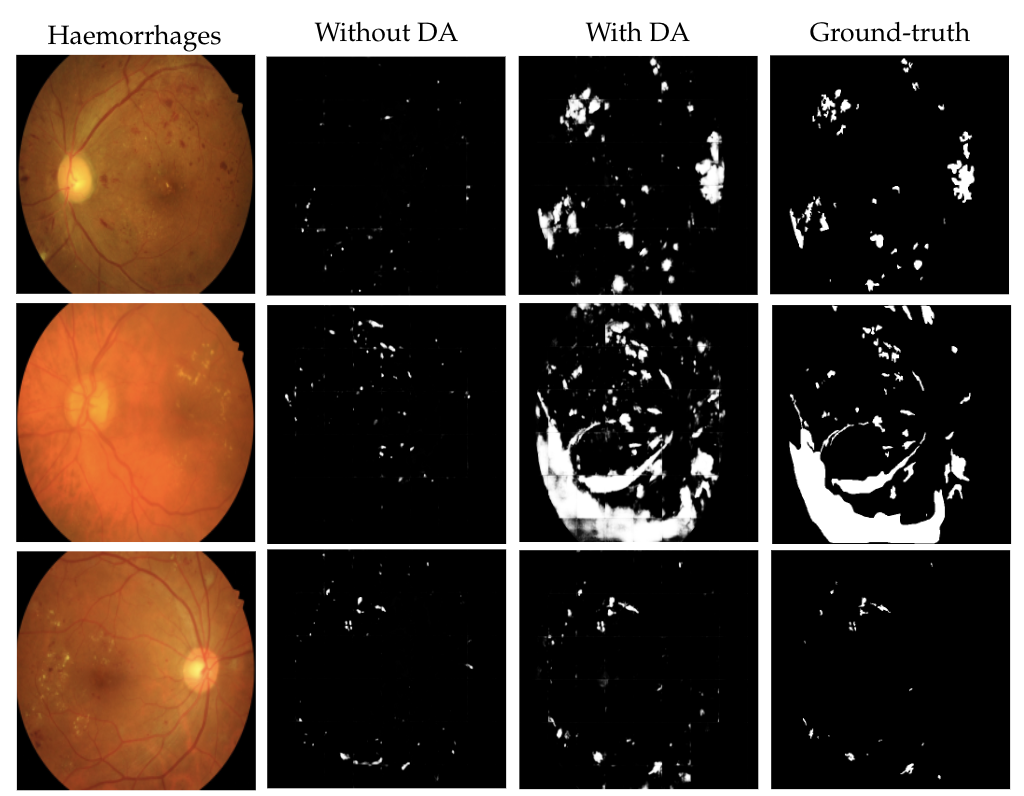}}  \\
    \hspace{0.05cm}
    \subfloat[Hard Exudates (EX)]{\includegraphics[width =3.5in]{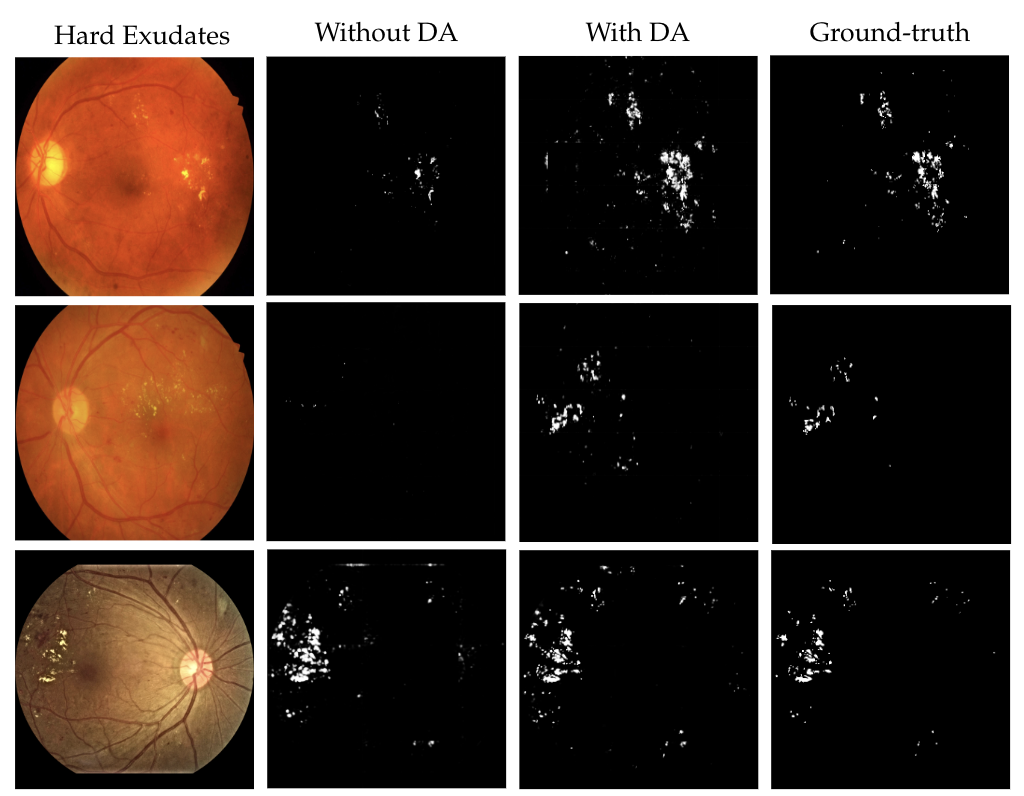}}%\hfill%
    \hspace{0.05cm}
    \subfloat[Soft Exudates (SE)]{\includegraphics[width =3.5in]{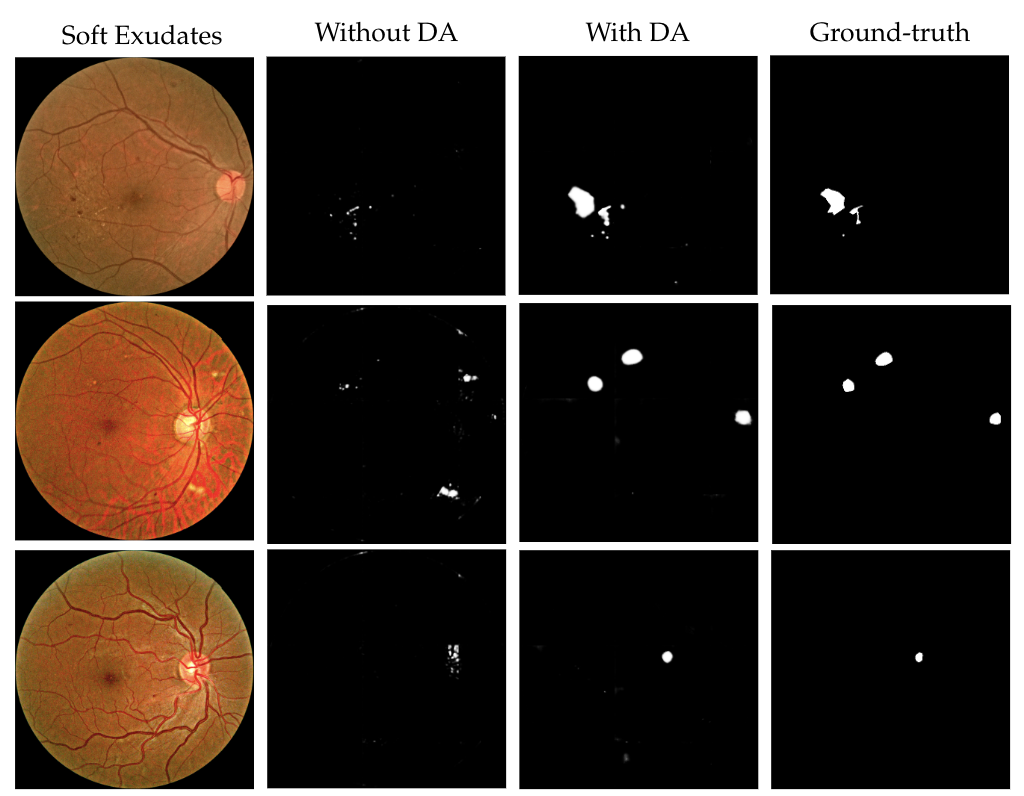}} \\
    \caption{Qualitative multi-lesion segmentation results for four different lesion classes. Comparison of lesion maps predicted by the segmentation models trained without (with) Domain Adaption (DA) \texttt{S-Net (S-Net + AD-Net + W-Net)}.}
    \label{fig:segmentation_images}
\end{figure*}

In table \ref{table:idrid_to_fgadr}, we compare our method with several state-of-the-art techniques: Fully convolutional segmentation \textit{(FCN-8s)} \cite{long2015fully}, Deep Lap model \textit{(DL\-V3+)}\cite{chen2017deeplab}, U-shaped models \textit{(U-Net)} \cite{ronneberger2015unet}, and \textit{Attention U-Net} \cite{islam2019brain}) using the results reported in \cite{zhou2020benchmark}. %The results for those baselines are taken from \cite{zhou2020benchmark}. 
The training are conducted directly on the target domain \textit{FGADR} using 100\% training labels. For a detailed comparison, we trained our domain adaptive model along with different percentages of labeled data from target domain. Results in green indicate the settings where we surpassed one or more methods under comparison while using $40\% \rightarrow 60\%$ labeled data. Results in bold indicate the best scores in all settings. In general, we observe that, for all of the lesion classes, by using only $40\%-60\%$ of labeled data in the target domain, we are able to outperform more than one baselines and with $80\%-100\%$ labeled data, we derive better AUC-ROC and AUC-PR scores compared with other methods in most cases. For instance, we gained 1.6\%, 0.5\%, 0.6\% and 0.4\% AUC-ROC score improvement on \textit{MA, HE, EX,} and \textit{SE} lesion classes respectively. The AUC-PR scores also improved for \textit{MA, EX and SE} by 2.7\%, 0.8\% and 1.2\% respectively. Figure \ref{fig:segmentation_images} illustrates qualitative comparisons of lesion map predictions between \texttt{S-Net} settings and our proposed multi-lesion segmentation model trained with domain adaption constraints \texttt{S-Net + AD-Net + W-Net}.

\paragraph{Results on FGADR $\rightarrow$ IDRID:} Table \ref{table:fgadr_to_idrid} shows the results on four lesion classes on \textit{IDRID-segmentation} test set where the segmentation model is trained using the labeled data from \textit{FGADR-segmentation} set and unlabeled images from \textit{IDRID-segmentation} training dataset. For different domain adaptation settings, we discover similar trends as the \textit{IDRID $\rightarrow$ FGADR} case. In particular, we compare our method with the following state-of-the-art techniques: adversarial learning based segmentation networks \textit{Adv. HEDNet}  \cite{Xiao_2019}, \textit{AdvSeg} \cite{hung2018adversarial} and semi-supervised collaborative learning networks \textit{ASDNet} and \textit{CoLL} \cite{zhou2019collaborative}. The training is conducted with $100\%$ labeled instances in the target domain (IDRID). With $0\%$ of label data, the configuration \texttt{S-Net + AD-Net + W-Net} already has comparable performance with the first baseline \textit{Adv. HEDNet}. When using $40\%$ labeled data, we are able to outperform most of competitive baselines in both AUC-ROC and AUC-PR metrics. This shows that, the proposed lesion generator model can be fine-tuned with minimal annotation data while still attains good performance in the target domain. 

\subsection{Performance of Diabetic Grading Tasks with Attentive Lesion Information}
\subsubsection{Influence of Attention Mechanism on Grading Networks}
We evaluated the effectiveness of predicted lesion segmentation maps in disease grading classification tasks.  Our evaluation is on two types of vision related foundation models: CNN-based architecture and Vision Transformer-based architecture. We conducted experiments with the following settings.

\begin{enumerate}

    \item \texttt{G-Net}: In this baseline, we directly train the disease grading classification network using retina fundus images and their associated disease grading labels using the classification loss $\mathcal{L}_{cls}$ in equation (\ref{eq:classification}). In this step only \textit{G-Net} is trained and no`attention functionality is used, therefore, we do not use any lesion information in the training process. 
    
    \item \texttt{G-Net + Attention (low-level)}: We integrate an attention model \textit{Att-Net} to highlight generated lesion features in the training process of the grading network \textit{G-Net}. The attention mechanism we integrated here is on structural feature level of the grading network described in section \ref{subsec:lesion-cnn}. Here, the \textit{G-Net} and \textit{Att-Net} are jointly trained using equation (\ref{eq:classification_attention}).
    
    \item \texttt{G-Net + Attention (high-level)}: In this setting, we integrate the attention mechanism for lesion features following our high-level concept described in section \ref{subsec:lesion-high-level}. Unlike the previous low-level concept, this attention constraint is more intuitive where the normalized class activation maps are directly compared with the lesion feature maps to compute the attention overlapping loss $\mathcal{L}_{overlap}$ in equation (\ref{eq:overlap}). The \textit{G-Net} and \textit{Att-Net} then are jointly trained for classification.
    
    \item \texttt{G-Net + Attention (two-level):} We combine both of the proposed \textit{low-level} and \textit{high-level} attention concept in the training process of the disease grading network \textit{G-Net}. This is our final proposed method for the lesion attentive DR grading model. The \textit{G-Net} and \textit{Att-Net} are jointly optimized using the equation (\ref{eq:grading_final}).
    
\end{enumerate}

\begin{table*}[!t]
\centering
% \captionsetup{skip=4pt}
% \vspace*{1 cm}
\caption{Performance of the proposed attention mechanisms on the CNN-based architecture (ResNet-50). Results are evaluated on three datasets \textit{IDRID}, \textit{EyePACS}, and \textit{FGADR}.}% lesion attentive disease grading on CNN based disease grading classification model \textbf{G-Net (CNN)}. Evaluation Results on \textbf{IDRID}, \textbf{EyePACS} and \textbf{FGADR} datasets. All of the "G-Net" in this tables indicates CNN based architecture is used for prediction.}
\label{table:grading_cnn}
\scalebox{0.88}{
\begin{tabular}{lcccccc}
% \toprule
\hline
\multirow{2}{*}{Methods} & \multicolumn{2}{c}{IDRID} & \multicolumn{2}{c}{EyePACS} & \multicolumn{2}{c}{FGADR} \\ \cline{2-7}
  & Accuracy  & Kappa & Accuracy  & Kappa  & Accuracy  & Kappa \\ \hline
G-Net (ResNet-50)                                                              & 0.823              & 0.844          & 0.836               & 0.819           & 0.807             & 0.751          \\
G-Net + Attention (low-level)                                            & 0.897              & 0.904          & 0.881               & 0.867           & 0.860             & 0.852          \\
\begin{tabular}[c]{@{}c@{}}G-Net + Attention (high-level)\end{tabular} & 0.859              & 0.873          & 0.862               & 0.849           & 0.839             & 0.827          \\
G-Net + Attention (two-level)                                   & \textbf{0.904}     & \textbf{0.911} & \textbf{0.895}      & \textbf{0.883}  & \textbf{0.872}    & \textbf{0.861} \\ %\bottomrule
\hline
\end{tabular}}
\end{table*}
\begin{table*}[!t]
\centering
\caption{Performance of the proposed attention mechanisms on the Vision Transformer architecture (MIL-VIT). Results are evaluated on three datasets \textit{IDRID}, \textit{EyePACS}, and \textit{FGADR}
% Evaluation of effectiveness of lesion attentive disease grading on Transformer-based disease grading classification model \textbf{G-Net (MIL-ViT)}. Evaluation Results on \textbf{IDRID}, \textbf{EyePACS} and \textbf{FGADR} datasets. All of the "G-Net" in this tables indicates Vision Transformer based architecture is used for prediction.
. Because MIL-VIT already integrated attentions at the feature-level, we only extend it with the high-level case.}
\label{table:grading_vit}
% \captionsetup{skip=4pt}
% \vspace*{1 cm}
\scalebox{0.88}{
\begin{tabular}{lcccccc}
% \toprule
\hline
\multirow{2}{*}{Methods} & \multicolumn{2}{c}{IDRID} & \multicolumn{2}{c}{EyePACS} & \multicolumn{2}{c}{FGADR} \\ \cline{2-7}
  & Accuracy  & Kappa & Accuracy  & Kappa  & Accuracy  & Kappa \\ \hline
{\color[HTML]{202124} G-Net (MIL-VIT)}                        & {\color[HTML]{202124} 0.780}          & {\color[HTML]{202124} 0.824}          & {\color[HTML]{202124} 0.755}          & {\color[HTML]{202124} 0.791}          & {\color[HTML]{202124} 0.751}          & {\color[HTML]{202124} 0.781}          \\
{\color[HTML]{202124} G-Net + Attention (high-level)} & {\color[HTML]{202124} \textbf{0.822}} & {\color[HTML]{202124} \textbf{0.841}} & {\color[HTML]{202124} \textbf{0.819}} & {\color[HTML]{202124} \textbf{0.849}} & {\color[HTML]{202124} \textbf{0.797}} & {\color[HTML]{202124} \textbf{0.826}} \\ %\bottomrule
\hline
\end{tabular}}
\end{table*}

\begin{table*}[!t]
\centering
\caption{Comparison of our lesion attentive disease grading models  with other \textit{state-of-the-art} techniques. These methods are also used lesion information in solving the DR grading task.
% Last two methods indicates the results for our proposed method. Our CNN-based method \textbf{G-Net (CNN) + Attention} has achieve comparable performance on IDRID and EyePACS datasets and outperforms other methods on \textbf{FGADR} grading dataset in both \textit{accuracy} and \textit{kappa} metrics. 
}
\label{table:grading_comparison}
% \captionsetup{skip=4pt}
\scalebox{0.85}{
\begin{tabular}{lcccccc}
% \toprule
\hline
\multirow{2}{*}{Methods} & \multicolumn{2}{c}{IDRID} & \multicolumn{2}{c}{EyePACS} & \multicolumn{2}{c}{FGADR} \\ \cline{2-7}
  & Accuracy  & Kappa & Accuracy  & Kappa  & Accuracy  & Kappa \\ \hline
{\color[HTML]{202124} JCS \cite{Wu_2021}}                                & {\color[HTML]{202124} -}              & {\color[HTML]{202124} -}              & {\color[HTML]{202124} 0.886}          & {\color[HTML]{202124} 0.877}          & {\color[HTML]{202124} 0.856}          & {\color[HTML]{202124} 0.842}          \\
{\color[HTML]{202124} AFN-Net \cite{zhou2020benchmark}}                            & {\color[HTML]{202124} -}              & {\color[HTML]{202124} -}              & {\color[HTML]{202124} 0.861}          & {\color[HTML]{202124} 0.856}          & {\color[HTML]{202124} 0.836}          & {\color[HTML]{202124} 0.784}          \\
% {\color[HTML]{202124} LAT \cite{sun2021lesion}}                                & {\color[HTML]{202124} -}              & {\color[HTML]{202124} -}              & {\color[HTML]{202124} -}              & {\color[HTML]{202124} \textbf{0.884}} & {\color[HTML]{202124} -}              & {\color[HTML]{202124} -}              \\
{\color[HTML]{202124} CoLL \cite{zhou2019collaborative}}                               & {\color[HTML]{202124} \textbf{0.913}} & {\color[HTML]{202124} 0.904} & {\color[HTML]{202124} 0.891}          & {\color[HTML]{202124} 0.872}          & {\color[HTML]{202124} 0.86}           & {\color[HTML]{202124} 0.848}          \\ \hline
{\color[HTML]{202124} G-Net (MIL-ViT) + Attention} & {\color[HTML]{202124} 0.822}          & {\color[HTML]{202124} 0.841}          & {\color[HTML]{202124} 0.819}          & {\color[HTML]{202124} 0.849}          & {\color[HTML]{202124} 0.797}          & {\color[HTML]{202124} 0.826}          \\
{\color[HTML]{202124} G-Net (ResNet-50) + Attention}       & {\color[HTML]{202124} 0.904}          & {\color[HTML]{202124} \textbf{0.911}}          & {\color[HTML]{202124} \textbf{0.895}} & {\color[HTML]{202124} \textbf{0.883}}          & {\color[HTML]{202124} \textbf{0.872}} & {\color[HTML]{202124} \textbf{0.861}} \\ %\bottomrule
\hline
\end{tabular}}
\end{table*}

\begin{figure*}[ht]
\centering
\subfloat[accuracy (FGADR)]{\includegraphics[width = 3.0in]{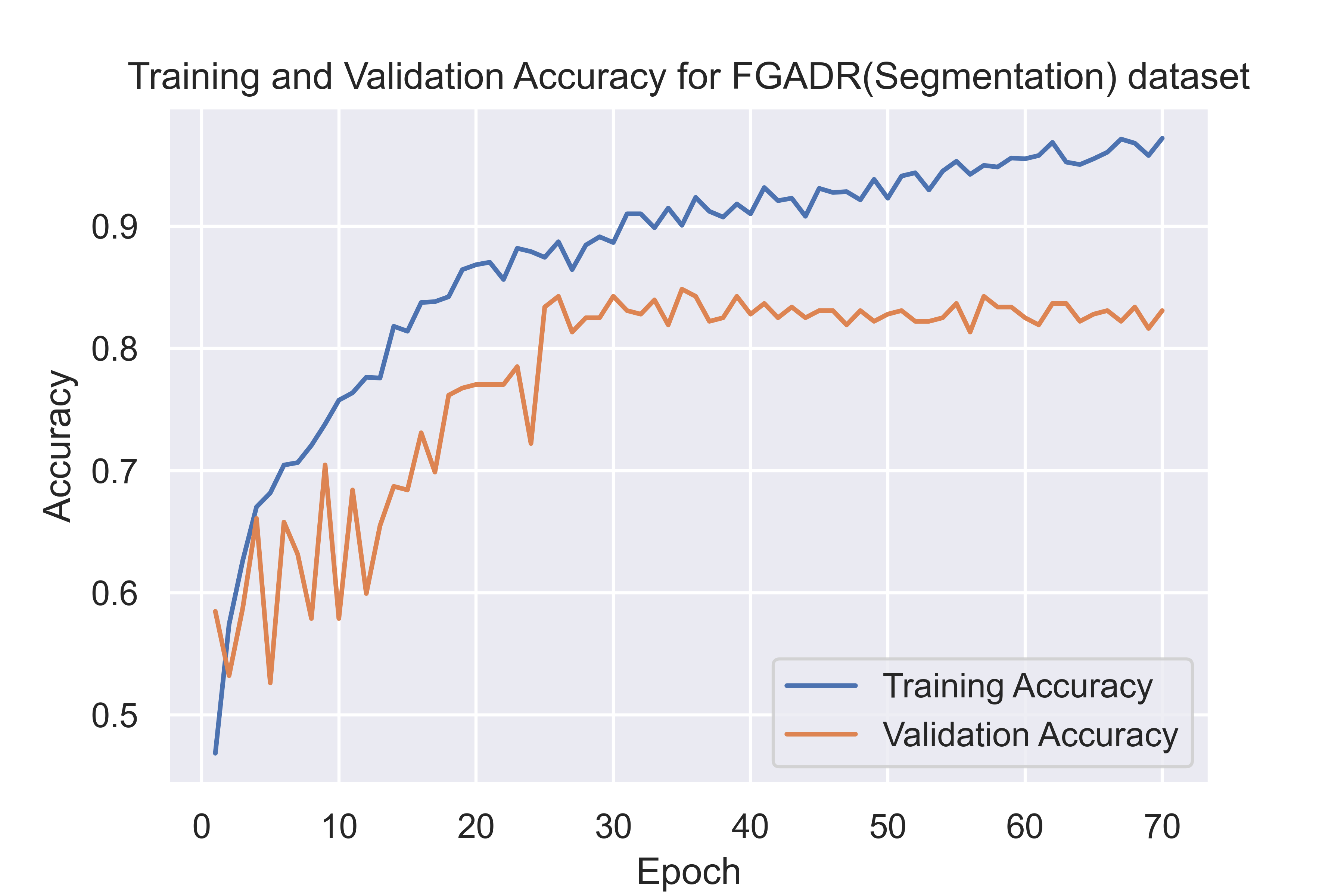}}\hspace{-2em}
\subfloat[kappa (FGADR)]{\includegraphics[width = 3.0in]{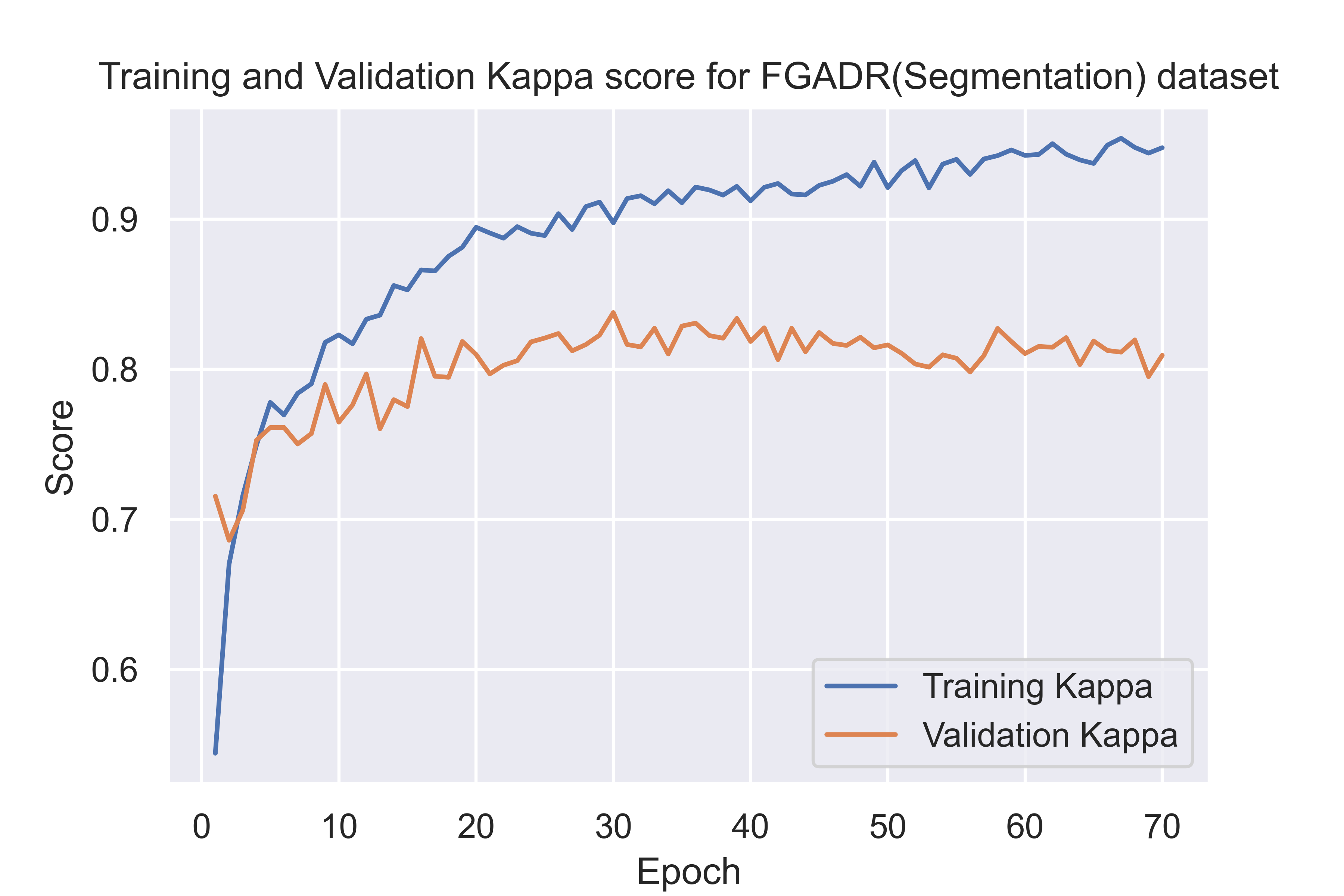}}\\
\subfloat[accuracy (EyePACS)]{\includegraphics[width = 3.0in]{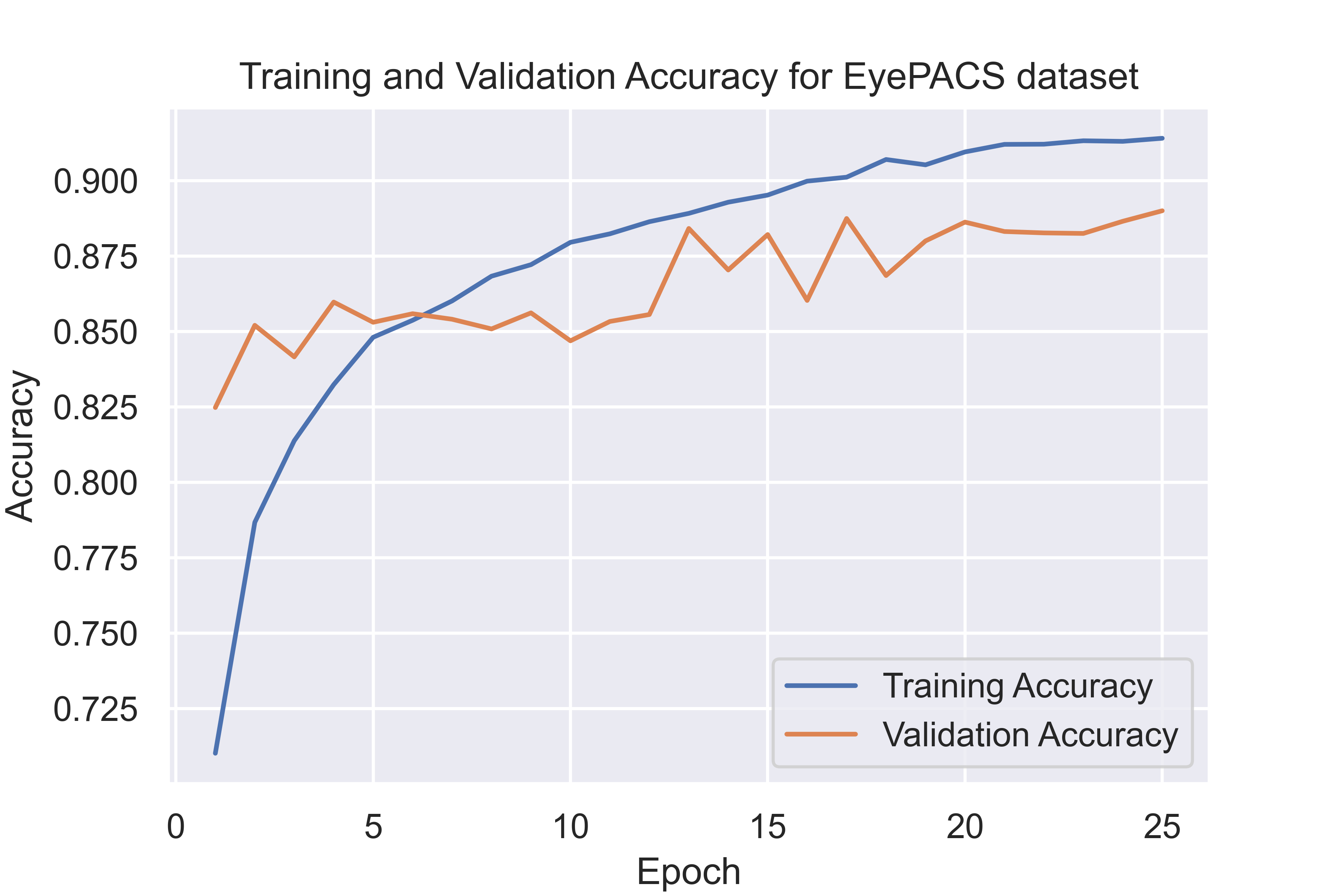}}\hspace{-2em}
\subfloat[kappa (EyePACS)]{\includegraphics[width = 3.0in]{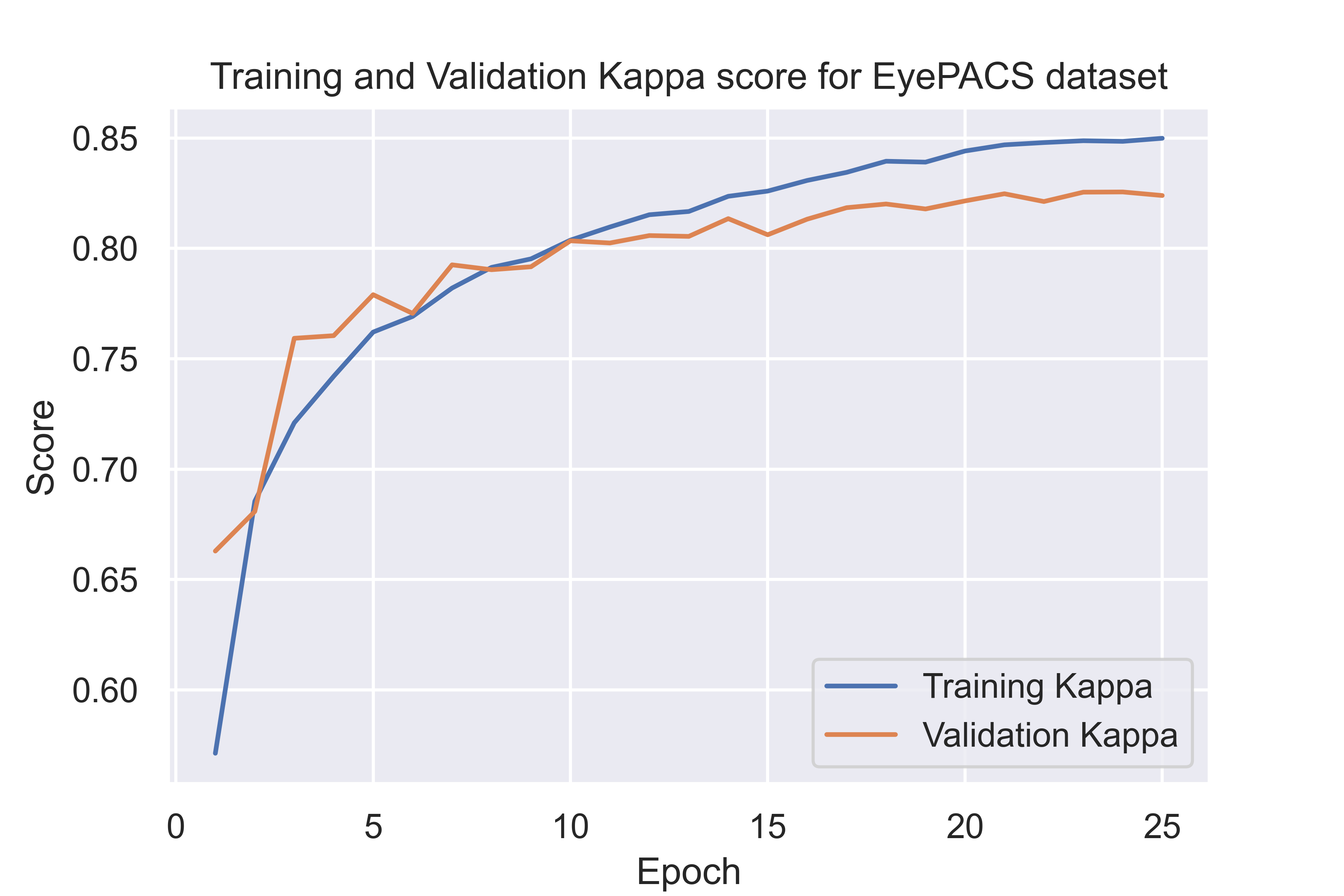}}
% \subfloat[fig 3]{\includegraphics[width = 3in]{something}}
% \subfloat[fig 4]{\includegraphics[width = 3in]{something}} 
\caption{Illustration our training and validation classification accuracy and kappa measurement using \texttt{G-Net + Attention (two levels)}. Plots in the top row (a) and (b) are the accuracy and kappa score curve in \textit{FGADR} dataset. Plots in the bottom row (c) and (d) are results in \textit{EyePACS} dataset.}
\label{fig:accuray_curve_grading}
\end{figure*}

For the CNN-based architecture, we evaluated all four of our proposed settings using \textit{ResNet-50} for the \textit{G-Net} model. For transformer-based architecture, we considered multi-head vision transformer \textit{(MIL-Vit)} \cite{yu2021mil} as \textit{G-Net}. Transformer models are inherently built on attention modules at the structural level (low-level); therefore, we only test \texttt{G-Net} and \texttt{G-Net + Attention (high-level)} methods for the transformer architecture. 

We tested our approach on the \textit{IDRID}, \textit{EyePACS} and \textit{FGADR} DR grading datasets. Table \ref{table:grading_cnn} and \ref{table:grading_vit} present the evaluation results for different settings on these datasets for CNN-based and Transformer-based models respectively. For CNN-based models in table \ref{table:grading_cnn}, we observe that both low-level and high-level attention mechanisms improved the performance for the original classification network \texttt{G-Net}. Integrating low-level attention with grading network in settings \texttt{G-Net + Attention (low-level)}, we could increase in Kappa scores by 6\%, 5\% and 10\% for IDRID, EyePACS and FGADR datasets respectively. For the \texttt{G-Net + Attention (high-level)}, the kappa scores are improved by 3\%, 3\% and 7\% respectively. We also gained best performance (in bold) for each of the datasets when combining both low-level and high-level attention mechanisms for the setting \texttt{G-Net + Attention (two-level)}. Figure \ref{fig:accuray_curve_grading} describes the comparison between training and validation accuracy in terms of kappa scores for the \textit{FGADR} and \textit{EyePACS} data. We can see that the networks tend to achieve higher performance given more epochs.

Table \ref{table:grading_vit} presents the results for transformer-based model. We can see that, integrating high level attention mechanism with \textit{G-Net} has improved both classification accuracy and kappa scores for all three grading datasets. In table \ref{table:grading_comparison}, we compared our best lesion attentive disease grading models (one each for CNN-based and Transformer-based methods) with recent competitive methods: \textit{JCS} (\cite{Wu_2021}), Denoising Attention Fusion Network (\textit{AFN-Net}) \cite{zhou2020benchmark}) and Collaborative DR grading (\textit{CoLL}) \cite{zhou2019collaborative}. Results of our CNN-based attention model \texttt{G-Net (CNN) + Attention} on \textit{IDRID} is comparable with the other methods. For EyePACS and FGADR dataset, we outperform other methods by around 2\% in accuracy and kappa scores.

\subsubsection{DR Grading Prediction with Explainable Property}
\label{subsec:dr-explain}
To construct an explainable intelligent decision support system for diabetic retinopathy diagnosis, in the inference step, along with the predicted class output, we compute the class activation map (CAM) for the input image. CAM highlights the discriminative regions for the predicted class based on which the deep neural model has taken its decision. For DR grading tasks, different lesion regions observed in retinal fundus image are the key clinical factors to diagnose the patient disease progression to a particular grading class. %The explanation decision of a model for it predictions needed to be consistent with the human expert reasoning. In terms of diabetic retinopathy diagnosis, class activation map (CAM) for the predicted grading class should be able to highlight relevant lesion regions. 
Figure \ref{fig:explantion_outputs} illustrates an example of the explainable prediction in our system. For a given fundus image, our domain invariant lesion generator model \textit{S-Net} predicts lesion maps for four different lesion classes (figure \ref{fig:explantion_outputs}\,(j) to \ref{fig:explantion_outputs}\,(l)). Then the attention-based grading model \textit{G-Net} provides the disease grade and its decision explanation in the form of class activation map (figure \ref{fig:explantion_outputs}\,(b)). By comparing the CAM region with the predicted lesion positions, we can compare the quality of the model explanation. Intuitively, the overlapping rate of CAM on lesion maps and input images can provide explanation properties to the experts about the network's decisions (figure \ref{fig:explantion_outputs}\,(c) and \ref{fig:explantion_outputs}\,(d)).

In figure \ref{fig:cam_comparision_update}, we qualitatively compare the explanation results between the grading model that is trained only using classification loss in \texttt{G-Net} settings and our proposed attention based grading model that incorporate both low-level and high-level attention concepts in settings \texttt{G-Net (CNN) + Attention (two-level)}. For some typical images and their lesion maps, we discover that our attention-based grading (figure \ref{fig:cam_comparision_update}\,(d)) was able to capture almost all the important lesion regions whereas \texttt{G-Net} (figure \ref{fig:cam_comparision_update}\,(c)) failed to capture these lesion regions and considered some irrelevant positions as the class discriminating mapping. In opposite, even some tiny lesion regions are also considered in the prediction decision of our grading model.

% \begin{figure}[!hbt]

%     \centering
%     \includegraphics[width=1.0\columnwidth]{images/experiment/interactive/grad_cam_compare.png}
%     \caption{Comparison on explainability of the disease classification model trained with and without lesion attention. (a) original inputs, (b) union of ground-truth lesion maps, (c) overlapping of class activation map (CAM) of the predicted grading class with the (a) and (b) for \textit{G-Net} trained without lesion attention model \textit{Att-Net}, (d) overlapping of class activation map (CAM) of the predicted grading class with the (a) and (b) for \textit{G-Net} trained with lesion attention model \textit{Att-Net}. We can observe that CAMs in column (d) overlaps with most of the lesion regions.}
%     \label{fig:cam_comparision}
% \end{figure}
\begin{figure*}[h!]
\centering
\includegraphics[width=1.5\columnwidth]{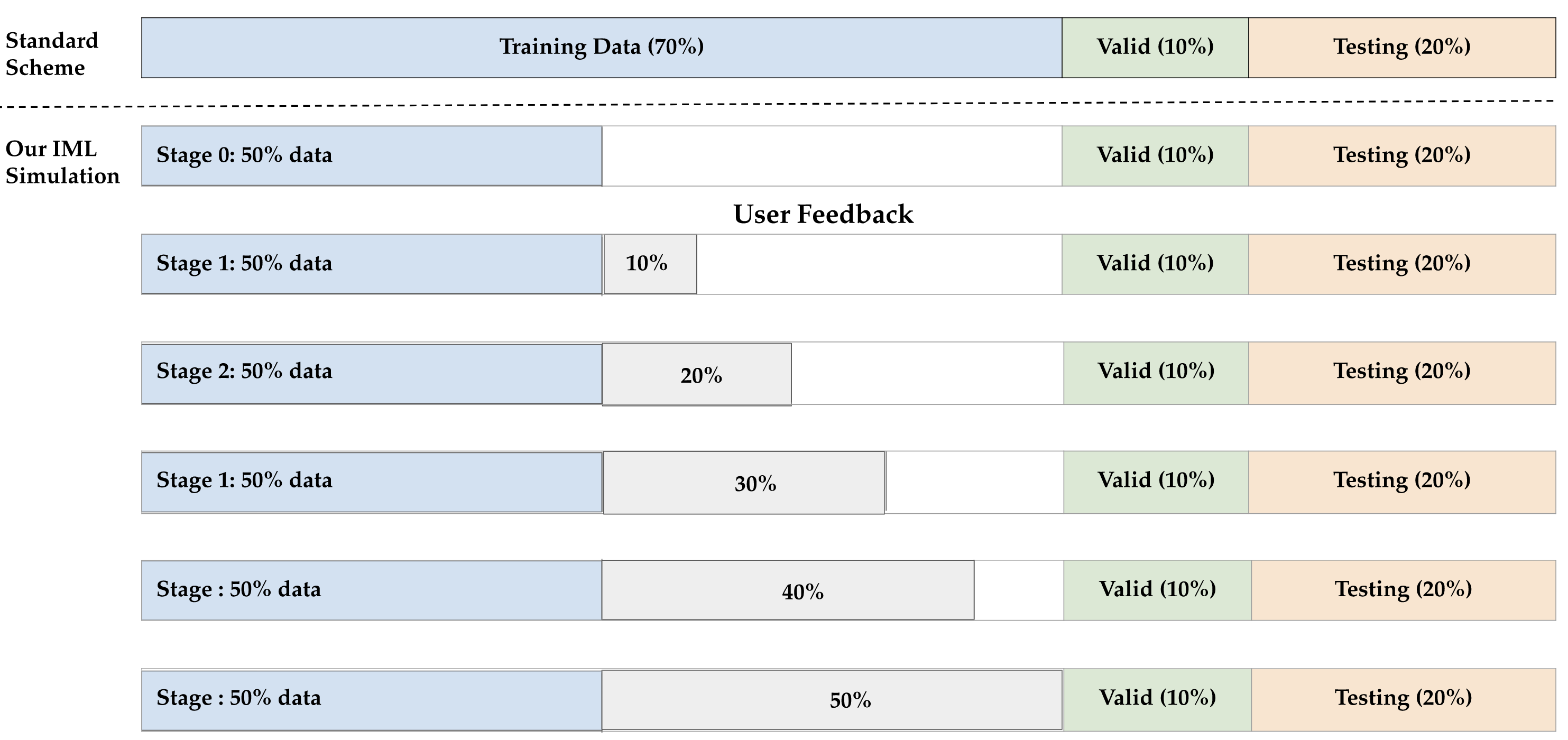}\\
\caption{Illustration of user-feedback simulation on slices of data. In a standard scheme, traditional data split strategy is followed. In our interactive simulation case, we initially split the training data into two equal parts, and used one part of the split (50\% of training data) to train our proposed attention-based grading model. For the remaining part of $50\%$ training data, it is further split  into five equal slices, serving for increasingly data feedback. After each time getting data feedback, the model is fine-tuned using both initial data and new ones. The performance of a new model then is evaluated on the hold-out testing set.}
\label{fig:user_simulation}
\end{figure*}

\begin{figure*}
    \centering
    \subfloat[original image]{\includegraphics[width =0.9in]{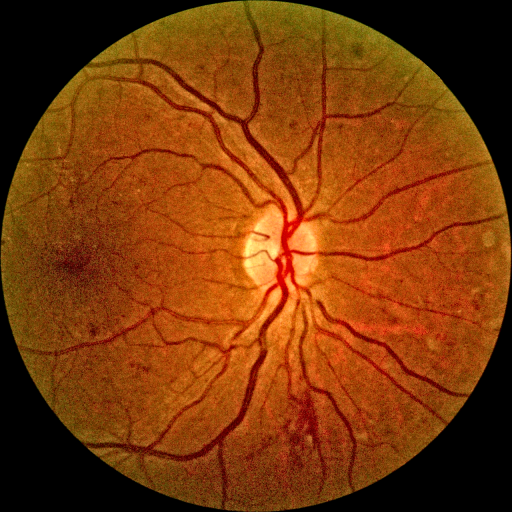}}\hspace{1em}
    % \hfill%
    % 
    \subfloat[CAM - \textit{grade-3}]{\includegraphics[width =0.9in]{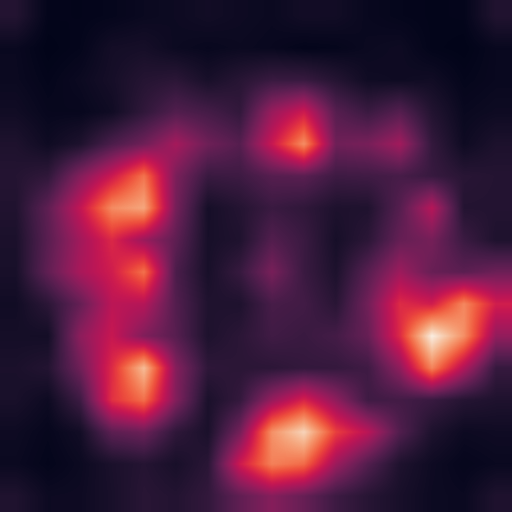}}\hspace{1em}
    % \hfill%
    % \hspace{0.05cm}
    \subfloat[CAM + lesion (union)]{\includegraphics[width = 0.9in]{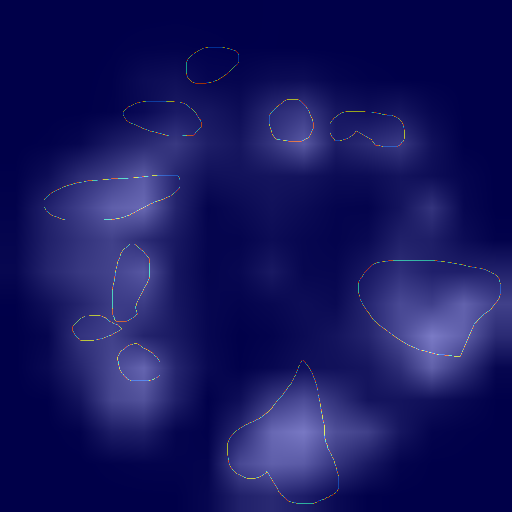}}\hspace{1em}
    % \hfill%
    % \hspace{0.05cm}
    \subfloat[CAM + lesion + image]{\includegraphics[width = 0.9in]{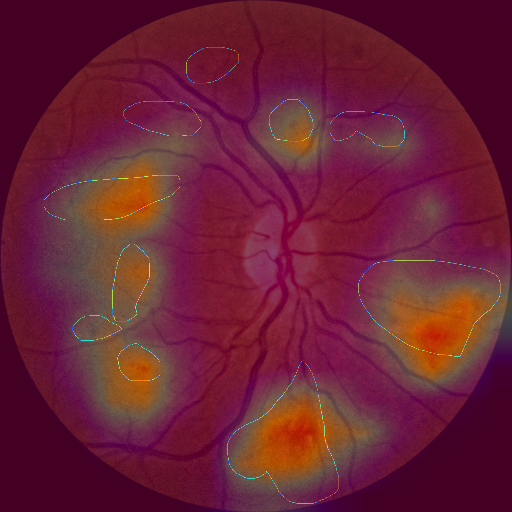}}\\ 
    \subfloat[CAM + MA]{\includegraphics[width =0.9in]{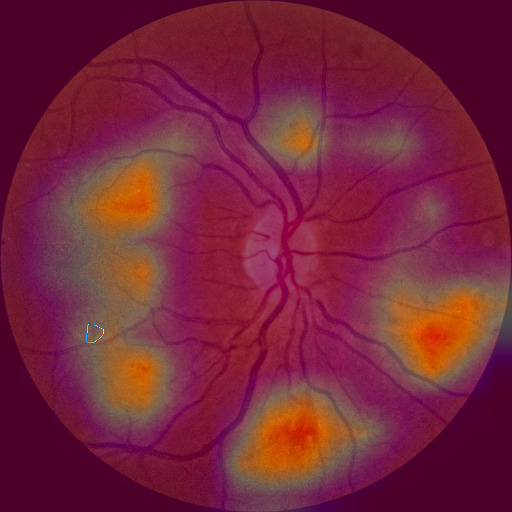}}\hspace{1em}
    %\hfill%
    % \hspace{0.05cm}
    \subfloat[CAM + HE]{\includegraphics[width =0.9in]{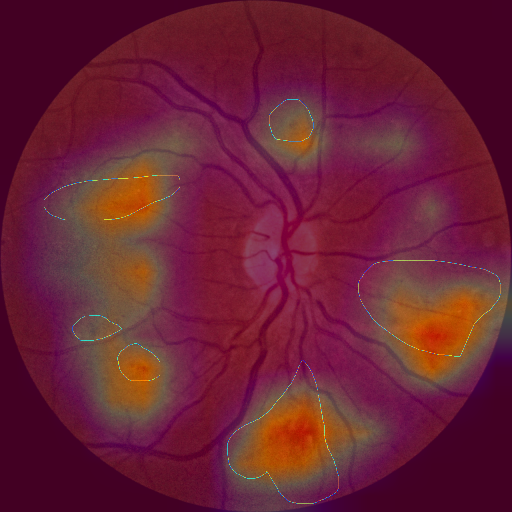}}\hspace{1em}
    % \hfill%
    % \hspace{0.05cm}
    \subfloat[CAM + EX]{\includegraphics[width = 0.9in]{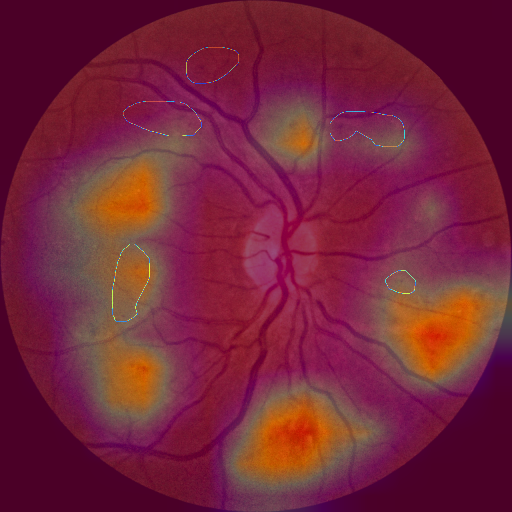}}\hspace{1em}
    %\hfill%
    % \hspace{0.05cm}
    \subfloat[CAM + SE]{\includegraphics[width = 0.9in]{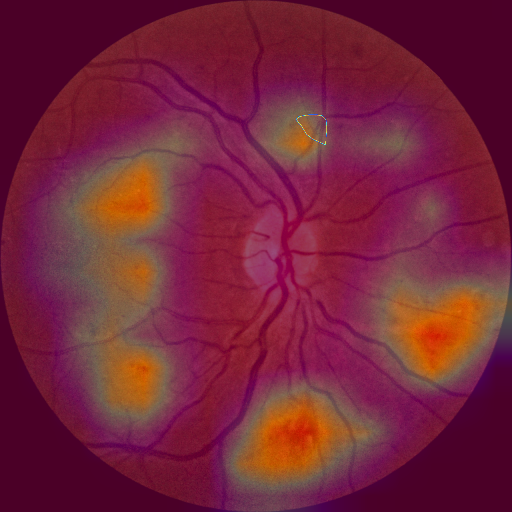}}\\
    \subfloat[Microaneurysms]{\includegraphics[width =0.9in]{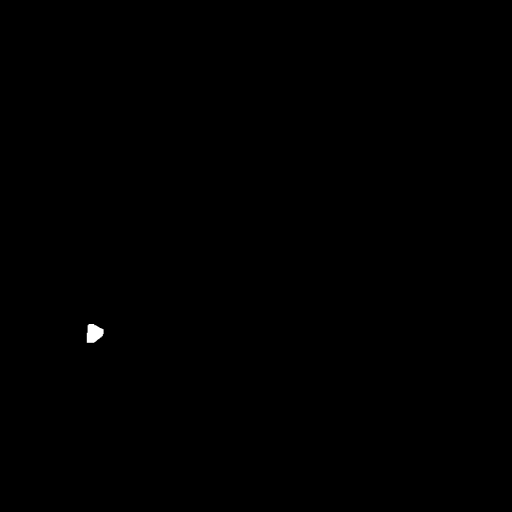}}\hspace{1em}
    %\hfill%
    % \hspace{0.05cm}
    \subfloat[Haemorrhages]{\includegraphics[width =0.9in]{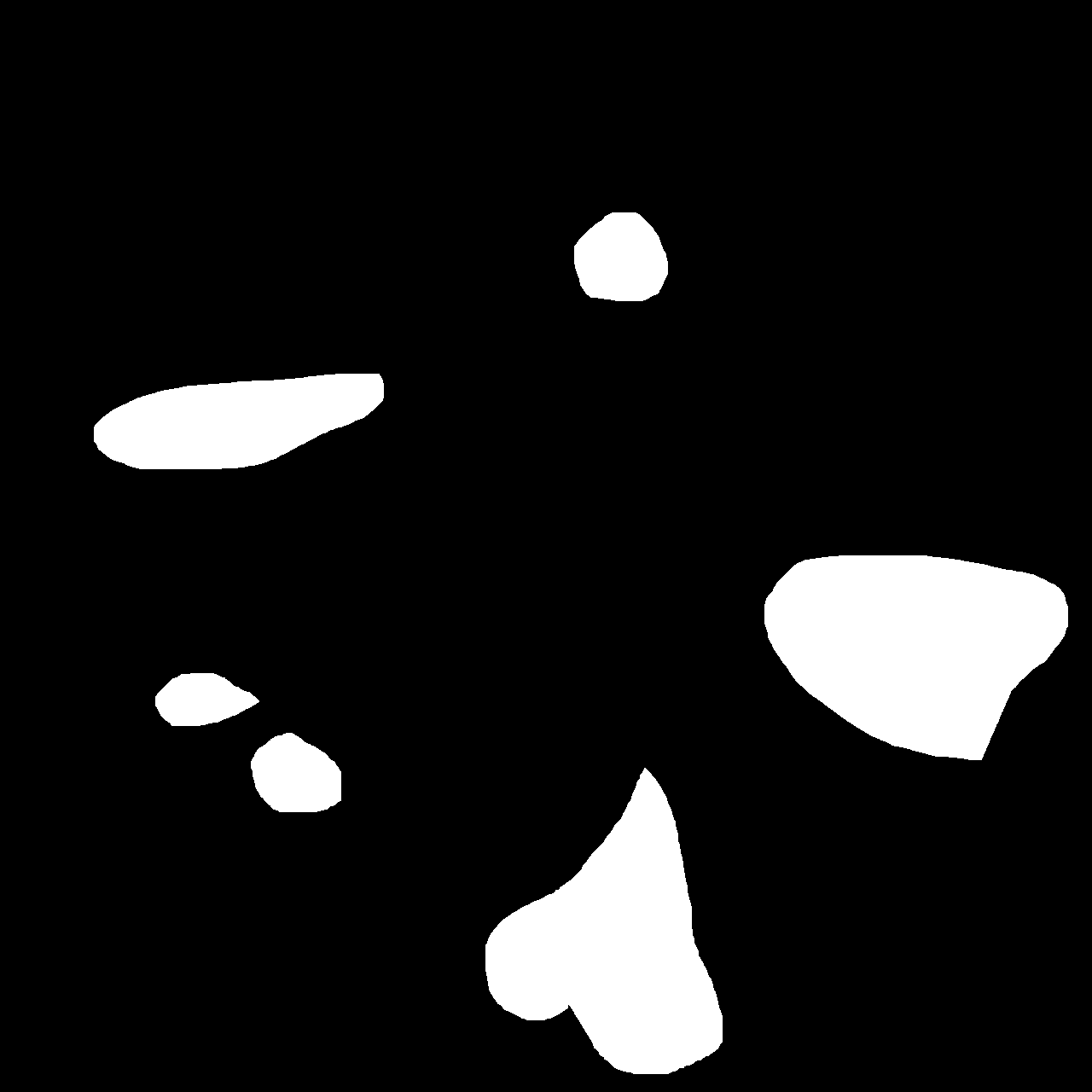}}\hspace{1em}
    %\hfill%
    % \hspace{0.05cm}
    \subfloat[Hard Exudates]{\includegraphics[width = 0.9in]{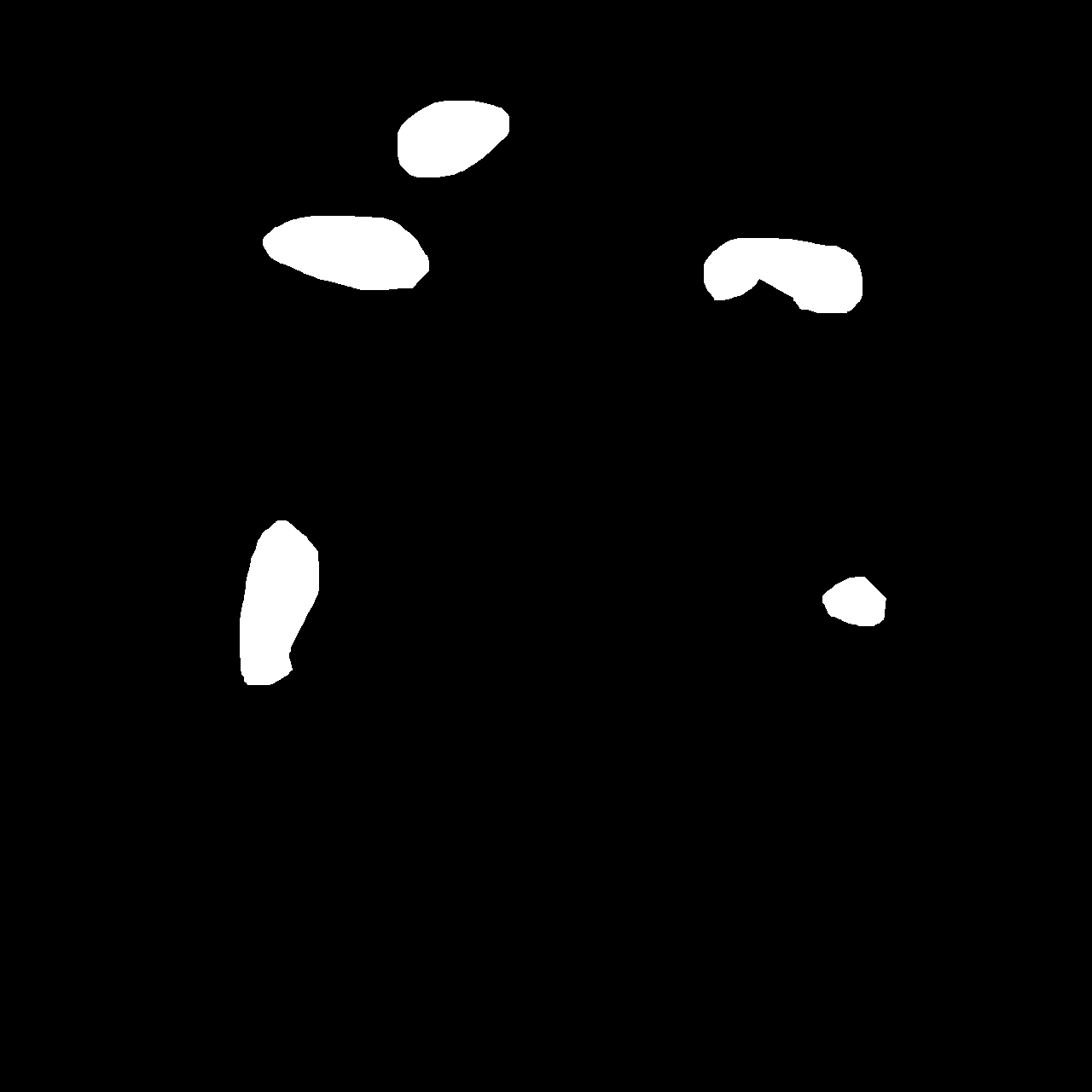}}\hspace{1em}
    % \hfill%
    % \hspace{0.05cm}
    \subfloat[Soft Exudates]{\includegraphics[width = 0.9in]{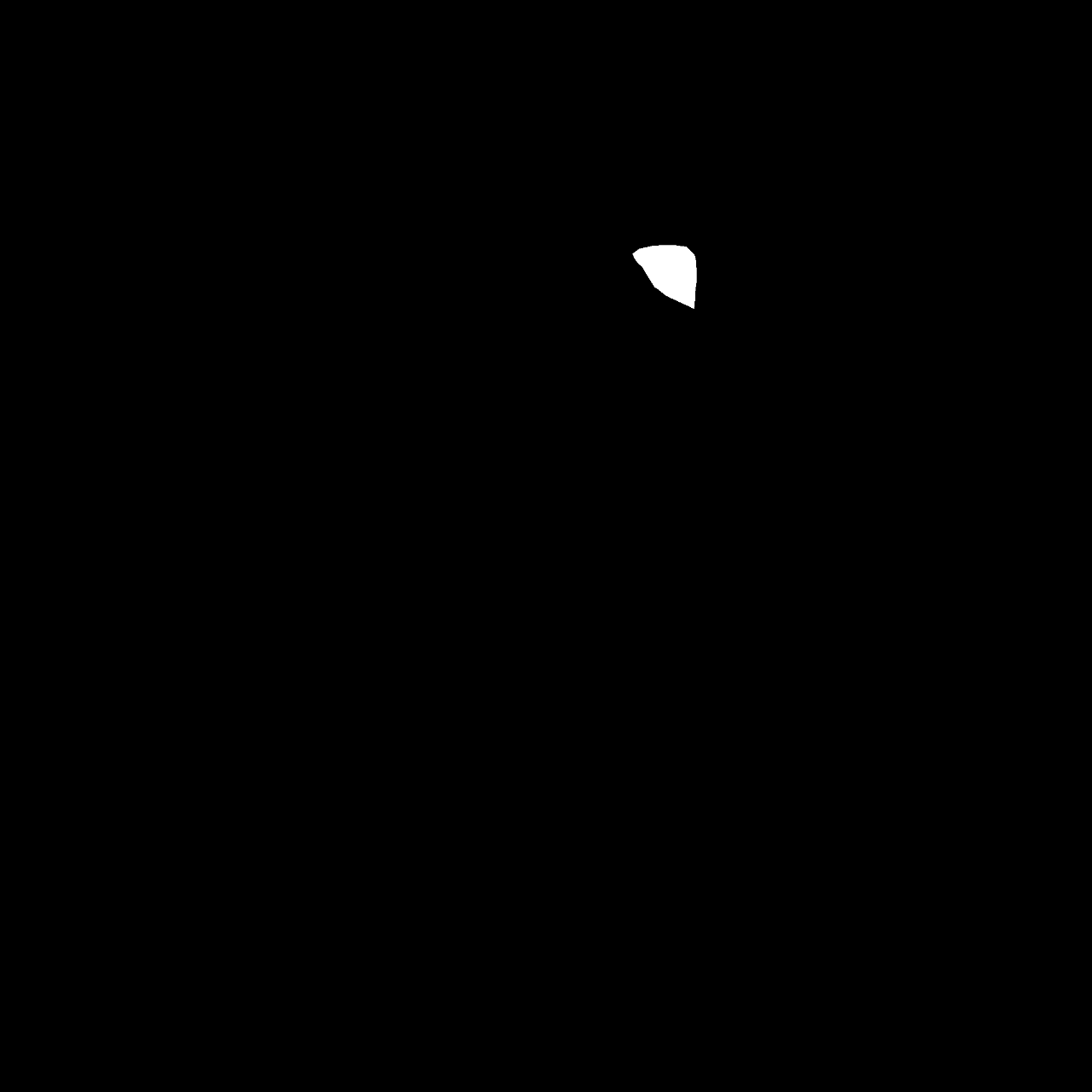}}\\
    \caption{For an input retina fundus image (a), we provide the lesion maps for four lesion classes (the bottom row). In the top row, the network provides a (b) class activation map (CAM) for the predicted class (\textit{grade-3}) (c) the overlap of CAM with the union of the predicted lesion maps, and (d) overlapping of CAM and union of predicted lesion maps on the original input image. In the second row, the CAM overlap with each of the lesion maps.}
    \label{fig:explantion_outputs}
\end{figure*}

% Input and outputs of our proposed intelligent diagnosis system for diabetic retinopathy (DR) disease.

\subsection{Performance of Trained System using User Feedback}
\begin{table}[h!]
\centering
% \captionsetup{skip=4pt}
\scalebox{0.85}{
\begin{tabular}{lcccc}
\hline
Method & \% direct user feedback & Accuracy & Kappa & Explanation \\ \hline
\multirow{6}{*}{G-Net + Attention}   & 0\%                              & 0.773             & 0.781          & 0.272                                   \\
                                     & \textit{10\%}                    & 0.786             & 0.801          & 0.295                                    \\
                                     & \textit{20\%}                    & 0.822             & 0.838          & 0.331                                    \\
                                     & \textit{30\%}                    & 0.841             & 0.853          & 0.363                                    \\
                                     & \textit{40\%}                    & 0.850             & 0.863          & 0.366                                    \\
                                     & \textit{50\%}                    & 0.855             & 0.866          & 0.387                                    \\ \hline
\end{tabular}}
\caption{Performance of our framework utilizing increasingly user-feedback. We iteratively fine-tune the model using simulated user-feedback. Experiments are conducted on \textit{FGADR} dataset.}
\label{table:user_feedback}
\end{table}

Within our interactive framework, given an input image, ophthalmologists can validate predicted lesion feature maps and grading class outputs, as well as provide feedback on these predictions. For example, if the predicted lesion maps have false positive or false negative regions, they can highlight the region and upon verification, these lesion maps can be used to further fine-tune our intelligent diagnosis system. As discussed in section \ref{subsec:feedback}, we can use these expert annotations directly to improve the performance of our lesion attentive grading model. In order to evaluate the effect of expert's validation and feedback process on our learning system, we conducted experiments by simulating user-feedback action. For this, we used \textit{FGADR} dataset to simulate user feedback because this dataset has both annotations for lesion maps and grading tasks.

\begin{figure}[H]
    \centering
    \scalebox{0.78}{
    \subfloat[original]{\includegraphics[width =1.3in]{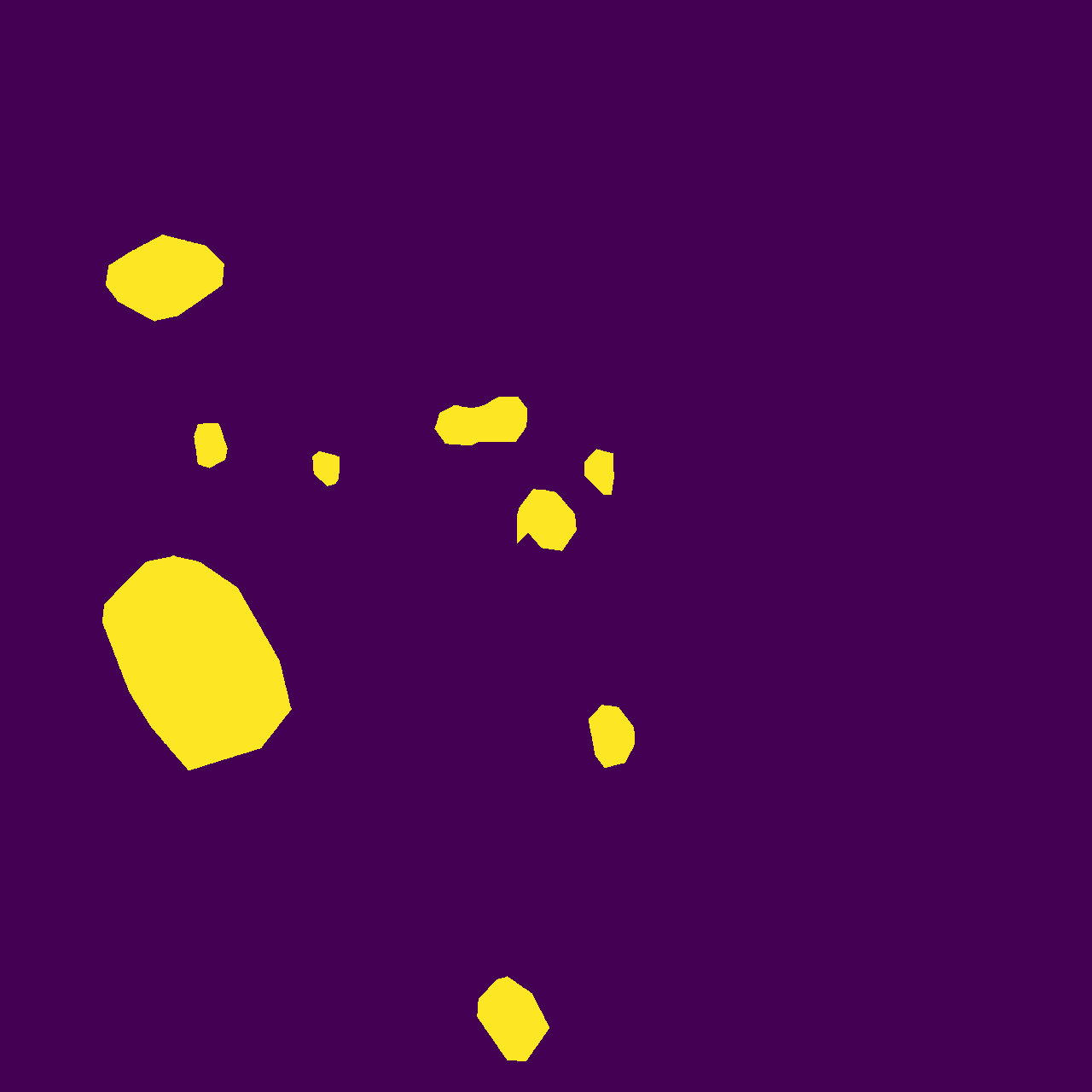}}\hspace{1em}
    %\hfill%
    % \hspace{-0.3cm}
    \subfloat[dilation]{\includegraphics[width =1.3in]{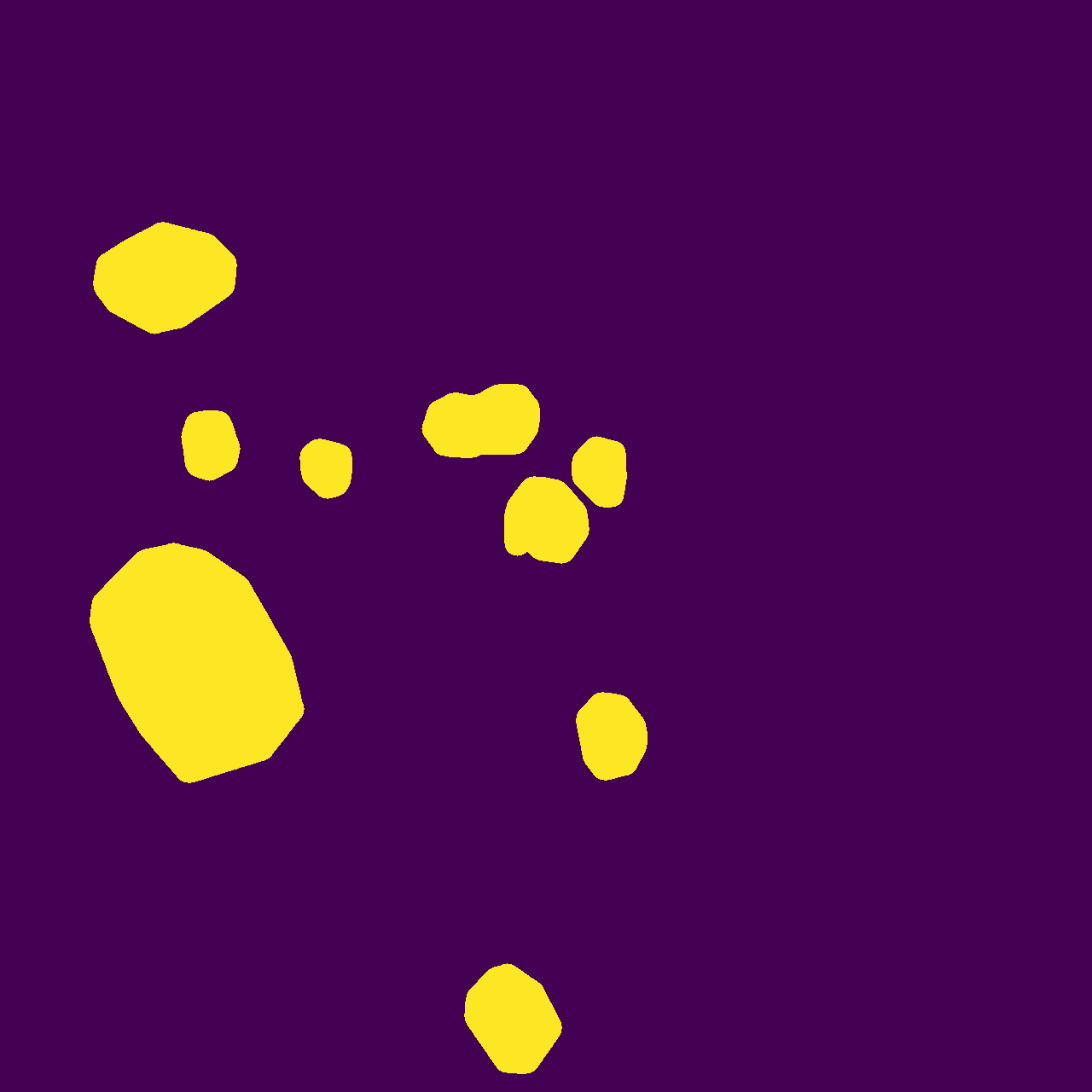}}\hspace{1em}
    %\hfill%
    % \hspace{0.05cm}
    \subfloat[erosion]{\includegraphics[width = 1.3in]{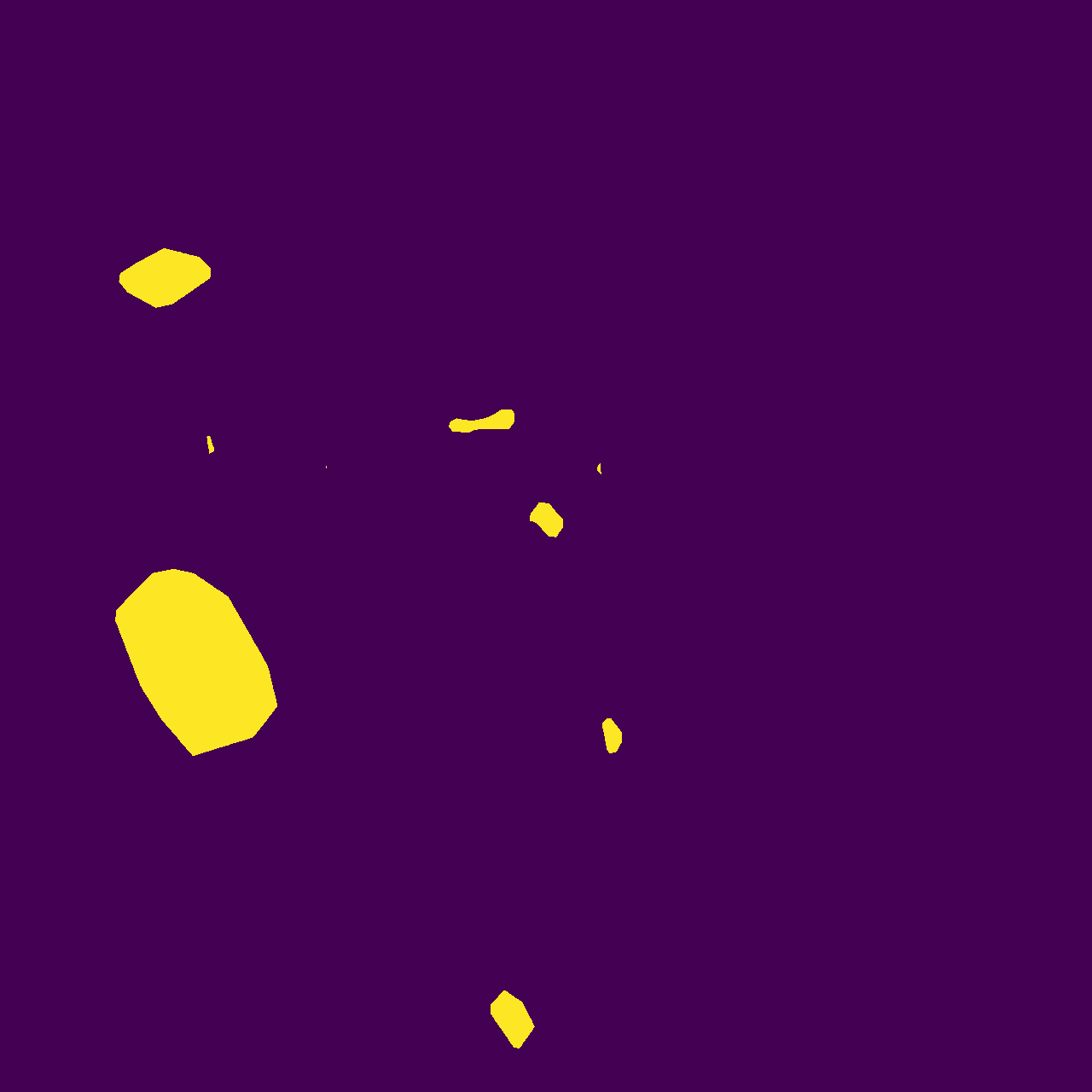}}}\\
        \caption{Illustration of noisy user-feedback on lesion features maps. To simulate the user behaviour, we add  randomly noises by performing dilation and erosion on original lesion masks and training models using these perturbed data. The examples are (a) original segmentation map, (b) lesion map after performing dilation, and (c) lesion map after erosion morphological operations using kernel size 15.}
    \label{fig:erosion_dilation}
\end{figure}

\begin{figure*}
    \captionsetup[subfigure]{labelformat=empty}
    \centering
    \subfloat{\includegraphics[width =0.9in]{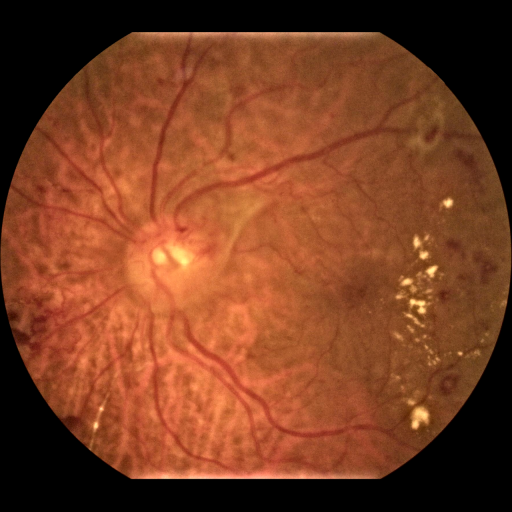}}\hspace{1em}
    % \hfill%
    % \hspace{0.05cm}
    \subfloat{\includegraphics[width =0.9in]{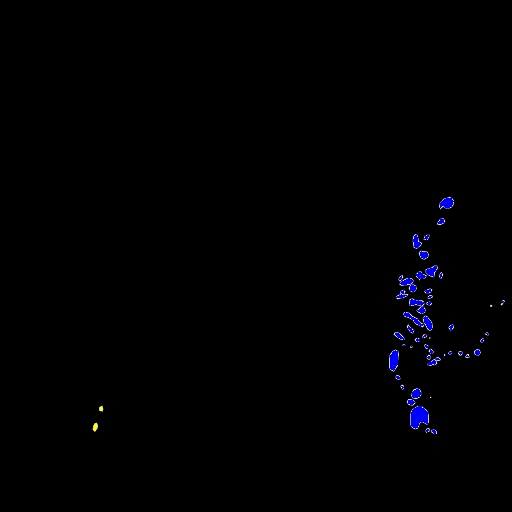}}\hspace{1em}
    % \hfill%
    % \hspace{0.05cm}
    \subfloat{\includegraphics[width = 0.9in]{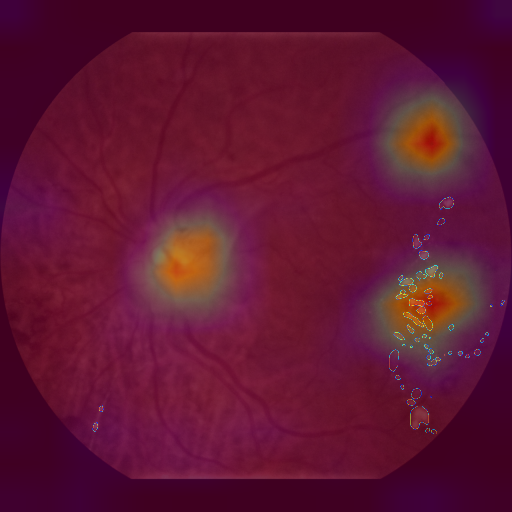}}\hspace{1em}
    %\hfill%
    % \hspace{0.05cm}
    \subfloat{\includegraphics[width = 0.9in]{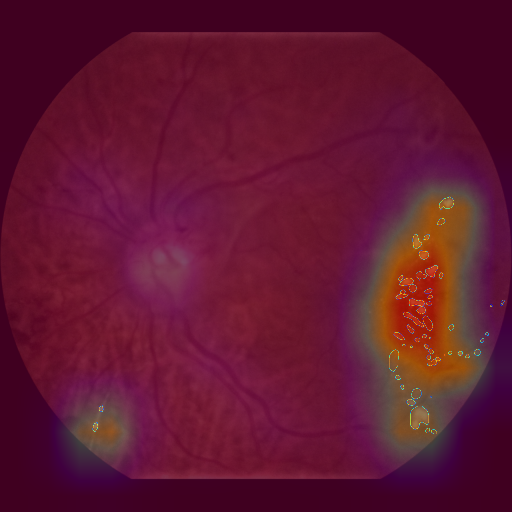}}\\
    \subfloat{\includegraphics[width =0.9in]{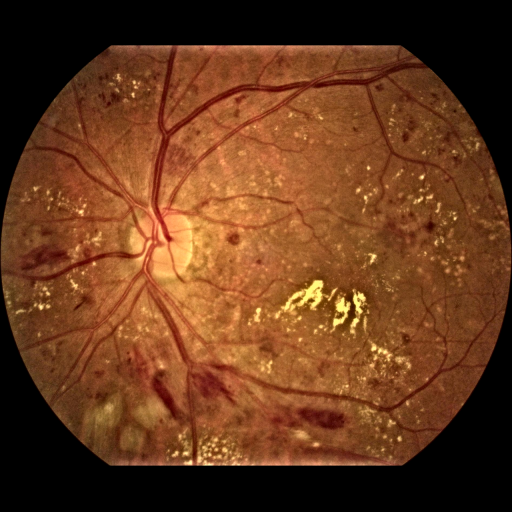}}\hspace{1em}
    % \hfill%
    % \hspace{0.05cm}
    \subfloat{\includegraphics[width =0.9in]{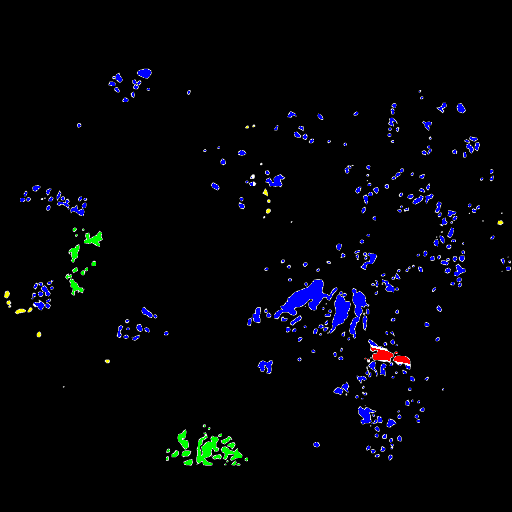}}\hspace{1em}
    %\hfill%
    % \hspace{0.05cm}
    \subfloat{\includegraphics[width = 0.9in]{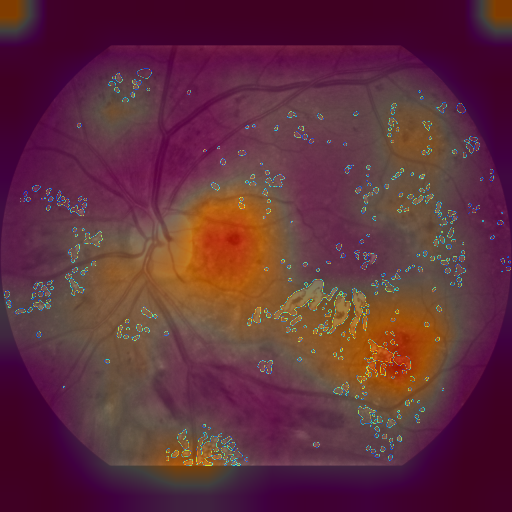}}\hspace{1em}
    %\hfill%
    % \hspace{0.05cm}
    \subfloat{\includegraphics[width = 0.9in]{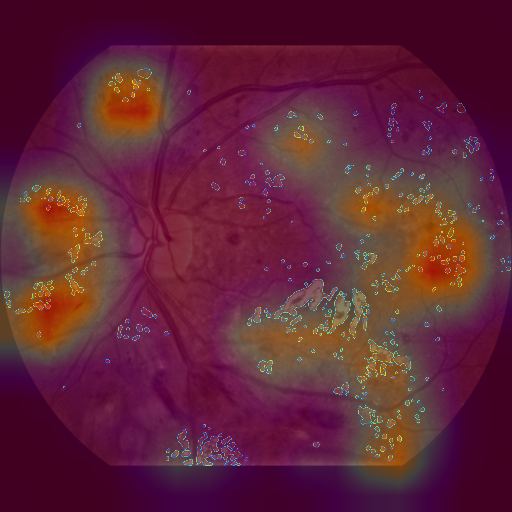}}\\
    \subfloat[(a) original image]{\includegraphics[width =0.9in]{images/experiment/interactive/fgard_0108_without_attention/ori_image_0108_1.png}}\hspace{1em}
    % \hfill%
    % \hspace{0.05cm}
    \subfloat[(b) predicted lesion masks]{\includegraphics[width =0.9in]{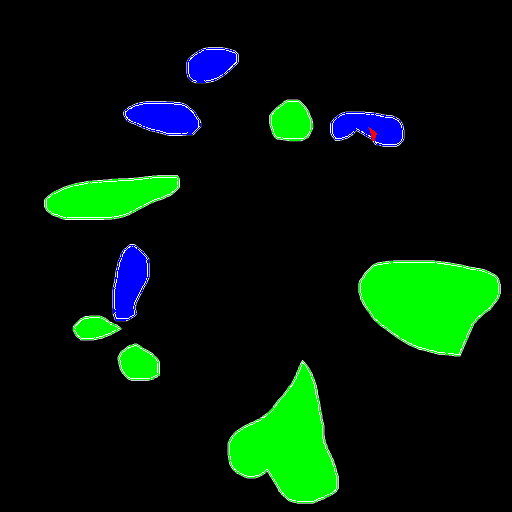}}\hspace{1em}
    %\hfill%
    % \hspace{0.05cm}
    \subfloat[(c) Activation map overlap \\without lesion attention]{\includegraphics[width = 0.9in]{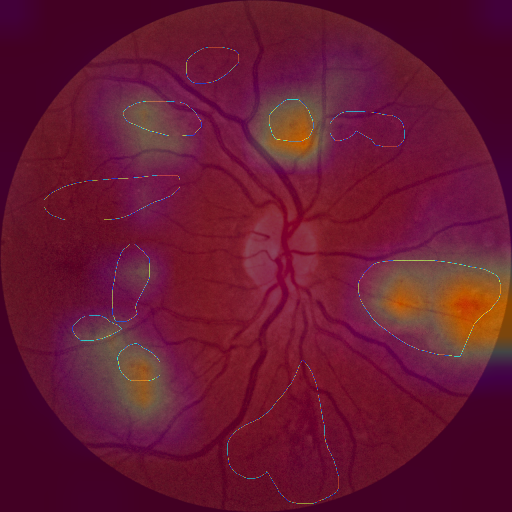}}\hspace{1em}
    %\hfill%
    % \hspace{0.05cm}
    \subfloat[(d) Activation map overlap with lesion attention (\textbf{ours})]{\includegraphics[width = 0.9in]{images/experiment/interactive/fgard_0108_with_attention/comb_all_three_0108_1.png}}
    % \subfloat[fig 3]{\includegraphics[width = 3in]{something}}
    % \subfloat[fig 4]{\includegraphics[width = 3in]{something}} 
    \caption{Comparison on explainability of the disease classification model trained with and without lesion attention. (a) original inputs, (b) union of ground-truth lesion maps, (c) overlapping of class activation map (CAM) of the predicted grading class with the (a) and (b) for \textit{G-Net} trained without lesion attention model \textit{Att-Net}, (d) overlapping of class activation map (CAM) of the predicted grading class with the (a) and (b) for \textit{G-Net} trained with lesion attention model \textit{Att-Net}. We can observe that CAMs in column (d) overlaps with most of the lesion regions.}
    \label{fig:cam_comparision_update}
\end{figure*}

In figure \ref{fig:user_simulation}, we illustrate our data split approach for user-feedback simulation. The original training data is divided into two equal splits. The disease grading model \textit{G-Net} in beginning is trained using a half of training data and their lesion masks predicted by \textit{S-Net}. The rest of training data is divided into five parts serving as increasingly data collected from users. After each update, a new model will be tested again on the fixed hold-out test set to justify performance. In our setting, instead of directly utilizing ground-truth samples as user feedback, we assumed that these data contain a certain noise level. This assumption makes sense in practice as we cannot guarantee accurate annotations from the user in all cases. Therefore, we simulated this scenario by applying some morphological operations on the ground-truth data and used them in the fine-tuning step. Specifically we randomly applied $\mathrm{erosion}$ and $\mathrm{dilation}$ operations for the lesion maps with a kernel size $15$. Figure \ref{fig:erosion_dilation} demonstrates the effect of these operations on a ground-truth lesion mask. 

% \usepackage{multirow}

Table \ref{table:user_feedback} presents our obtained results on \textit{FGADR} data for the DR classifier network with rising user feedback data. 
% The visualizations for these results are also illustrated in figure \ref{fig:user_feedback_graph}.
At each stage, 10\% of additional simulated user-feedback data is used to fine-tune a current stage of the grading network. We report both classification and explanation scores in our experiments, where the explanation score is computed based on the overlapping between the network's heatmap regions and detected lesion positions using the Jaccard similarly \cite{murphy1996finley}. Table \ref{table:user_feedback} demonstrates to us a trend of improvement in both classification and explanation scores given enlarged user-feedback data. For example, with 30\% or more response data, we achieved comparable scores against the other competitive methods discussed previously. In summary, these tests shows that the system can improve even further if there are feedback from users, and these feedback can be provided with very little effort from the users. 

% \begin{figure*}[h!]
% \centering
% \subfloat[classification accuracy]{\includegraphics[width = 2.8in]{images/experiment/interactive/fgadr_user_feedback_curve.png}}%\hfill% 
% \hspace{1em}
% \subfloat[explanation accuracy]{\includegraphics[width = 2.8in]{images/experiment/interactive/fgadr_user_feedback_curve_jaccard.png}}
% % \subfloat[fig 3]{\includegraphics[width = 3in]{something}}
% % \subfloat[fig 4]{\includegraphics[width = 3in]{something}} 
% \caption{Visualizing performance of our framework utilizing increasingly user-feedback computed by (a) accuracy and kappa metrics and (b) explanation scores. Experiments are conducted on \textit{FGADR} dataset.
% }
% \label{fig:user_feedback_graph}
% \end{figure*}
\section*{Conclusions and Future Works} \label{sec-conclusion}
This work provides a unified system for DR grading, which simultaneously learns to predict the disease grading and important lesion features. %Motivated by ophthalmologists' clinical behavior in identifying DR disease, we incorporate lesion characteristics into the learning process of our disease grade prediction neural model. Experiments revealed that this framework significantly improves baseline performance and outperforms other competitive benchmarks. Generally, our method has the following strengths:
The method can automatically detect different types of lesions across domains while using only annotations in the source domain and unlabeled data in the target domain. This allows for the rapid deployment of applications in new datasets without the need for extensive pixel-level annotation, which is often expensive and time-consuming to prepare. The attention network component of the proposed method  identifies and exploits the most significant lesion locations for improving performance and for explainability. It also makes the annotation progress more comfortable for users and make neural networks more robust in the presence of noise in new data. Moreover, the model also offers a variety of interpretative outputs to the user, such as DR Grading prediction, lesion positions, heatmap activation of the network, and the overlapping between these regions. However, there are various aspects that can be improved and explored.
First, even though our approach has integrated lesion information for the DR Grading task, there is still another medical criterion that can be utilized to improve performance. For instance, extensive characteristics of the lesion regions such as geometry, area, radius, or the degree of occurrence of different types of lesions, are other essential factors that need to formulate during the learning strategies. However, expressing such constraints is not straightforward since most of them are not differentiable, making end-to-end learning unfeasible. Fortunately, recent advances in machine learning subjects like discrete optimization \cite{poganvcic2019differentiation}, geometric deep learning \cite{monti2017geometric,bronstein2021geometric}, and physic-informed machine learning \cite{karniadakis2021physics} may offer us viable methods for incorporating these restrictions. For that reason, we believe that expanding our suggested method in those directions is worth investigating.

Second, the lack of training data is a primary obstacle that hinders the robustness and generalizability of a trained deep network. While we proposed techniques based on transfer learning and domain adaptation to alleviate these challenges, having a powerful pre-trained model is still in high demand. Currently, self-supervised learning methods trained on large-scale unlabeled data, namely the foundation model, have succeeded in various downstream tasks in natural language processing with well-known models such as BERT \cite{vit}, DALL-E \cite{ramesh2021zero}, and GPT-3 \cite{brown2020language}. This raises the question of whether foundation models trained on large-scale medical datasets can bring similar performance for downstream medical tasks \cite{nguyen2022joint}. In our setting, given such a foundation model, we expect to advance accuracy for neural networks in both lesion generator and DR Grading tasks, yielding increasing the performance of the whole system and reducing user efforts in preparing data annotations. 

Third, our proposed method can be applied to other medical applications such as brain tumor \cite{nguyen20173d,nguyen2022asmcnn,li2022dacbt} or skin cancer detection \cite{esteva2017dermatologist} where jointly localizing lesion information and data classification plays a crucial one. Therefore, designing experiments for these scenarios is an important aspect to validate the algorithm's generalization.

Last but not least, this study organized experiments to confirm that the proposed method grows accuracy over time when provided user feedback in weakly-supervised forms. However, because these results are simulated in the computer system, they may not cover all real scenarios in practice. This encourages us to build a real intelligent user interface and deploy it for real-world applications. Such a system when operating in practice requires overcoming various barriers. For example, developing a user-friendly and intuitive UI/UX system so that experts easily engage with and provide feedback to systems. When experts are in the annotation step, equipping computer-aided detection (CAD) \cite{yanase2019systematic} to discover and recommend similar marked locations to users is also necessary to save time and accelerate the progress. Finally, when there is a large amount of data feedback accessible, it poses concerns about learning for new instances while not forgetting past samples. Such questions are active topic research in continual learning and active learning, which is also a future direction for investigation.

%\clearpage
%\newpage

%\bibliographystyle{model2-names.bst}\biboptions{authoryear}
%\bibliography{references}

\clearpage

\bibliographystyle{model2-names.bst}
\bibliography{aaai23}

\begin{thebibliography}{83}
\expandafter\ifx\csname natexlab\endcsname\relax\def\natexlab#1{#1}\fi
\providecommand{\url}[1]{\texttt{#1}}
\providecommand{\href}[2]{#2}
\providecommand{\path}[1]{#1}
\providecommand{\DOIprefix}{doi:}
\providecommand{\ArXivprefix}{arXiv:}
\providecommand{\URLprefix}{URL: }
\providecommand{\Pubmedprefix}{pmid:}
\providecommand{\doi}[1]{\href{http://dx.doi.org/#1}{\path{#1}}}
\providecommand{\Pubmed}[1]{\href{pmid:#1}{\path{#1}}}
\providecommand{\bibinfo}[2]{#2}
\ifx\xfnm\relax \def\xfnm[#1]{\unskip,\space#1}\fi
%Type = Inproceedings
\bibitem[{Abnar and Zuidema(2020)}]{att-roll}
\bibinfo{author}{Abnar, S.}, \bibinfo{author}{Zuidema, W.},
  \bibinfo{year}{2020}.
\newblock \bibinfo{title}{Quantifying attention flow in transformers}, in:
  \bibinfo{booktitle}{Proceedings of the 58th Annual Meeting of the Association
  for Computational Linguistics}, \bibinfo{publisher}{Association for
  Computational Linguistics}, \bibinfo{address}{Online}. pp.
  \bibinfo{pages}{4190--4197}.
\newblock \URLprefix \url{https://aclanthology.org/2020.acl-main.385},
  \DOIprefix\doi{10.18653/v1/2020.acl-main.385}.
%Type = Article
\bibitem[{Al-Masni et~al.(2020)Al-Masni, Kim and Kim}]{al2020multiple}
\bibinfo{author}{Al-Masni, M.A.}, \bibinfo{author}{Kim, D.H.},
  \bibinfo{author}{Kim, T.S.}, \bibinfo{year}{2020}.
\newblock \bibinfo{title}{Multiple skin lesions diagnostics via integrated deep
  convolutional networks for segmentation and classification}.
\newblock \bibinfo{journal}{Computer methods and programs in biomedicine}
  \bibinfo{volume}{190}, \bibinfo{pages}{105351}.
%Type = Article
\bibitem[{Alban and Gilligan(2016)}]{alban2016automated}
\bibinfo{author}{Alban, M.}, \bibinfo{author}{Gilligan, T.},
  \bibinfo{year}{2016}.
\newblock \bibinfo{title}{Automated detection of diabetic retinopathy using
  fluorescein angiography photographs}.
\newblock \bibinfo{journal}{Report of standford education} .
%Type = Article
\bibitem[{Amershi et~al.(2014)Amershi, Cakmak, Knox and
  Kulesza}]{amershi2014power}
\bibinfo{author}{Amershi, S.}, \bibinfo{author}{Cakmak, M.},
  \bibinfo{author}{Knox, W.B.}, \bibinfo{author}{Kulesza, T.},
  \bibinfo{year}{2014}.
\newblock \bibinfo{title}{Power to the people: The role of humans in
  interactive machine learning}.
\newblock \bibinfo{journal}{Ai Magazine} \bibinfo{volume}{35},
  \bibinfo{pages}{105--120}.
%Type = Article
\bibitem[{Antal and Hajdu(2012)}]{antal2012ensemble}
\bibinfo{author}{Antal, B.}, \bibinfo{author}{Hajdu, A.}, \bibinfo{year}{2012}.
\newblock \bibinfo{title}{An ensemble-based system for microaneurysm detection
  and diabetic retinopathy grading}.
\newblock \bibinfo{journal}{IEEE transactions on biomedical engineering}
  \bibinfo{volume}{59}, \bibinfo{pages}{1720--1726}.
%Type = Inproceedings
\bibitem[{Arjovsky et~al.(2017)Arjovsky, Chintala and Bottou}]{wasser_gan}
\bibinfo{author}{Arjovsky, M.}, \bibinfo{author}{Chintala, S.},
  \bibinfo{author}{Bottou, L.}, \bibinfo{year}{2017}.
\newblock \bibinfo{title}{Wasserstein generative adversarial networks}, in:
  \bibinfo{booktitle}{International conference on machine learning},
  \bibinfo{organization}{PMLR}. pp. \bibinfo{pages}{214--223}.
%Type = Article
\bibitem[{Bronstein et~al.(2021)Bronstein, Bruna, Cohen and
  Veli{\v{c}}kovi{\'c}}]{bronstein2021geometric}
\bibinfo{author}{Bronstein, M.M.}, \bibinfo{author}{Bruna, J.},
  \bibinfo{author}{Cohen, T.}, \bibinfo{author}{Veli{\v{c}}kovi{\'c}, P.},
  \bibinfo{year}{2021}.
\newblock \bibinfo{title}{Geometric deep learning: Grids, groups, graphs,
  geodesics, and gauges}.
\newblock \bibinfo{journal}{arXiv preprint arXiv:2104.13478} .
%Type = Article
\bibitem[{Brown et~al.(2020)Brown, Mann, Ryder, Subbiah, Kaplan, Dhariwal,
  Neelakantan, Shyam, Sastry, Askell et~al.}]{brown2020language}
\bibinfo{author}{Brown, T.}, \bibinfo{author}{Mann, B.},
  \bibinfo{author}{Ryder, N.}, \bibinfo{author}{Subbiah, M.},
  \bibinfo{author}{Kaplan, J.D.}, \bibinfo{author}{Dhariwal, P.},
  \bibinfo{author}{Neelakantan, A.}, \bibinfo{author}{Shyam, P.},
  \bibinfo{author}{Sastry, G.}, \bibinfo{author}{Askell, A.}, et~al.,
  \bibinfo{year}{2020}.
\newblock \bibinfo{title}{Language models are few-shot learners}.
\newblock \bibinfo{journal}{Advances in neural information processing systems}
  \bibinfo{volume}{33}, \bibinfo{pages}{1877--1901}.
%Type = Article
\bibitem[{Chen et~al.(2017)Chen, Papandreou, Kokkinos, Murphy and
  Yuille}]{chen2017deeplab}
\bibinfo{author}{Chen, L.C.}, \bibinfo{author}{Papandreou, G.},
  \bibinfo{author}{Kokkinos, I.}, \bibinfo{author}{Murphy, K.},
  \bibinfo{author}{Yuille, A.L.}, \bibinfo{year}{2017}.
\newblock \bibinfo{title}{Deeplab: Semantic image segmentation with deep
  convolutional nets, atrous convolution, and fully connected crfs}.
\newblock \bibinfo{journal}{IEEE transactions on pattern analysis and machine
  intelligence} \bibinfo{volume}{40}, \bibinfo{pages}{834--848}.
%Type = Article
\bibitem[{Chen et~al.(2016)Chen, Duan, Houthooft, Schulman, Sutskever and
  Abbeel}]{chen2016infogan}
\bibinfo{author}{Chen, X.}, \bibinfo{author}{Duan, Y.},
  \bibinfo{author}{Houthooft, R.}, \bibinfo{author}{Schulman, J.},
  \bibinfo{author}{Sutskever, I.}, \bibinfo{author}{Abbeel, P.},
  \bibinfo{year}{2016}.
\newblock \bibinfo{title}{Infogan: Interpretable representation learning by
  information maximizing generative adversarial nets}.
\newblock \bibinfo{journal}{Advances in neural information processing systems}
  \bibinfo{volume}{29}.
%Type = Article
\bibitem[{Cheplygina(2019)}]{cats2019}
\bibinfo{author}{Cheplygina, V.}, \bibinfo{year}{2019}.
\newblock \bibinfo{title}{Cats or cat scans: Transfer learning from natural or
  medical image source data sets?}
\newblock \bibinfo{journal}{Current Opinion in Biomedical Engineering}
  \bibinfo{volume}{9}, \bibinfo{pages}{21--27}.
%Type = Article
\bibitem[{Dai et~al.(2021)Dai, Wu, Li, Cai, Wu, Kong, Liu, Wang, Hou, Liu
  et~al.}]{dai2021deep}
\bibinfo{author}{Dai, L.}, \bibinfo{author}{Wu, L.}, \bibinfo{author}{Li, H.},
  \bibinfo{author}{Cai, C.}, \bibinfo{author}{Wu, Q.}, \bibinfo{author}{Kong,
  H.}, \bibinfo{author}{Liu, R.}, \bibinfo{author}{Wang, X.},
  \bibinfo{author}{Hou, X.}, \bibinfo{author}{Liu, Y.}, et~al.,
  \bibinfo{year}{2021}.
\newblock \bibinfo{title}{A deep learning system for detecting diabetic
  retinopathy across the disease spectrum}.
\newblock \bibinfo{journal}{Nature communications} \bibinfo{volume}{12},
  \bibinfo{pages}{1--11}.
%Type = Article
\bibitem[{Dosovitskiy et~al.(2021)Dosovitskiy, Beyer, Kolesnikov, Weissenborn,
  Zhai, Unterthiner, Dehghani, Minderer, Heigold, Gelly et~al.}]{vit}
\bibinfo{author}{Dosovitskiy, A.}, \bibinfo{author}{Beyer, L.},
  \bibinfo{author}{Kolesnikov, A.}, \bibinfo{author}{Weissenborn, D.},
  \bibinfo{author}{Zhai, X.}, \bibinfo{author}{Unterthiner, T.},
  \bibinfo{author}{Dehghani, M.}, \bibinfo{author}{Minderer, M.},
  \bibinfo{author}{Heigold, G.}, \bibinfo{author}{Gelly, S.}, et~al.,
  \bibinfo{year}{2021}.
\newblock \bibinfo{title}{An image is worth 16x16 words: Transformers for image
  recognition at scale}.
\newblock \bibinfo{journal}{International Conference on Learning
  Representations (ICLR)} .
%Type = Article
\bibitem[{Esteva et~al.(2017)Esteva, Kuprel, Novoa, Ko, Swetter, Blau and
  Thrun}]{esteva2017dermatologist}
\bibinfo{author}{Esteva, A.}, \bibinfo{author}{Kuprel, B.},
  \bibinfo{author}{Novoa, R.A.}, \bibinfo{author}{Ko, J.},
  \bibinfo{author}{Swetter, S.M.}, \bibinfo{author}{Blau, H.M.},
  \bibinfo{author}{Thrun, S.}, \bibinfo{year}{2017}.
\newblock \bibinfo{title}{Dermatologist-level classification of skin cancer
  with deep neural networks}.
\newblock \bibinfo{journal}{nature} \bibinfo{volume}{542},
  \bibinfo{pages}{115--118}.
%Type = Misc
\bibitem[{EyePACS(2015)}]{kaggleEyePacsDataset}
\bibinfo{author}{EyePACS}, \bibinfo{year}{2015}.
\newblock \bibinfo{title}{Eyepacs challenge kaggle diabetic retinopathy
  dataset}.
\newblock
  \bibinfo{howpublished}{\url{https://www.kaggle.com/c/diabetic-retinopathy-detection/data}}.
\newblock \bibinfo{note}{Accessed: 2021-10-12}.
%Type = Article
\bibitem[{Goodfellow et~al.(2014)Goodfellow, Pouget-Abadie, Mirza, Xu,
  Warde-Farley, Ozair, Courville and Bengio}]{goodfellow2014generative}
\bibinfo{author}{Goodfellow, I.}, \bibinfo{author}{Pouget-Abadie, J.},
  \bibinfo{author}{Mirza, M.}, \bibinfo{author}{Xu, B.},
  \bibinfo{author}{Warde-Farley, D.}, \bibinfo{author}{Ozair, S.},
  \bibinfo{author}{Courville, A.}, \bibinfo{author}{Bengio, Y.},
  \bibinfo{year}{2014}.
\newblock \bibinfo{title}{Generative adversarial nets}.
\newblock \bibinfo{journal}{Advances in neural information processing systems}
  \bibinfo{volume}{27}.
%Type = Article
\bibitem[{Gretton et~al.(2012)Gretton, Borgwardt, Rasch, Sch{\"o}lkopf and
  Smola}]{gretton2012kernel}
\bibinfo{author}{Gretton, A.}, \bibinfo{author}{Borgwardt, K.M.},
  \bibinfo{author}{Rasch, M.J.}, \bibinfo{author}{Sch{\"o}lkopf, B.},
  \bibinfo{author}{Smola, A.}, \bibinfo{year}{2012}.
\newblock \bibinfo{title}{A kernel two-sample test}.
\newblock \bibinfo{journal}{The Journal of Machine Learning Research}
  \bibinfo{volume}{13}, \bibinfo{pages}{723--773}.
%Type = Article
\bibitem[{Gulrajani et~al.(2017)Gulrajani, Ahmed, Arjovsky, Dumoulin and
  Courville}]{Improved-wgan}
\bibinfo{author}{Gulrajani, I.}, \bibinfo{author}{Ahmed, F.},
  \bibinfo{author}{Arjovsky, M.}, \bibinfo{author}{Dumoulin, V.},
  \bibinfo{author}{Courville, A.C.}, \bibinfo{year}{2017}.
\newblock \bibinfo{title}{Improved training of wasserstein gans}.
\newblock \bibinfo{journal}{Advances in neural information processing systems}
  \bibinfo{volume}{30}.
%Type = Article
\bibitem[{Gulshan et~al.(2016)Gulshan, Peng, Coram, Stumpe, Wu, Narayanaswamy,
  Venugopalan, Widner, Madams, Cuadros et~al.}]{gulshan2016development}
\bibinfo{author}{Gulshan, V.}, \bibinfo{author}{Peng, L.},
  \bibinfo{author}{Coram, M.}, \bibinfo{author}{Stumpe, M.C.},
  \bibinfo{author}{Wu, D.}, \bibinfo{author}{Narayanaswamy, A.},
  \bibinfo{author}{Venugopalan, S.}, \bibinfo{author}{Widner, K.},
  \bibinfo{author}{Madams, T.}, \bibinfo{author}{Cuadros, J.}, et~al.,
  \bibinfo{year}{2016}.
\newblock \bibinfo{title}{Development and validation of a deep learning
  algorithm for detection of diabetic retinopathy in retinal fundus
  photographs}.
\newblock \bibinfo{journal}{Jama} \bibinfo{volume}{316},
  \bibinfo{pages}{2402--2410}.
%Type = Inproceedings
\bibitem[{He et~al.(2016)He, Zhang, Ren and Sun}]{resnet50}
\bibinfo{author}{He, K.}, \bibinfo{author}{Zhang, X.}, \bibinfo{author}{Ren,
  S.}, \bibinfo{author}{Sun, J.}, \bibinfo{year}{2016}.
\newblock \bibinfo{title}{Deep residual learning for image recognition}, in:
  \bibinfo{booktitle}{Proceedings of the IEEE conference on computer vision and
  pattern recognition}, pp. \bibinfo{pages}{770--778}.
%Type = Article
\bibitem[{He et~al.(2020)He, Yang, Zhang, Zhao, Zhang, Xing and
  Xie}]{he2020sample}
\bibinfo{author}{He, X.}, \bibinfo{author}{Yang, X.}, \bibinfo{author}{Zhang,
  S.}, \bibinfo{author}{Zhao, J.}, \bibinfo{author}{Zhang, Y.},
  \bibinfo{author}{Xing, E.}, \bibinfo{author}{Xie, P.}, \bibinfo{year}{2020}.
\newblock \bibinfo{title}{Sample-efficient deep learning for covid-19 diagnosis
  based on ct scans}.
\newblock \bibinfo{journal}{medrxiv} .
%Type = Article
\bibitem[{Hung et~al.(2018)Hung, Tsai, Liou, Lin and
  Yang}]{hung2018adversarial}
\bibinfo{author}{Hung, W.C.}, \bibinfo{author}{Tsai, Y.H.},
  \bibinfo{author}{Liou, Y.T.}, \bibinfo{author}{Lin, Y.Y.},
  \bibinfo{author}{Yang, M.H.}, \bibinfo{year}{2018}.
\newblock \bibinfo{title}{Adversarial learning for semi-supervised semantic
  segmentation}.
\newblock \bibinfo{journal}{Proceedings of the British Machine Vision
  Conference (BMVC)} .
%Type = Inproceedings
\bibitem[{Islam et~al.(2019)Islam, Vibashan, Jose, Wijethilake, Utkarsh and
  Ren}]{islam2019brain}
\bibinfo{author}{Islam, M.}, \bibinfo{author}{Vibashan, V.},
  \bibinfo{author}{Jose, V.}, \bibinfo{author}{Wijethilake, N.},
  \bibinfo{author}{Utkarsh, U.}, \bibinfo{author}{Ren, H.},
  \bibinfo{year}{2019}.
\newblock \bibinfo{title}{Brain tumor segmentation and survival prediction
  using 3d attention unet}, in: \bibinfo{booktitle}{International MICCAI
  Brainlesion Workshop}, \bibinfo{organization}{Springer}. pp.
  \bibinfo{pages}{262--272}.
%Type = Inproceedings
\bibitem[{Isola et~al.(2017)Isola, Zhu, Zhou and Efros}]{isola2018imagetoimage}
\bibinfo{author}{Isola, P.}, \bibinfo{author}{Zhu, J.Y.},
  \bibinfo{author}{Zhou, T.}, \bibinfo{author}{Efros, A.A.},
  \bibinfo{year}{2017}.
\newblock \bibinfo{title}{Image-to-image translation with conditional
  adversarial networks}, in: \bibinfo{booktitle}{Proceedings of the IEEE
  conference on computer vision and pattern recognition}, pp.
  \bibinfo{pages}{1125--1134}.
%Type = Inproceedings
\bibitem[{Jadon(2020)}]{jadon2020survey}
\bibinfo{author}{Jadon, S.}, \bibinfo{year}{2020}.
\newblock \bibinfo{title}{A survey of loss functions for semantic
  segmentation}, in: \bibinfo{booktitle}{2020 IEEE Conference on Computational
  Intelligence in Bioinformatics and Computational Biology (CIBCB)},
  \bibinfo{organization}{IEEE}. pp. \bibinfo{pages}{1--7}.
%Type = Inproceedings
\bibitem[{Jalwana et~al.(2021)Jalwana, Akhtar, Bennamoun and
  Mian}]{jalwana2021cameras}
\bibinfo{author}{Jalwana, M.A.}, \bibinfo{author}{Akhtar, N.},
  \bibinfo{author}{Bennamoun, M.}, \bibinfo{author}{Mian, A.},
  \bibinfo{year}{2021}.
\newblock \bibinfo{title}{Cameras: Enhanced resolution and sanity preserving
  class activation mapping for image saliency}, in:
  \bibinfo{booktitle}{Proceedings of the IEEE/CVF Conference on Computer Vision
  and Pattern Recognition}, pp. \bibinfo{pages}{16327--16336}.
%Type = Article
\bibitem[{Karniadakis et~al.(2021)Karniadakis, Kevrekidis, Lu, Perdikaris, Wang
  and Yang}]{karniadakis2021physics}
\bibinfo{author}{Karniadakis, G.E.}, \bibinfo{author}{Kevrekidis, I.G.},
  \bibinfo{author}{Lu, L.}, \bibinfo{author}{Perdikaris, P.},
  \bibinfo{author}{Wang, S.}, \bibinfo{author}{Yang, L.}, \bibinfo{year}{2021}.
\newblock \bibinfo{title}{Physics-informed machine learning}.
\newblock \bibinfo{journal}{Nature Reviews Physics} \bibinfo{volume}{3},
  \bibinfo{pages}{422--440}.
%Type = Inproceedings
\bibitem[{Le et~al.(2021)Le, Le, Yamazaki, Bui, Luu and Savides}]{le2021offset}
\bibinfo{author}{Le, N.}, \bibinfo{author}{Le, T.}, \bibinfo{author}{Yamazaki,
  K.}, \bibinfo{author}{Bui, T.}, \bibinfo{author}{Luu, K.},
  \bibinfo{author}{Savides, M.}, \bibinfo{year}{2021}.
\newblock \bibinfo{title}{Offset curves loss for imbalanced problem in medical
  segmentation}, in: \bibinfo{booktitle}{2020 25th International Conference on
  Pattern Recognition (ICPR)}, \bibinfo{organization}{IEEE}. pp.
  \bibinfo{pages}{9189--9195}.
%Type = Article
\bibitem[{Li et~al.(2022)Li, Khan, Alshara, Alotaibi, Mawuli
  et~al.}]{li2022dacbt}
\bibinfo{author}{Li, J.P.}, \bibinfo{author}{Khan, S.},
  \bibinfo{author}{Alshara, M.A.}, \bibinfo{author}{Alotaibi, R.M.},
  \bibinfo{author}{Mawuli, C.}, et~al., \bibinfo{year}{2022}.
\newblock \bibinfo{title}{Dacbt: deep learning approach for classification of
  brain tumors using mri data in iot healthcare environment}.
\newblock \bibinfo{journal}{Scientific Reports} \bibinfo{volume}{12},
  \bibinfo{pages}{1--14}.
%Type = Inproceedings
\bibitem[{Li et~al.(2018a)Li, Wu, Peng, Ernst and Fu}]{visual_attention}
\bibinfo{author}{Li, K.}, \bibinfo{author}{Wu, Z.}, \bibinfo{author}{Peng,
  K.C.}, \bibinfo{author}{Ernst, J.}, \bibinfo{author}{Fu, Y.},
  \bibinfo{year}{2018}a.
\newblock \bibinfo{title}{Tell me where to look: Guided attention inference
  network}, in: \bibinfo{booktitle}{Proceedings of the IEEE Conference on
  Computer Vision and Pattern Recognition}, pp. \bibinfo{pages}{9215--9223}.
%Type = Inproceedings
\bibitem[{Li et~al.(2020)Li, Verma, Nakashima, Kawasaki and
  Nagahara}]{li2020joint}
\bibinfo{author}{Li, L.}, \bibinfo{author}{Verma, M.},
  \bibinfo{author}{Nakashima, Y.}, \bibinfo{author}{Kawasaki, R.},
  \bibinfo{author}{Nagahara, H.}, \bibinfo{year}{2020}.
\newblock \bibinfo{title}{Joint learning of vessel segmentation and artery/vein
  classification with post-processing}, in: \bibinfo{booktitle}{Medical Imaging
  with Deep Learning}, \bibinfo{organization}{PMLR}. pp.
  \bibinfo{pages}{440--453}.
%Type = Article
\bibitem[{Li et~al.(2018b)Li, Chen, Qi, Dou, Fu and Heng}]{li2018h}
\bibinfo{author}{Li, X.}, \bibinfo{author}{Chen, H.}, \bibinfo{author}{Qi, X.},
  \bibinfo{author}{Dou, Q.}, \bibinfo{author}{Fu, C.W.}, \bibinfo{author}{Heng,
  P.A.}, \bibinfo{year}{2018}b.
\newblock \bibinfo{title}{H-denseunet: hybrid densely connected unet for liver
  and tumor segmentation from ct volumes}.
\newblock \bibinfo{journal}{IEEE transactions on medical imaging}
  \bibinfo{volume}{37}, \bibinfo{pages}{2663--2674}.
%Type = Inproceedings
\bibitem[{Lin et~al.(2018)Lin, Guo, Wang, Wu, Chen, Wang, Chen and
  Wu}]{lin2018framework}
\bibinfo{author}{Lin, Z.}, \bibinfo{author}{Guo, R.}, \bibinfo{author}{Wang,
  Y.}, \bibinfo{author}{Wu, B.}, \bibinfo{author}{Chen, T.},
  \bibinfo{author}{Wang, W.}, \bibinfo{author}{Chen, D.Z.},
  \bibinfo{author}{Wu, J.}, \bibinfo{year}{2018}.
\newblock \bibinfo{title}{A framework for identifying diabetic retinopathy
  based on anti-noise detection and attention-based fusion}, in:
  \bibinfo{booktitle}{International Conference on Medical Image Computing and
  Computer-Assisted Intervention}, \bibinfo{organization}{Springer}. pp.
  \bibinfo{pages}{74--82}.
%Type = Article
\bibitem[{Liu and Belkin(2018)}]{liu2018accelerating}
\bibinfo{author}{Liu, C.}, \bibinfo{author}{Belkin, M.}, \bibinfo{year}{2018}.
\newblock \bibinfo{title}{Accelerating sgd with momentum for over-parameterized
  learning}.
\newblock \bibinfo{journal}{arXiv preprint arXiv:1810.13395} .
%Type = Inproceedings
\bibitem[{Liu et~al.(2019)Liu, Gu and Samaras}]{liu2019wasserstein}
\bibinfo{author}{Liu, H.}, \bibinfo{author}{Gu, X.}, \bibinfo{author}{Samaras,
  D.}, \bibinfo{year}{2019}.
\newblock \bibinfo{title}{Wasserstein gan with quadratic transport cost}, in:
  \bibinfo{booktitle}{Proceedings of the IEEE/CVF international conference on
  computer vision}, pp. \bibinfo{pages}{4832--4841}.
%Type = Inproceedings
\bibitem[{Long et~al.(2015)Long, Shelhamer and Darrell}]{long2015fully}
\bibinfo{author}{Long, J.}, \bibinfo{author}{Shelhamer, E.},
  \bibinfo{author}{Darrell, T.}, \bibinfo{year}{2015}.
\newblock \bibinfo{title}{Fully convolutional networks for semantic
  segmentation}, in: \bibinfo{booktitle}{Proceedings of the IEEE conference on
  computer vision and pattern recognition}, pp. \bibinfo{pages}{3431--3440}.
%Type = Article
\bibitem[{Ma et~al.(2021)Ma, Li, Lu, Zhu and Shen}]{ma2021adversarial}
\bibinfo{author}{Ma, A.}, \bibinfo{author}{Li, J.}, \bibinfo{author}{Lu, K.},
  \bibinfo{author}{Zhu, L.}, \bibinfo{author}{Shen, H.T.},
  \bibinfo{year}{2021}.
\newblock \bibinfo{title}{Adversarial entropy optimization for unsupervised
  domain adaptation}.
\newblock \bibinfo{journal}{IEEE Transactions on Neural Networks and Learning
  Systems} .
%Type = Article
\bibitem[{Maadi et~al.(2021)Maadi, Akbarzadeh~Khorshidi and
  Aickelin}]{maadi2021review}
\bibinfo{author}{Maadi, M.}, \bibinfo{author}{Akbarzadeh~Khorshidi, H.},
  \bibinfo{author}{Aickelin, U.}, \bibinfo{year}{2021}.
\newblock \bibinfo{title}{A review on human--ai interaction in machine learning
  and insights for medical applications}.
\newblock \bibinfo{journal}{International journal of environmental research and
  public health} \bibinfo{volume}{18}, \bibinfo{pages}{2121}.
%Type = Misc
\bibitem[{McHugh(2012)}]{mchugh2012interrater}
\bibinfo{author}{McHugh, M.}, \bibinfo{year}{2012}.
\newblock \bibinfo{title}{interrater reliability: the kappa statistic.
  biochemica medica, 22 (3), 276--282}.
%Type = Inproceedings
\bibitem[{Mehta et~al.(2018)Mehta, Mercan, Bartlett, Weaver, Elmore and
  Shapiro}]{mehta2018ynet}
\bibinfo{author}{Mehta, S.}, \bibinfo{author}{Mercan, E.},
  \bibinfo{author}{Bartlett, J.}, \bibinfo{author}{Weaver, D.},
  \bibinfo{author}{Elmore, J.G.}, \bibinfo{author}{Shapiro, L.},
  \bibinfo{year}{2018}.
\newblock \bibinfo{title}{Y-net: joint segmentation and classification for
  diagnosis of breast biopsy images}, in: \bibinfo{booktitle}{International
  Conference on Medical Image Computing and Computer-Assisted Intervention},
  \bibinfo{organization}{Springer}. pp. \bibinfo{pages}{893--901}.
%Type = Article
\bibitem[{Mirza and Osindero(2014)}]{mirza2014conditional}
\bibinfo{author}{Mirza, M.}, \bibinfo{author}{Osindero, S.},
  \bibinfo{year}{2014}.
\newblock \bibinfo{title}{Conditional generative adversarial nets}.
\newblock \bibinfo{journal}{arXiv preprint arXiv:1411.1784} .
%Type = Inproceedings
\bibitem[{Monti et~al.(2017)Monti, Boscaini, Masci, Rodola, Svoboda and
  Bronstein}]{monti2017geometric}
\bibinfo{author}{Monti, F.}, \bibinfo{author}{Boscaini, D.},
  \bibinfo{author}{Masci, J.}, \bibinfo{author}{Rodola, E.},
  \bibinfo{author}{Svoboda, J.}, \bibinfo{author}{Bronstein, M.M.},
  \bibinfo{year}{2017}.
\newblock \bibinfo{title}{Geometric deep learning on graphs and manifolds using
  mixture model cnns}, in: \bibinfo{booktitle}{Proceedings of the IEEE
  conference on computer vision and pattern recognition}, pp.
  \bibinfo{pages}{5115--5124}.
%Type = Article
\bibitem[{Murphy(1996)}]{murphy1996finley}
\bibinfo{author}{Murphy, A.H.}, \bibinfo{year}{1996}.
\newblock \bibinfo{title}{The finley affair: A signal event in the history of
  forecast verification}.
\newblock \bibinfo{journal}{Weather and forecasting} \bibinfo{volume}{11},
  \bibinfo{pages}{3--20}.
%Type = Article
\bibitem[{Nguyen et~al.(2022a)Nguyen, Nguyen, Mai, Nguyen, Tran, Nguyen, Pham
  and Nguyen}]{nguyen2022asmcnn}
\bibinfo{author}{Nguyen, D.H.}, \bibinfo{author}{Nguyen, D.M.},
  \bibinfo{author}{Mai, T.T.N.}, \bibinfo{author}{Nguyen, T.},
  \bibinfo{author}{Tran, K.T.}, \bibinfo{author}{Nguyen, A.T.},
  \bibinfo{author}{Pham, B.T.}, \bibinfo{author}{Nguyen, B.T.},
  \bibinfo{year}{2022}a.
\newblock \bibinfo{title}{Asmcnn: An efficient brain extraction using active
  shape model and convolutional neural networks}.
\newblock \bibinfo{journal}{Information Sciences} \bibinfo{volume}{591},
  \bibinfo{pages}{25--48}.
%Type = Inproceedings
\bibitem[{Nguyen et~al.(2021a)Nguyen, Mai, Than, Prange and
  Sonntag}]{nguyen2021self}
\bibinfo{author}{Nguyen, D.M.}, \bibinfo{author}{Mai, T.T.N.},
  \bibinfo{author}{Than, N.T.}, \bibinfo{author}{Prange, A.},
  \bibinfo{author}{Sonntag, D.}, \bibinfo{year}{2021}a.
\newblock \bibinfo{title}{Self-supervised domain adaptation for diabetic
  retinopathy grading using vessel image reconstruction}, in:
  \bibinfo{booktitle}{German Conference on Artificial Intelligence
  (K{\"u}nstliche Intelligenz)}, \bibinfo{organization}{Springer}. pp.
  \bibinfo{pages}{349--361}.
%Type = Inproceedings
\bibitem[{Nguyen et~al.(2021b)Nguyen, Nguyen, Vu, Nguyen, Nunnari and
  Sonntag}]{nguyen2021attention}
\bibinfo{author}{Nguyen, D.M.}, \bibinfo{author}{Nguyen, D.M.},
  \bibinfo{author}{Vu, H.}, \bibinfo{author}{Nguyen, B.T.},
  \bibinfo{author}{Nunnari, F.}, \bibinfo{author}{Sonntag, D.},
  \bibinfo{year}{2021}b.
\newblock \bibinfo{title}{An attention mechanism using multiple knowledge
  sources for covid-19 detection from ct images}, in: \bibinfo{booktitle}{The
  Thirty-Fifth AAAI Conference on Artificial Intelligence (AAAI-21), Workshop:
  Trustworthy AI for Healthcare}.
%Type = Article
\bibitem[{Nguyen et~al.(2022b)Nguyen, Nguyen, Vu, Pham, Nguyen, Nguyen and
  Sonntag}]{nguyen2021tatl}
\bibinfo{author}{Nguyen, D.M.}, \bibinfo{author}{Nguyen, T.T.},
  \bibinfo{author}{Vu, H.}, \bibinfo{author}{Pham, Q.},
  \bibinfo{author}{Nguyen, M.D.}, \bibinfo{author}{Nguyen, B.T.},
  \bibinfo{author}{Sonntag, D.}, \bibinfo{year}{2022}b.
\newblock \bibinfo{title}{{TATL}: task agnostic transfer learning for skin
  attributes detection}.
\newblock \bibinfo{journal}{Medical Image Analysis} \bibinfo{volume}{78},
  \bibinfo{pages}{102359}.
%Type = Inproceedings
\bibitem[{Nguyen et~al.(2017)Nguyen, Vu, Ung and Nguyen}]{nguyen20173d}
\bibinfo{author}{Nguyen, D.M.}, \bibinfo{author}{Vu, H.T.},
  \bibinfo{author}{Ung, H.Q.}, \bibinfo{author}{Nguyen, B.T.},
  \bibinfo{year}{2017}.
\newblock \bibinfo{title}{3d-brain segmentation using deep neural network and
  gaussian mixture model}, in: \bibinfo{booktitle}{2017 IEEE Winter Conference
  on Applications of Computer Vision (WACV)}, \bibinfo{organization}{IEEE}. pp.
  \bibinfo{pages}{815--824}.
%Type = Article
\bibitem[{Nguyen et~al.(2022c)Nguyen, Nguyen, Truong, Cao, Nguyen, Ho, Swoboda,
  Albarqouni, Xie and Sonntag}]{nguyen2022joint}
\bibinfo{author}{Nguyen, D.M.H.}, \bibinfo{author}{Nguyen, H.},
  \bibinfo{author}{Truong, M.T.N.}, \bibinfo{author}{Cao, T.},
  \bibinfo{author}{Nguyen, B.T.}, \bibinfo{author}{Ho, N.},
  \bibinfo{author}{Swoboda, P.}, \bibinfo{author}{Albarqouni, S.},
  \bibinfo{author}{Xie, P.}, \bibinfo{author}{Sonntag, D.},
  \bibinfo{year}{2022}c.
\newblock \bibinfo{title}{Joint self-supervised image-volume representation
  learning with intra-inter contrastive clustering}.
\newblock \bibinfo{journal}{arXiv preprint arXiv:2212.01893} .
%Type = Incollection
\bibitem[{Paszke et~al.(2019)Paszke, Gross, Massa, Lerer, Bradbury, Chanan,
  Killeen, Lin, Gimelshein, Antiga, Desmaison, Kopf, Yang, DeVito, Raison,
  Tejani, Chilamkurthy, Steiner, Fang, Bai and Chintala}]{NEURIPS2019_9015}
\bibinfo{author}{Paszke, A.}, \bibinfo{author}{Gross, S.},
  \bibinfo{author}{Massa, F.}, \bibinfo{author}{Lerer, A.},
  \bibinfo{author}{Bradbury, J.}, \bibinfo{author}{Chanan, G.},
  \bibinfo{author}{Killeen, T.}, \bibinfo{author}{Lin, Z.},
  \bibinfo{author}{Gimelshein, N.}, \bibinfo{author}{Antiga, L.},
  \bibinfo{author}{Desmaison, A.}, \bibinfo{author}{Kopf, A.},
  \bibinfo{author}{Yang, E.}, \bibinfo{author}{DeVito, Z.},
  \bibinfo{author}{Raison, M.}, \bibinfo{author}{Tejani, A.},
  \bibinfo{author}{Chilamkurthy, S.}, \bibinfo{author}{Steiner, B.},
  \bibinfo{author}{Fang, L.}, \bibinfo{author}{Bai, J.},
  \bibinfo{author}{Chintala, S.}, \bibinfo{year}{2019}.
\newblock \bibinfo{title}{Pytorch: An imperative style, high-performance deep
  learning library}, in: \bibinfo{editor}{Wallach, H.},
  \bibinfo{editor}{Larochelle, H.}, \bibinfo{editor}{Beygelzimer, A.},
  \bibinfo{editor}{d\textquotesingle Alch\'{e}-Buc, F.}, \bibinfo{editor}{Fox,
  E.}, \bibinfo{editor}{Garnett, R.} (Eds.), \bibinfo{booktitle}{Advances in
  Neural Information Processing Systems 32}. \bibinfo{publisher}{Curran
  Associates, Inc.}, pp. \bibinfo{pages}{8024--8035}.
%Type = Inproceedings
\bibitem[{Pogan{\v{c}}i{\'c} et~al.(2019)Pogan{\v{c}}i{\'c}, Paulus, Musil,
  Martius and Rolinek}]{poganvcic2019differentiation}
\bibinfo{author}{Pogan{\v{c}}i{\'c}, M.V.}, \bibinfo{author}{Paulus, A.},
  \bibinfo{author}{Musil, V.}, \bibinfo{author}{Martius, G.},
  \bibinfo{author}{Rolinek, M.}, \bibinfo{year}{2019}.
\newblock \bibinfo{title}{Differentiation of blackbox combinatorial solvers},
  in: \bibinfo{booktitle}{International Conference on Learning
  Representations}.
%Type = Article
\bibitem[{Porwal et~al.(2018)Porwal, Pachade, Kamble, Kokare, Deshmukh,
  Sahasrabuddhe and Meriaudeau}]{h25w98-18}
\bibinfo{author}{Porwal, P.}, \bibinfo{author}{Pachade, S.},
  \bibinfo{author}{Kamble, R.}, \bibinfo{author}{Kokare, M.},
  \bibinfo{author}{Deshmukh, G.}, \bibinfo{author}{Sahasrabuddhe, V.},
  \bibinfo{author}{Meriaudeau, F.}, \bibinfo{year}{2018}.
\newblock \bibinfo{title}{Indian diabetic retinopathy image dataset (idrid): a
  database for diabetic retinopathy screening research}.
\newblock \bibinfo{journal}{Data} \bibinfo{volume}{3}, \bibinfo{pages}{25}.
%Type = Inproceedings
\bibitem[{Prange et~al.(2017)Prange, Chikobava, Poller, Barz and
  Sonntag}]{prange2017multimodal}
\bibinfo{author}{Prange, A.}, \bibinfo{author}{Chikobava, M.},
  \bibinfo{author}{Poller, P.}, \bibinfo{author}{Barz, M.},
  \bibinfo{author}{Sonntag, D.}, \bibinfo{year}{2017}.
\newblock \bibinfo{title}{A multimodal dialogue system for medical decision
  support inside virtual reality}, in: \bibinfo{booktitle}{Proceedings of the
  18th Annual SIGdial Meeting on Discourse and Dialogue}, pp.
  \bibinfo{pages}{23--26}.
%Type = Inproceedings
\bibitem[{Ramesh et~al.(2021)Ramesh, Pavlov, Goh, Gray, Voss, Radford, Chen and
  Sutskever}]{ramesh2021zero}
\bibinfo{author}{Ramesh, A.}, \bibinfo{author}{Pavlov, M.},
  \bibinfo{author}{Goh, G.}, \bibinfo{author}{Gray, S.}, \bibinfo{author}{Voss,
  C.}, \bibinfo{author}{Radford, A.}, \bibinfo{author}{Chen, M.},
  \bibinfo{author}{Sutskever, I.}, \bibinfo{year}{2021}.
\newblock \bibinfo{title}{Zero-shot text-to-image generation}, in:
  \bibinfo{booktitle}{International Conference on Machine Learning},
  \bibinfo{organization}{PMLR}. pp. \bibinfo{pages}{8821--8831}.
%Type = Inproceedings
\bibitem[{Ronneberger et~al.(2015)Ronneberger, Fischer and
  Brox}]{ronneberger2015unet}
\bibinfo{author}{Ronneberger, O.}, \bibinfo{author}{Fischer, P.},
  \bibinfo{author}{Brox, T.}, \bibinfo{year}{2015}.
\newblock \bibinfo{title}{U-net: Convolutional networks for biomedical image
  segmentation}, in: \bibinfo{booktitle}{International Conference on Medical
  image computing and computer-assisted intervention},
  \bibinfo{organization}{Springer}. pp. \bibinfo{pages}{234--241}.
%Type = Article
\bibitem[{Ruder(2016)}]{ruder2016overview}
\bibinfo{author}{Ruder, S.}, \bibinfo{year}{2016}.
\newblock \bibinfo{title}{An overview of gradient descent optimization
  algorithms}.
\newblock \bibinfo{journal}{arXiv preprint arXiv:1609.04747} .
%Type = Article
\bibitem[{Saeedi et~al.(2019)Saeedi, Petersohn, Salpea, Malanda, Karuranga,
  Unwin, Colagiuri, Guariguata, Motala, Ogurtsova et~al.}]{saeedi2019global}
\bibinfo{author}{Saeedi, P.}, \bibinfo{author}{Petersohn, I.},
  \bibinfo{author}{Salpea, P.}, \bibinfo{author}{Malanda, B.},
  \bibinfo{author}{Karuranga, S.}, \bibinfo{author}{Unwin, N.},
  \bibinfo{author}{Colagiuri, S.}, \bibinfo{author}{Guariguata, L.},
  \bibinfo{author}{Motala, A.A.}, \bibinfo{author}{Ogurtsova, K.}, et~al.,
  \bibinfo{year}{2019}.
\newblock \bibinfo{title}{Global and regional diabetes prevalence estimates for
  2019 and projections for 2030 and 2045: Results from the international
  diabetes federation diabetes atlas}.
\newblock \bibinfo{journal}{Diabetes research and clinical practice}
  \bibinfo{volume}{157}, \bibinfo{pages}{107843}.
%Type = Misc
\bibitem[{Saini and Prasad(2021)}]{lime}
\bibinfo{author}{Saini, A.}, \bibinfo{author}{Prasad, R.},
  \bibinfo{year}{2021}.
\newblock \bibinfo{title}{Select wisely and explain: Active learning and
  probabilistic local post-hoc explainability}.
\newblock \URLprefix \url{https://arxiv.org/abs/2108.06907},
  \DOIprefix\doi{10.48550/ARXIV.2108.06907}.
%Type = Article
\bibitem[{Sankar et~al.(2016)Sankar, Batri and Parvathi}]{sankar2016earliest}
\bibinfo{author}{Sankar, M.}, \bibinfo{author}{Batri, K.},
  \bibinfo{author}{Parvathi, R.}, \bibinfo{year}{2016}.
\newblock \bibinfo{title}{Earliest diabetic retinopathy classification using
  deep convolution neural networks. pdf}.
\newblock \bibinfo{journal}{Int. J. Adv. Eng. Technol} \bibinfo{volume}{10},
  \bibinfo{pages}{M9}.
%Type = Inproceedings
\bibitem[{Selvaraju et~al.(2017)Selvaraju, Cogswell, Das, Vedantam, Parikh and
  Batra}]{selvaraju2017grad}
\bibinfo{author}{Selvaraju, R.R.}, \bibinfo{author}{Cogswell, M.},
  \bibinfo{author}{Das, A.}, \bibinfo{author}{Vedantam, R.},
  \bibinfo{author}{Parikh, D.}, \bibinfo{author}{Batra, D.},
  \bibinfo{year}{2017}.
\newblock \bibinfo{title}{Grad-cam: Visual explanations from deep networks via
  gradient-based localization}, in: \bibinfo{booktitle}{Proceedings of the IEEE
  international conference on computer vision}, pp. \bibinfo{pages}{618--626}.
%Type = Article
\bibitem[{Shannon(2001)}]{entropy}
\bibinfo{author}{Shannon, C.E.}, \bibinfo{year}{2001}.
\newblock \bibinfo{title}{A mathematical theory of communication}.
\newblock \bibinfo{journal}{ACM SIGMOBILE mobile computing and communications
  review} \bibinfo{volume}{5}, \bibinfo{pages}{3--55}.
%Type = Inproceedings
\bibitem[{Shen et~al.(2018)Shen, Qu, Zhang and Yu}]{shen2018wasserstein}
\bibinfo{author}{Shen, J.}, \bibinfo{author}{Qu, Y.}, \bibinfo{author}{Zhang,
  W.}, \bibinfo{author}{Yu, Y.}, \bibinfo{year}{2018}.
\newblock \bibinfo{title}{Wasserstein distance guided representation learning
  for domain adaptation}, in: \bibinfo{booktitle}{Thirty-second AAAI conference
  on artificial intelligence}.
%Type = Inproceedings
\bibitem[{Smith(2017)}]{smith2017cyclical}
\bibinfo{author}{Smith, L.N.}, \bibinfo{year}{2017}.
\newblock \bibinfo{title}{Cyclical learning rates for training neural
  networks}, in: \bibinfo{booktitle}{2017 IEEE winter conference on
  applications of computer vision (WACV)}, \bibinfo{organization}{IEEE}. pp.
  \bibinfo{pages}{464--472}.
%Type = Article
\bibitem[{Sonntag et~al.(2020)Sonntag, Nunnari and
  Profitlich}]{sonntag2020skincare}
\bibinfo{author}{Sonntag, D.}, \bibinfo{author}{Nunnari, F.},
  \bibinfo{author}{Profitlich, H.J.}, \bibinfo{year}{2020}.
\newblock \bibinfo{title}{The skincare project, an interactive deep learning
  system for differential diagnosis of malignant skin lesions. technical
  report}.
\newblock \bibinfo{journal}{arXiv preprint arXiv:2005.09448} .
%Type = Inproceedings
\bibitem[{Sonntag et~al.(2012)Sonntag, Schulz, Reuschling and
  Galarraga}]{sonntag2012radspeech}
\bibinfo{author}{Sonntag, D.}, \bibinfo{author}{Schulz, C.},
  \bibinfo{author}{Reuschling, C.}, \bibinfo{author}{Galarraga, L.},
  \bibinfo{year}{2012}.
\newblock \bibinfo{title}{Radspeech's mobile dialogue system for radiologists},
  in: \bibinfo{booktitle}{Proceedings of the 2012 ACM international conference
  on Intelligent User Interfaces}, pp. \bibinfo{pages}{317--318}.
%Type = Misc
\bibitem[{Sonntag et~al.(2010)Sonntag, Wennerberg, Buitelaar and
  Zillner}]{sonntag2010pillars}
\bibinfo{author}{Sonntag, D.}, \bibinfo{author}{Wennerberg, P.},
  \bibinfo{author}{Buitelaar, P.}, \bibinfo{author}{Zillner, S.},
  \bibinfo{year}{2010}.
\newblock \bibinfo{title}{Pillars of ontology treatment in the medical domain}.
%Type = Inproceedings
\bibitem[{Sun et~al.(2021)Sun, Li, Zhang, Mao, Wu and Zhang}]{sun2021lesion}
\bibinfo{author}{Sun, R.}, \bibinfo{author}{Li, Y.}, \bibinfo{author}{Zhang,
  T.}, \bibinfo{author}{Mao, Z.}, \bibinfo{author}{Wu, F.},
  \bibinfo{author}{Zhang, Y.}, \bibinfo{year}{2021}.
\newblock \bibinfo{title}{Lesion-aware transformers for diabetic retinopathy
  grading}, in: \bibinfo{booktitle}{Proceedings of the IEEE/CVF Conference on
  Computer Vision and Pattern Recognition}, pp. \bibinfo{pages}{10938--10947}.
%Type = Inproceedings
\bibitem[{Tsai et~al.(2018)Tsai, Hung, Schulter, Sohn, Yang and
  Chandraker}]{learning_to_adapt}
\bibinfo{author}{Tsai, Y.H.}, \bibinfo{author}{Hung, W.C.},
  \bibinfo{author}{Schulter, S.}, \bibinfo{author}{Sohn, K.},
  \bibinfo{author}{Yang, M.H.}, \bibinfo{author}{Chandraker, M.},
  \bibinfo{year}{2018}.
\newblock \bibinfo{title}{Learning to adapt structured output space for
  semantic segmentation}, in: \bibinfo{booktitle}{Proceedings of the IEEE
  conference on computer vision and pattern recognition}, pp.
  \bibinfo{pages}{7472--7481}.
%Type = Inproceedings
\bibitem[{Tzeng et~al.(2017)Tzeng, Hoffman, Saenko and
  Darrell}]{tzeng2017adversarial}
\bibinfo{author}{Tzeng, E.}, \bibinfo{author}{Hoffman, J.},
  \bibinfo{author}{Saenko, K.}, \bibinfo{author}{Darrell, T.},
  \bibinfo{year}{2017}.
\newblock \bibinfo{title}{Adversarial discriminative domain adaptation}, in:
  \bibinfo{booktitle}{Proceedings of the IEEE conference on computer vision and
  pattern recognition}, pp. \bibinfo{pages}{7167--7176}.
%Type = Article
\bibitem[{Tzeng et~al.(2014)Tzeng, Hoffman, Zhang, Saenko and
  Darrell}]{tzeng2014deep}
\bibinfo{author}{Tzeng, E.}, \bibinfo{author}{Hoffman, J.},
  \bibinfo{author}{Zhang, N.}, \bibinfo{author}{Saenko, K.},
  \bibinfo{author}{Darrell, T.}, \bibinfo{year}{2014}.
\newblock \bibinfo{title}{Deep domain confusion: Maximizing for domain
  invariance}.
\newblock \bibinfo{journal}{arXiv preprint arXiv:1412.3474} .
%Type = Inproceedings
\bibitem[{Vu et~al.(2019)Vu, Jain, Bucher, Cord and P{\'e}rez}]{advent}
\bibinfo{author}{Vu, T.H.}, \bibinfo{author}{Jain, H.},
  \bibinfo{author}{Bucher, M.}, \bibinfo{author}{Cord, M.},
  \bibinfo{author}{P{\'e}rez, P.}, \bibinfo{year}{2019}.
\newblock \bibinfo{title}{Advent: Adversarial entropy minimization for domain
  adaptation in semantic segmentation}, in: \bibinfo{booktitle}{Proceedings of
  the IEEE/CVF Conference on Computer Vision and Pattern Recognition}, pp.
  \bibinfo{pages}{2517--2526}.
%Type = Inproceedings
\bibitem[{Wang et~al.(2020)Wang, Wang, Du, Yang, Zhang, Ding, Mardziel and
  Hu}]{wang2020score}
\bibinfo{author}{Wang, H.}, \bibinfo{author}{Wang, Z.}, \bibinfo{author}{Du,
  M.}, \bibinfo{author}{Yang, F.}, \bibinfo{author}{Zhang, Z.},
  \bibinfo{author}{Ding, S.}, \bibinfo{author}{Mardziel, P.},
  \bibinfo{author}{Hu, X.}, \bibinfo{year}{2020}.
\newblock \bibinfo{title}{Score-cam: Score-weighted visual explanations for
  convolutional neural networks}, in: \bibinfo{booktitle}{Proceedings of the
  IEEE/CVF conference on computer vision and pattern recognition workshops},
  pp. \bibinfo{pages}{24--25}.
%Type = Inproceedings
\bibitem[{Wang et~al.(2017)Wang, Yin, Shi, Fang, Li and Wang}]{wang2017zoom}
\bibinfo{author}{Wang, Z.}, \bibinfo{author}{Yin, Y.}, \bibinfo{author}{Shi,
  J.}, \bibinfo{author}{Fang, W.}, \bibinfo{author}{Li, H.},
  \bibinfo{author}{Wang, X.}, \bibinfo{year}{2017}.
\newblock \bibinfo{title}{Zoom-in-net: Deep mining lesions for diabetic
  retinopathy detection}, in: \bibinfo{booktitle}{International Conference on
  Medical Image Computing and Computer-Assisted Intervention},
  \bibinfo{organization}{Springer}. pp. \bibinfo{pages}{267--275}.
%Type = Article
\bibitem[{Wu et~al.(2021)Wu, Gao, Mei, Xu, Fan, Zhang and Cheng}]{Wu_2021}
\bibinfo{author}{Wu, Y.H.}, \bibinfo{author}{Gao, S.H.}, \bibinfo{author}{Mei,
  J.}, \bibinfo{author}{Xu, J.}, \bibinfo{author}{Fan, D.P.},
  \bibinfo{author}{Zhang, R.G.}, \bibinfo{author}{Cheng, M.M.},
  \bibinfo{year}{2021}.
\newblock \bibinfo{title}{{JCS}: An explainable covid-19 diagnosis system by
  joint classification and segmentation}.
\newblock \bibinfo{journal}{IEEE Transactions on Image Processing}
  \bibinfo{volume}{30}, \bibinfo{pages}{3113--3126}.
%Type = Inproceedings
\bibitem[{Xiao et~al.(2019)Xiao, Zou, Yang, Gaudio, Kitani, Smailagic, Costa
  and Xu}]{Xiao_2019}
\bibinfo{author}{Xiao, Q.}, \bibinfo{author}{Zou, J.}, \bibinfo{author}{Yang,
  M.}, \bibinfo{author}{Gaudio, A.}, \bibinfo{author}{Kitani, K.},
  \bibinfo{author}{Smailagic, A.}, \bibinfo{author}{Costa, P.},
  \bibinfo{author}{Xu, M.}, \bibinfo{year}{2019}.
\newblock \bibinfo{title}{Improving lesion segmentation for diabetic
  retinopathy using adversarial learning}, in:
  \bibinfo{booktitle}{International Conference on Image Analysis and
  Recognition}, \bibinfo{organization}{Springer}. pp.
  \bibinfo{pages}{333--344}.
%Type = Inproceedings
\bibitem[{Yan et~al.(2019)Yan, Wang, Gu, Huang, Yan, Xia and
  Tao}]{yan2019domain}
\bibinfo{author}{Yan, W.}, \bibinfo{author}{Wang, Y.}, \bibinfo{author}{Gu,
  S.}, \bibinfo{author}{Huang, L.}, \bibinfo{author}{Yan, F.},
  \bibinfo{author}{Xia, L.}, \bibinfo{author}{Tao, Q.}, \bibinfo{year}{2019}.
\newblock \bibinfo{title}{The domain shift problem of medical image
  segmentation and vendor-adaptation by unet-gan}, in:
  \bibinfo{booktitle}{International Conference on Medical Image Computing and
  Computer-Assisted Intervention}, \bibinfo{organization}{Springer}. pp.
  \bibinfo{pages}{623--631}.
%Type = Article
\bibitem[{Yanase and Triantaphyllou(2019)}]{yanase2019systematic}
\bibinfo{author}{Yanase, J.}, \bibinfo{author}{Triantaphyllou, E.},
  \bibinfo{year}{2019}.
\newblock \bibinfo{title}{A systematic survey of computer-aided diagnosis in
  medicine: Past and present developments}.
\newblock \bibinfo{journal}{Expert Systems with Applications}
  \bibinfo{volume}{138}, \bibinfo{pages}{112821}.
%Type = Inproceedings
\bibitem[{Yang et~al.(2017)Yang, Li, Li, Wu, Fan and Zhang}]{yang2017lesion}
\bibinfo{author}{Yang, Y.}, \bibinfo{author}{Li, T.}, \bibinfo{author}{Li, W.},
  \bibinfo{author}{Wu, H.}, \bibinfo{author}{Fan, W.}, \bibinfo{author}{Zhang,
  W.}, \bibinfo{year}{2017}.
\newblock \bibinfo{title}{Lesion detection and grading of diabetic retinopathy
  via two-stages deep convolutional neural networks}, in:
  \bibinfo{booktitle}{International Conference on Medical Image Computing and
  Computer-Assisted Intervention}, \bibinfo{organization}{Springer}. pp.
  \bibinfo{pages}{533--540}.
%Type = Article
\bibitem[{Yang et~al.(2021)Yang, Shang, Wu, Yang, Wang, Xu, Zhang and
  Zhang}]{yang2021robust}
\bibinfo{author}{Yang, Y.}, \bibinfo{author}{Shang, F.}, \bibinfo{author}{Wu,
  B.}, \bibinfo{author}{Yang, D.}, \bibinfo{author}{Wang, L.},
  \bibinfo{author}{Xu, Y.}, \bibinfo{author}{Zhang, W.},
  \bibinfo{author}{Zhang, T.}, \bibinfo{year}{2021}.
\newblock \bibinfo{title}{Robust collaborative learning of patch-level and
  image-level annotations for diabetic retinopathy grading from fundus image}.
\newblock \bibinfo{journal}{IEEE Transactions on Cybernetics} .
%Type = Inproceedings
\bibitem[{Yu et~al.(2021)Yu, Ma, Bi, Bian, Ning, He, Li, Liu and
  Zheng}]{yu2021mil}
\bibinfo{author}{Yu, S.}, \bibinfo{author}{Ma, K.}, \bibinfo{author}{Bi, Q.},
  \bibinfo{author}{Bian, C.}, \bibinfo{author}{Ning, M.}, \bibinfo{author}{He,
  N.}, \bibinfo{author}{Li, Y.}, \bibinfo{author}{Liu, H.},
  \bibinfo{author}{Zheng, Y.}, \bibinfo{year}{2021}.
\newblock \bibinfo{title}{Mil-vt: Multiple instance learning enhanced vision
  transformer for fundus image classification}, in:
  \bibinfo{booktitle}{International Conference on Medical Image Computing and
  Computer-Assisted Intervention}, \bibinfo{organization}{Springer}. pp.
  \bibinfo{pages}{45--54}.
%Type = Inproceedings
\bibitem[{Zhou et~al.(2016)Zhou, Khosla, Lapedriza, Oliva and
  Torralba}]{zhou2016learning}
\bibinfo{author}{Zhou, B.}, \bibinfo{author}{Khosla, A.},
  \bibinfo{author}{Lapedriza, A.}, \bibinfo{author}{Oliva, A.},
  \bibinfo{author}{Torralba, A.}, \bibinfo{year}{2016}.
\newblock \bibinfo{title}{{Learning deep features for discriminative
  localization}}, in: \bibinfo{booktitle}{Proceedings of the IEEE Conference on
  Computer Vision and Pattern Recognition}, pp. \bibinfo{pages}{2921--2929}.
%Type = Inproceedings
\bibitem[{Zhou et~al.(2019)Zhou, He, Huang, Liu, Zhu, Cui and
  Shao}]{zhou2019collaborative}
\bibinfo{author}{Zhou, Y.}, \bibinfo{author}{He, X.}, \bibinfo{author}{Huang,
  L.}, \bibinfo{author}{Liu, L.}, \bibinfo{author}{Zhu, F.},
  \bibinfo{author}{Cui, S.}, \bibinfo{author}{Shao, L.}, \bibinfo{year}{2019}.
\newblock \bibinfo{title}{Collaborative learning of semi-supervised
  segmentation and classification for medical images}, in:
  \bibinfo{booktitle}{Proceedings of the IEEE/CVF Conference on Computer Vision
  and Pattern Recognition}, pp. \bibinfo{pages}{2079--2088}.
%Type = Article
\bibitem[{Zhou et~al.(2020)Zhou, Wang, Huang, Cui and Shao}]{zhou2020benchmark}
\bibinfo{author}{Zhou, Y.}, \bibinfo{author}{Wang, B.}, \bibinfo{author}{Huang,
  L.}, \bibinfo{author}{Cui, S.}, \bibinfo{author}{Shao, L.},
  \bibinfo{year}{2020}.
\newblock \bibinfo{title}{A benchmark for studying diabetic retinopathy:
  Segmentation, grading, and transferability}.
\newblock \bibinfo{journal}{IEEE Transactions on Medical Imaging}
  \bibinfo{volume}{40}, \bibinfo{pages}{818--828}.

\end{thebibliography}

\end{document}